\documentclass{article}
\usepackage{comment}

\DeclareUnicodeCharacter{2212}{-} 
\usepackage{qtree}
\usepackage{arxiv}
\usepackage{longtable}
\usepackage{todonotes}
\usepackage{float}
\usepackage{tablefootnote}
\usepackage[utf8]{inputenc} 
\usepackage[T1]{fontenc}    
\usepackage[para]{footmisc}
\usepackage{hyperref}       
\usepackage{url}            
\usepackage{booktabs}       
\usepackage{amsfonts}       
\usepackage{nicefrac}       
\usepackage{microtype}      
\usepackage{lipsum}		
\usepackage{booktabs}
\usepackage{graphicx}
\usepackage{doi}
\usepackage{longtable}
\usepackage{multirow}
\usepackage{pdflscape}
\usepackage{amsmath,amssymb}

\usepackage{cite}

\usepackage{pdflscape}
\usepackage{tikz}
\usetikzlibrary{positioning}
\usetikzlibrary{shapes,arrows,chains,shadows,trees,mindmap}
\tikzset{every node/.append style={scale=1}}

\DeclareMathOperator{\EX}{\mathbb{E}}

\DeclareMathOperator*{\argmax}{arg\,max}
\DeclareMathOperator*{\argmin}{arg\,min}

\usepackage{tikz,colortbl}
\usetikzlibrary{calc}
\usepackage{zref-savepos}

\newcounter{NoTableEntry}
\renewcommand*{\theNoTableEntry}{NTE-\the\value{NoTableEntry}}

\title{Mixed-Precision Neural Networks: A Survey}

\author{ \href{https://orcid.org/0000-0002-2514-7960}{\includegraphics[scale=0.06]{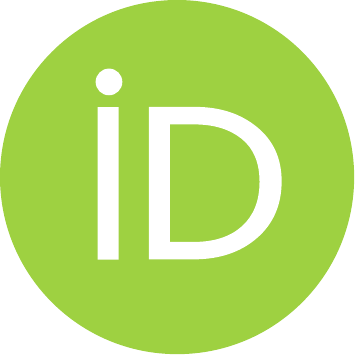}\hspace{1mm}Mariam Rakka}, \href{https://orcid.org/0000-0001-7139-3428}{\includegraphics[scale=0.06]{orcid.pdf}\hspace{1mm}Mohammed E. Fouda\thanks{Currently with Rain Neuromorphics Inc}}, 
\hspace{1mm}Pramod Khargonekar
,\,  and \href{https://orcid.org/0000-0002-6982-365X}{\includegraphics[scale=0.06]{orcid.pdf}
\hspace{1mm}Fadi Kurdahi}\\
Electrical Engineering and Computer Science Dept., University of California, Irvine, CA, USA, 92697. \\
	\texttt{Emails: \{mrakka, foudam, pramod.khargonekar, kurdahi\}@uci.edu} \\
}

\renewcommand{\shorttitle}{\textit{arXiv} Template}

\hypersetup{
pdftitle={A template for the arxiv style},
pdfsubject={q-bio.NC, q-bio.QM},
pdfauthor={David S.~Hippocampus, Elias D.~Striatum},
pdfkeywords={First keyword, Second keyword, More},
}

\begin{document}
\setlength{\footskip}{60pt}
\maketitle

\begin{abstract}

Mixed-precision Deep Neural Networks achieve the energy efficiency and throughput needed for hardware deployment, particularly when the resources are limited, without sacrificing accuracy. However, the optimal per-layer bit precision that preserves accuracy is not easily found, especially with the abundance of models, datasets, and quantization techniques that creates an enormous search space. In order to tackle this difficulty, a body of literature has emerged recently, and several frameworks that achieved promising accuracy results have been proposed. In this paper, we start by summarizing the quantization techniques used generally in literature. Then, we present a thorough survey of the mixed-precision frameworks, categorized according to their optimization techniques such as reinforcement learning and quantization techniques like deterministic rounding. Furthermore, the advantages and shortcomings of each framework are discussed, where we present a juxtaposition. We finally give guidelines for future mixed-precision frameworks.

\end{abstract}


\section{Introduction}

With the thrive of Deep Neural Networks (DNNs) in the fields of object detection \cite{szegedy2013deep}, autonomous cars \cite{tian2018deeptest}, the internet of things (IoT) \cite{hadidi2020toward}, genomics \cite{rentzsch2021cadd}, and smart cities \cite{mohammadi2018enabling}, questions started arising regarding the feasibility of deploying DNNs on hardware platforms limited by computational resources and/or power constraints. Even though they achieved high accuracy, DNNs have not been widely deployed yet on a large scale, especially on embedded systems and mobile phones. In hopes of mitigating the expensive overhead usually imposed by DNNs, many works in literature proposed "hardware-friendly" solutions. These include the following; pruning \cite{molchanov2019importance,li2016pruning, han2015learning, lecun1989optimal,yang2017designing,mao2017exploring}, new DNN designs \cite{iandola2016squeezenet, sandler2018mobilenetv2, tan2019efficientnet}, DNN architectures and hardware co-design \cite{gholami2018squeezenext, han2017efficient, howard2019searching, wu2019fbnet}, knowledge distillation \cite{hinton2015distilling, mishra2017apprentice, polino2018model,yin2020dreaming}, and quantization \cite{morgan1991experimental,bhalgat2020lsq+,chin2020one,hubara2016binarized,jacob2018quantization,rastegari2016xnor,zhang2018lq,zhou2017incremental,sharma2018bit,dong2019hawq,bulat2021bit}. Quantization is the focus of this survey which is shown to be best way to achieve orders of improvement in the energy and a latency.

Quantization which dates back to 1990 \cite{fiesler1990weight} presents itself as a promising solution as it allows representing floating-point values of weights and/or activations and/or gradients in a fewer number of bits. As such, it reduces the memory footprint, computational complexity, and memory traffic volume \cite{cai2020zeroq,horowitz20141,cai2020rethinking}, and renders itself suitable for support by existing hardware like CPUs and FPGAs. 

Networks that have a 1-bit precision for weights and/or activations are known as Binary Neural Networks (BNNs) \cite{qin2020binary}. While Floating Point (FP) DNNs are the most accurate, BNNs are the most efficient. However, quantizing the DNN parameters into ultra-low precision severely degrades the accuracy. One way to mitigate this degradation is to carry out "fine-tuning", which is retraining for a short number of epochs, such as what was done in \cite{zhou2016dorefa,choi2018pact,rastegari2016xnor} and \cite{zhang2018lq}. We denote quantization techniques that rely on fine-tuning for accuracy purposes as "retraining" quantization techniques. This introduces a new trade-off: accuracy vs. time consumption and computational complexity. In fact, there exists a factorial complexity in order to decide on the order of fine-tuning \cite{dong2019hawq}. As such, works in literature like \cite{kravchik2019low} and \cite{banner2018post} have been proposed to quantize the network without fine-tuning, and these techniques became famous as "post-training" quantization techniques. But some of these techniques still suffer from an unexpected loss in performance. Moreover, some of the "post-training" techniques need to access some unlabeled data for directing the training process, which is not always possible such as in the case of medical data where information is confidential \cite{cai2020zeroq}. Other recent works in literature try to reduce accuracy degradation by relying on non-uniform quantizers \cite{zhang2018lq} and channel-wise quantization \cite{krishnamoorthi2018quantizing}.

Early works on quantization \cite{zhou2016dorefa,hubara2016binarized,rastegari2016xnor,jacob2018quantization,jin2019towards,zhang2018lq,mellempudi2017ternary} used to quantize some/all the parameters in a DNN in the same fashion across all layers, for example allocating 8-bit precision for weights/activations in all the layers. These frameworks are known as fixed-precision frameworks. While promising higher accuracy compared to BNNs, one could argue that fixed-precision frameworks are nonoptimal. In particular, the distribution of weights (and activations) has been shown by \cite{han2015deep} to be bell-shaped, so allocating the same precision across all the layers does not match that distribution. Also, each layer in the DNN has its unique structure, role, and characteristics which are portrayed in the weight/activation distribution \cite{park2018value,huang2021mxqn} and \cite{elthakeb2018releq}. Since DNNs are widely known to be over-parametrized \cite{allen2018learning}, each layer has its own redundancy profile. Hence, in each layer, the parameters should be, intuitively, allocated different bitwidths. Otherwise, the allocation would be sub-optimal.

DNNs with different bit precision allocated for the parameters across the different layers are known as mixed-precision DNNs (MXPDNNs).  In MXPDNNs, some layers are maintained at higher precision, while others are reduced to a lower precision.  In addition to per-layer mixed-precision allocation, bitwidths' allocation can differ in granularity: per-network, per-channel(group, block), or per-parameter \cite{krishnamoorthi2018quantizing}. MXPDNNs promise a trade-off between accuracy and efficiency as they fall in the middle of the spectrum between FP models and BNNs. In addition, they promise more optimal solutions compared with fixed-precision frameworks. Though intuitive, however; the task of allocating different bitwidths for DNN parameters is in fact challenging for several reasons. We summarize the challenges below. 
\begin{enumerate}

    \item The hyper search space for per-layer parameters' bitwidths is exponential, in particular when multiple possible bitwidths are considered and as the granularity becomes finer, which renders manual assignments of these bitwidths laborious if not impossible \cite{elthakeb2018releq,cai2020rethinking}. Also, a brute force approach is not feasible due to the fact that the search space is an exponential function of the number of layers \cite{yao2021hawq}. This calls for having an "automated" or "learned" sort of way to allocate the mixed-precision. 
    
    \item The number of hardware platforms that the DNN algorithm needs to be compatible with is huge. Each of these platforms is characterized by its unique capabilities, so the task of quantizing networks that are compatible with all these platforms is not easy \cite{bulat2021bit}. 
    
    \item On a single hardware platform, the available computing resources will vary at run-time due to changes in other parallel running processes, the depletion of battery, the rise in temperature, and/or the change in task priorities. This means that one quantized network for a single platform might not be optimal for this platform at different times and conditions of operation \cite{bulat2021bit}.
    
    \item The fact that each quantization technique has its own unique calculations imposes a hardship on the gradient descend, whereby the convergence during training in back-propagation becomes non-trivial \cite{huang2021mxqn}.

\end{enumerate}

In this survey, we investigate those works that tackle the challenges posed by mixed-precision on the layer or block granularities. In particular, our contributions can be summarized as follows: 1) Collecting, investigating, and summarizing the early and recent works on MXPDNNs, 2) comparing the different MXPDNN frameworks by commenting on the pros and cons of each of the compiled frameworks, 3) Comparing best performing MXPDNNs with best performing BNNs, and 4) providing guidelines for future MXPDNN frameworks.

The rest of this survey is organized as follows. Section 2 summarizes the types used for quantization in general. Section 3 elaborates on the main frameworks on mixed-precision found in the literature. In section 4, we lay out our discussion, and in section 5 we conclude the work.

\section{Classification of Quantization Techniques} 
In MXPDNNs, each layer is assigned different precision and hence is quantized during the training or after the training. In this section, we summarize the quantization techniques that are used in MXPDNNs. For a more detailed description, we invite the readers to read the surveys done by \cite{guo2018survey} and \cite{gholami2021survey} in this context. We categorize and define the techniques according to their type of classification. As such, the quantization technique is either classified to be 1) deterministic, 2) stochastic, according to the 3) component that is being quantized, or according to the 4) nature of the quantization operator that maps a floating point value to a quantized one. We summarize the quantization techniques along with the relevant references in Table \ref{tab:my-table0}.

\subsection{Deterministic}
What is meant by deterministic quantization is a one-to-one mapping that exists between the "real" (hereon used to mean a floating, non-quantized value) and quantized values. One type of deterministic quantization is rounding. Works like \cite{courbariaux2015binaryconnect}, \cite{zhou2016dorefa} and \cite{rastegari2016xnor} round the real value into a "0" or "1", i.e. utilizing binary representation. For example, \cite{courbariaux2015binaryconnect} used the \textit{sign} function as follows

\begin{equation}
\label{eq1}
    q=sign(r)=  \begin{cases} 
      1 & r\geq 0,\\
     -1 & o.w.  \\
   \end{cases}
\end{equation}

where \textit{q} is the quantized value and \textit{r} is the real value.
Moreover, \cite{qin2020binary} compiled the works done on binary neural networks, which utilize rounding techniques. In \cite{polino2018model}, is a general form of the $sign$ function, where the real value is quantized into multi-levels. To quantize the real value into its closest fixed point, \cite{gupta2015deep} proposed a rounding scheme as follows.

\begin{equation}
    quant(r,[i,f])= \begin{cases} 
      \lfloor{r}\rfloor & \lfloor{r}\rfloor\leq r\leq \lfloor{r}\rfloor+\frac{\epsilon}{2}\\
     \lfloor{r}\rfloor+\epsilon, & \lfloor{r}\rfloor+\frac{\epsilon}{2}\textless r\leq \lfloor{r}\rfloor+\epsilon \\
   \end{cases}
\end{equation}

where r, i, f, and $\epsilon$ denote the real value, number of integer bits, number of fractional bits, and the smallest number (positive) that one can represent in the fixed-point format in the equation. The real values $\not\in [i,f]$ are normalized to either the upper bound or the lower bound of the fixed-point representation. Moreover, \cite{wu2018training}'s rounding scheme converts the real value into a k-bit integer. Vector quantization is yet another deterministic quantization technique where the real parameters are grouped into clusters and then quantization is applied to each cluster. Several extensions to vector quantization have been proposed as well. \cite{han2015deep,choi2016towards,wu2016quantized,park2017weighted}. In \cite{gong2014compressing} the k-means clustering is used to form the weight clusters as follows.

\begin{equation}
    min\sum^{u*v}_i\sum^w_j||weight_{i}-centroid_{j}||^2_2
\end{equation}

where for the weight matrix $W \in R^{uxv}$, the scalars $weight \in R^{1xu*v}$. Each element in \textit{weight} is, then after the clustering, given a cluster index and one can form a codebook with the cluster centers $centroid$ which is a $1xw$ row vector. 
Then, the centroid of each cluster is used to quantize the real weights in that cluster. In addition, deterministic quantization also comprises formulations whereby the quantization is regarded as an optimization problem whose purpose is to find good approximations of the real values like in \cite{rastegari2016xnor,li2016ternary} and \cite{leng2018extremely}. \cite{khoram2018adaptive} formulated the loss function as part of an optimization problem in order to allocate a larger bitwidth for significant weights compared to non-significant weights. Their optimization formulation is as follows

\begin{equation}
    \begin{cases}
    min_{weight} \quad N_{total}(\textbf{weight})=\sum^n_k N_{min}(weight_k) \\
    such \quad that \quad loss(\textbf{weight})\leq U.B.
    \end{cases}
\end{equation}

where $n$ is the total number of parameters, $N_{total}(.)$ is the total number of bits in the network, $N_{min}(.)$ is the minimum number of bits used to represent a \textit{weight} and \textit{U.B.} is utilized an upper bound on the acceptable accuracy lost. \cite{hou2016loss} proposed to binarize the weights that do not increase the loss through the following optimization formulation.

\begin{equation}
    \begin{cases}
    min_{\textbf{weight}} \quad loss(\textbf{weight}) \\
    such \quad that \quad \textbf{weight$_{layer}$}=\alpha_{layer} \textbf{q$_{layer}$} & \alpha_{layer} \textgreater 0, \textbf{q$_{layer}$}\in\{+1,-1\}^{N_{params}}, layer=1,2...,N_{layers}
    \end{cases}
\end{equation}

where $N_{params}$ and $N_{layers}$ are the total number of parameters in \textit{layer} and the total number of layers respectively. In addition, codebook quantization, where the real value is quantized into a codebook, is listed as deterministic quantization. It is worth mentioning that the codebook can be either predefined or learned from the data. In \cite{courbariaux2015binaryconnect}, the predefined codebook $\{-1,1\}$ is used. \cite{rastegari2016xnor} uses a 2-element codebook (binarized) but where the codebook values are scaled to some values. Similarly, \cite{hwang2014fixed} uses a predefined ternary codebook $\{-1,0,1\}$ while \cite{zhu2016trained} uses a scaled ternary codebook. A powers-of-two predefined codebook $\{...,-2^{(-2)},-2^{(-1)},-1,0,+1,+2^{(-1)},2^{(-2)},...\}$ is used in \cite{tang1993multilayer}. Examples of leaned codebooks can be found in \cite{achterhold2018variational,gong2014compressing} and \cite{choi2016towards}.

\subsection{Stochastic}
Stochastic quantization is when the quantized value is sampled from the discrete distribution of its non-quantized counterpart. The first type of stochastic quantization is random rounding. In \cite{courbariaux2015binaryconnect} and \cite{muller2015rounding}, an equation similar to Eqn. \ref{eq1} is used, except that in this case, the conditions are probabilistic:

\begin{equation}
\label{eq1s}
        q=\begin{cases} 
      1 & w/ \quad p=max(0,min(1,\frac{r+1}{2}))\\
     -1 & 1-p  \\
   \end{cases}
\end{equation}

where \textit{p} and \textit{1-p} represent probabilities. \cite{lin2015neural} extended \ref{eq1s} to the ternary case, and in \cite{polino2018model}, the following Eqn. is used for rounding.

\begin{equation}
    quant_{[0,1]}(r)=f^{-1}(quant(f(r)))
\end{equation}

where $quant_{[0,1]}(.)$ is the quantized value scaled by \textit{f} to the range $[0,1]$. $quant(.)$ is the quantization function that is represented as $\frac{\lfloor{r*l}\rfloor}{l}+\frac{\epsilon}{l}$, \textit{l} being the parameter of quantization level. $\epsilon$ is sampled according to the Bernoulli distribution $\epsilon \sim Bernoulli(r*s-\lfloor{r*s\rfloor})$. Also under stochastic quantization is probabilistic quantization, where the parameters are assumed to be discretely distributed or when a discrete posterior distribution on weights is learned via variational inference \cite{soudry2014expectation,shayer2017learning,jordan1999introduction,achterhold2018variational}. For this, the neural network is considered in the Bayesian setting.  

\subsection{DNN Component-Wise}
Here the quantization techniques are categorized according to the DNN component being quantized. In a DNN, there are three main components that one can quantize; the weights, the activations, and the gradients. The motivation behind quantizing the weights is the fact that it can decrease the size of the model and its associated storage as well as the communication cost \cite{wu2018mixed}. Moreover, it reduces the execution time of training and inference \cite{guo2018survey}. The idea behind quantizing activations, besides saving memory overhead, is to make the training process faster by utilizing inner products with binary operations. As for gradients, quantization can cut the cost of the communication during training of large parallel DNNs, besides saving memory \cite{guo2018survey}. 
In \cite{han2015deep,zhu2016trained} and \cite{leng2018extremely} the weights are quantized, while activations are used in full precision. In \cite{courbariaux2015binaryconnect,rastegari2016xnor,zhou2016dorefa,choi2018pact,jung2018joint} and \cite{zhuang2018training} both weights and activations are quantized. Though being the hardest to quantize because of the need for high precision for the optimization algorithm to converge, some works have explored gradient quantization \cite{seide20141,strom2015scalable,alistarh2017qsgd,dryden2016communication}. Mainly those works focus on the context of the data-parallel training framework of DNNs. In such a framework, each node holds a copy of the weights and requires the computation of the sub-gradient, after which it should send the computed sub-gradient to the other nodes. Then to update the weights, the sub-gradient at each node is accumulated from broadcast received from the other nodes \cite{guo2018survey}.

\subsection{Nature of the Quantization Operator}
In this subsection, we categorize the quantization techniques by the quantization operator. The quantization operator maps the floating point value into its quantized counterpart. According to the nature of the quantization operator, we can classify the quantization as uniform/non-uniform and as symmetric/asymmetric. A uniform quantization scheme has a quantization operator that maps the real value into an integer value whereby the resulting quantization levels are uniformly spaced. A popular quantization operator resulting in uniform quantization is as follows.
\begin{equation}
    Q(r)=Int(\frac{r}{S})-Z
\label{uniformmm}
\end{equation}
where $r$ is the real-valued input, $S$ is the real-valued scaling factor, and Z denotes an integer zero point. Moreover, $Int(.)$ function is a rounding function \cite{gholami2021survey}. Some works in literature that utilized uniform quantization include \cite{evoq}. Non-uniform quantization, on the other hand, allows the quantization steps as well as the quantization levels to be non-uniformly spaced. The quantization operator resulting in non-uniform quantization is defined as follows.
\begin{equation}
    Q(r)=X_i, \; if \; r\in [\Delta_i, \Delta_{i+1})
\end{equation}
where $X_i$ denotes the discrete quantization levels, and $\Delta_i$ represents the quantization steps. Both $X_i$ and $\Delta_i$ are non-uniformly spaced \cite{gholami2021survey}. Some of the works in literature that explored non-uniform quantization include \cite{cai2017deep,choi2016towards,zhou2018explicit,zhang2018lq, park2017weighted,park2018value}.

The real-valued scaling factor, $S$, is used in Eqn. \ref{uniformmm} is defined as follows.
\begin{equation}
    S=\frac{b-a}{2^{bw}-1}
\end{equation}
where $bw$ is the bit width, and $[a,b]$ denotes the clipping range that the real values are clipped with. The choice of the clipping range, also known as calibration, renders the quantization technique as symmetric or asymmetric. In a symmetric quantization scheme, the clipping range is chosen to be symmetric (i.e. $-a=b$). In asymmetric quantization (like in \cite{evoq}), $-a\neq b$. Some of the calibration techniques are proposed in \cite{mckinstry2018discovering, migacznvidia,wu2020integer}.

\begin{table}[h]
\centering
\caption{Summary of the quantization techniques.}
\label{tab:my-table0}
\resizebox{\textwidth}{!}{%
\begin{tabular}{|l|l|l|l|}
\hline
Classification Type & Method & Brief Description & References \\ \hline
\multirow{8}{*}{Deterministic} & Rounding & Rounds a real value into quantized value & \begin{tabular}[c]{@{}l@{}}\cite{courbariaux2015binaryconnect,zhou2016dorefa,qin2020binary,polino2018model}\\ \cite{gupta2015deep,wu2018training,rastegari2016xnor}\end{tabular} \\ \cline{2-4} 
 & Vector Quantization & \begin{tabular}[c]{@{}l@{}}Replaces weights by the centroids of the \\ k-mean clusters during inference\end{tabular} & \cite{gong2014compressing,han2015deep,choi2016towards} \\ \cline{2-4} 
 & Product Quantization (PQ) & \begin{tabular}[c]{@{}l@{}}Partitions the weight matrix into\\ multiple sub-matrices and quantizes the latters\end{tabular} & \cite{gong2014compressing} \\ \cline{2-4} 
 & PQ with Error Correction & Extends PQ with improvements & \cite{wu2016quantized} \\ \cline{2-4} 
 & Residual Quantization & Recursively quantized the residuals of clusters & \cite{gong2014compressing} \\ \cline{2-4} 
 & \begin{tabular}[c]{@{}l@{}}Automatic (Multi-Bit)\\ Quantization\end{tabular} & \begin{tabular}[c]{@{}l@{}}Groups weights in clusters\\ according to their entropies\end{tabular} & \cite{park2017weighted} \\ \cline{2-4} 
 & \begin{tabular}[c]{@{}l@{}}Quantization formulated\\ as an Optimization\end{tabular} & \begin{tabular}[c]{@{}l@{}}Solves an optimization problem\\ to approximate quantization values\end{tabular} & \begin{tabular}[c]{@{}l@{}}\cite{rastegari2016xnor,li2016ternary,hou2016loss}\\ \cite{leng2018extremely,khoram2018adaptive}\\\end{tabular} \\ \cline{2-4} 
 & Codebook Quantization & \begin{tabular}[c]{@{}l@{}}Quantizes values into a \\ predefined or learned codebook\end{tabular} & \begin{tabular}[c]{@{}l@{}}\cite{courbariaux2015binaryconnect,rastegari2016xnor,zhu2016trained,tang1993multilayer} \\ \cite{achterhold2018variational,hwang2014fixed,gong2014compressing,choi2016towards}.\end{tabular} \\ \hline
\multirow{2}{*}{Stochastic} & Random Rounding & \begin{tabular}[c]{@{}l@{}}Samples the quantized value\\ from discrete distributions\end{tabular} & \begin{tabular}[c]{@{}l@{}}\cite{courbariaux2015binaryconnect,muller2015rounding,lin2015neural,polino2018model}\end{tabular} \\ \cline{2-4} 
 & Probabilistic Quantization & Learns a multimodal posterior distribution & \begin{tabular}[c]{@{}l@{}}\cite{soudry2014expectation,shayer2017learning,jordan1999introduction,achterhold2018variational}\end{tabular} \\ \hline
\multirow{3}{*}{DNN Component-wise} & Weight Quantization & Quantizes the weights & \begin{tabular}[c]{@{}l@{}}\cite{han2015deep,zhu2016trained,leng2018extremely}\end{tabular} \\ \cline{2-4} 
 & Activation Quantization & \begin{tabular}[c]{@{}l@{}}Quantizes the activations\\ (i.e. the output of the DNN layers)\end{tabular} & \begin{tabular}[c]{@{}l@{}}\cite{courbariaux2015binaryconnect,rastegari2016xnor,choi2018pact,jung2018joint}\\\cite{zhuang2018training,zhou2016dorefa}\end{tabular} \\ \cline{2-4} 
 & Gradient Quantization & \begin{tabular}[c]{@{}l@{}}Quantizes the gradients \\ in the backward propagation\end{tabular} & \begin{tabular}[c]{@{}l@{}}\cite{seide20141,strom2015scalable,alistarh2017qsgd,dryden2016communication}\end{tabular} \\ \hline
 \multirow{2}{*}{Nature of Operator} &  Uniform/Non-uniform & \begin{tabular}[c]{@{}l@{}}The quantization levels and steps are\\ uniformly/non-uniformly spaced \end{tabular}  & \begin{tabular}[c]{@{}l@{}} \cite{cai2017deep,choi2016towards,zhou2018explicit,zhang2018lq} \\ \cite{ park2017weighted,park2018value,gholami2021survey, evoq} \end{tabular}\\ \cline{2-4} 
 & Symmetric/Asymmetric & The calibration is symmetric/asymmetric & \cite{evoq,gholami2021survey} \\ \hline
\end{tabular}%
}
\end{table}

\section{Frameworks for MXPDNNs}
In this section, we elaborate on different works of literature done on mixed-precision. In particular, we focus on four aspects for each framework: 1) goal and problem formulation, 2) Quantization and Optimization Techniques, 3) Benchmarks Used and 4) Results and Contributions.  At the end, we have summarized the key aspects of each framework in Tables \ref{tab:my-table11}, \ref{tab:my-table12}, and \ref{tab:my-table13}. We note that for the quantization technique, we either categorize it as "training-aware", "retraining" or "post-training" as shown in Fig. \ref{mxdpdnn}. We denote frameworks that train from scratch as "training-aware", those that start from a pre-trained full precision network and quantize with fine-tuning or retraining as "retraining", and those that start from a pre-trained full precision network and quantize without retraining as "post-training". We divide the MXPDNN frameworks according to their optimization technique. In particular, when the bitwidth is found by modifying the gradient descent, we denote the optimization technique "gradient-based". When a proposed heuristic is used to compute the sensitivity of the layers and accordingly assign precisions, we denote the optimization process as "heuristic-based". When a differential search or evolutionary search is utilized for the optimization, we categorize the framework under "meta-heuristic-based" optimization. Finally, "reinforcement learning-based" optimization is used to categorize frameworks that rely on reinforcement learning or deep reinforcement learning.

\tikzset{
  treenode/.style = {shape=rectangle, rounded corners,
                     draw, align=center,
                     top color=white,
                     bottom color=blue!20},
  root/.style     = {treenode, font=\Large,
                     bottom color=red!30},
  env/.style      = {treenode, font=\ttfamily\normalsize},
  dummy/.style    = {circle,draw},
}

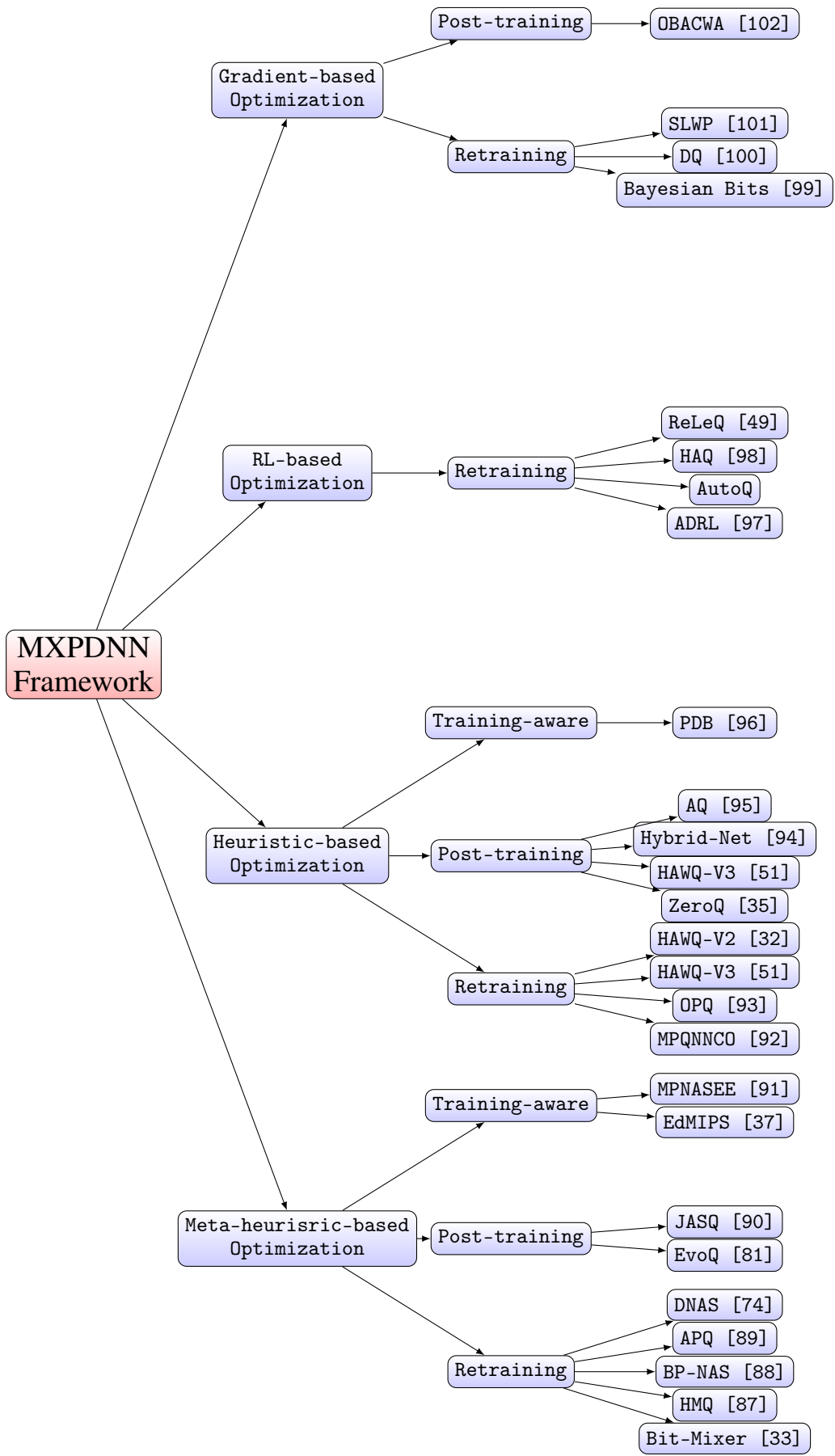
\begin{figure}
\centering
\begin{tikzpicture}   [
    grow                    = right,
    level distance          = 10em,
    edge from parent/.style = {draw, -latex},
    every node/.style       = {font=\footnotesize},
    sloped,
    level 1/.style={sibling distance=63mm},level 2/.style={sibling distance=21.9mm},level 3/.style={sibling distance=5.5mm}
  ]

\node [root] {MXPDNN\\Framework}
	child {node [env] {Meta-heurisric-based\\Optimization}
	child {node [env] {Retraining}
	child {node [env] {Bit-Mixer \cite{bulat2021bit}}}
	child {node [env] {HMQ \cite{habi2020hmq}}}
	child {node [env] {BP-NAS \cite{yu2020search}}}
	child {node [env] {APQ \cite{wang2020apq}}}
	child {node [env] {DNAS \cite{wu2018mixed}}}
	}
	child {node [env] {Post-training}
	child {node [env] {EvoQ\cite{evoq}}}
	child {node [env] {JASQ\cite{chen2018joint}}}
	}
	child {node [env] {Training-aware}
	child {node [env] {EdMIPS \cite{cai2020rethinking}}}
	child {node [env] {MPNASEE \cite{gong2019differentiable}}}
	}
	}
	child {node [env] {Heuristic-based\\Optimization}
	child {node [env] {Retraining}
	child {node [env] {MPQNNCO \cite{chen2021towards}}}
	child {node [env] {OPQ \cite{hu2021opq}}}
	child {node [env] {HAWQ-V3 \cite{yao2021hawq}}}
	child {node [env] {HAWQ-V2 \cite{dong2019hawq}}}
	}
	child {node [env] {Post-training}
	child {node [env] {ZeroQ \cite{cai2020zeroq}}}
	child {node [env] {HAWQ-V3 \cite{yao2021hawq}}}
	child {node [env] {Hybrid-Net \cite{chakraborty2020constructing}}}
	child {node [env] {AQ \cite{zhou2018adaptive}}}
	}
	child {node [env] {Training-aware}
	child {node [env] {PDB \cite{chu2019mixed}}}
	}
	}
	child {node [env] {RL-based\\Optimization}
	child {node [env] {Retraining}
	child {node [env] {ADRL \cite{ning2020simple}}}
	child {node [env] {AutoQ}}
	child {node [env] {HAQ \cite{wang2019haq}}}
	child {node [env] {ReLeQ \cite{elthakeb2018releq}}}
	}
	}
	child {node [env] {Gradient-based\\Optimization}
	child {node [env] {Retraining}
	child {node [env] {Bayesian Bits \cite{van2020bayesian}}}
	child {node [env] {DQ \cite{uhlich2019differentiable}}}
	child {node [env] {SLWP \cite{lacey2018stochastic}}}
	}
	child {node [env] {Post-training}
	child {node [env] {OBACWA \cite{zhe2019optimizing}}}
	}
	};

\end{tikzpicture}
\caption{Mixed Precision Frameworks grouped according to their optimization and quantization techniques.}
\label{mxdpdnn}
\end{figure}

\subsection{Gradient-Based Optimization}
\subsubsection{Stochastic Layer-Wise Precision in Deep Neural Networks (SLWP)}
In \cite{lacey2018stochastic}, the authors' purpose is to learn a heterogeneous (mixed) allocation of bitwidths for weights and activations across the layers. For that, they formulate their problem as a constrained, stochastically allocated mixed bit precision via modified gradient descend. For the quantization part (after bitwidth allocation), they rely on deterministic rounding to find the quantized value via Straight-Through Estimator (STE) \cite{bengio2013estimating}, after the bitwidths were stochastically sampled from a categorical distribution. Their quantization methodology can be classified as retraining.

The motivation behind the SLWP framework is to tackle two issues that are encountered when allocating mixed-precision bitwidths; the first is the fact that the number of bits assigned to a parameter (be it weight or activation) tends to be large as higher precision naturally leads to lower losses (and the algorithm will hence favor higher bitwidth precision allocation) and the second is the fact that discrete non-differentiable quantization cannot be integrated with the vanilla backpropagation techniques such as gradient descend. To solve the first issue, a precision budget, \textit{B}, is defined such that the sum of allocated bits across the different layers must equal \textit{B}. For the second issue, they rely on a stochastic allocation of precision inspired by \cite{srivastava2014dropout} and on modifying the backpropagation technique to handle stochastic discrete operations by utilizing the Gumbel-Softmax distribution \cite{jang2016categorical}. The Gumbel-Softmax distribution is a continuous distribution on the simplex which approximates categorical samples. A new layer, "the precision allocation layer", is introduced in the forward pass (as a leaf in the computational graph) of the training algorithm responsible for constructing the Gumbel-Softmax distribution. In particular, SLWP  can be summarized as follows; upon each incoming input and for bitwidth allocation, SLWP relies on the proposed "precision allocation layer" to sample the values (bitwidths) from the Gumbel-Softmax distribution for all layers, then each layer is quantized according to the sampled bitwidths. The process is repeated until the sum of the number of bits allocated across layers is equal to the budget.

In order to evaluate the SLWP framework, the authors rely on the following benchmarks: MNIST tested on a 5-layer model similar to LeNet-5 \cite{lecun1998gradient} and ImageNet tested on AlexNet \cite{krizhevsky2012imagenet}. They choose to quantize both weights and activations across all layers in the experiments, except for AlexNet, where the first layer is kept in full precision. For comparison, they choose their baselines as networks that adapt a fixed precision across all the layers. Three scenarios are mainly considered; 1) all layers are allocated 2 bitwidths for weights and activations, 2) all layers are allocated 4 bitwidths, and 3) all layers are allocated 8 bitwidths. For fairness, the number of layers per network is either 5 to match their first benchmark or 7 to match their second. In all scenarios, the budget is chosen according to the baseline; for example for the ImageNet on AlexNet benchmark, and for a baseline of 2 bitwidth allocation, the budget is $2*\#layers=2*7=14$. Moreover, the top-1\% error is reported as the metric for measuring performance.

The MNIST results show that the SLWP framework learns the per-layer bitwidths and finally allocates fixed bitwidth assignments across the 5 layers, similar to the baselines with lower \% error ( 2.32\% maximum error was reported for 2 bitwidth precision assignment) compared to the baselines. As for the experiments on ImageNet, the SLWP framework converges with different per-layer bitwidth assignments and lower \% errors compared to the baselines, which suggests that the framework is more effective in a "mixed" bitwidth assignment sense on more complex networks. Finally, the contributions of the authors of this framework can be summarized as follows; the introduction of a stochastic sampling technique for training in order to allocate different bitwidths across layers with a training-aware quantization approach. SLWP framework outperforms their chosen baselines in terms of speed and accuracy. The authors do not report model size savings as their baselines are chosen in low precision.

\subsubsection{Differentiable Quantization of Deep Neural Networks (DQ)}

\cite{uhlich2019differentiable} proposes differentiable quantization, DQ, where the framework learns the quantization step size and dynamic range then consequently infers the bitwidth (of weights and activations) per layer via gradient descent. The training process employed by DQ and that quantizes the parameters has computational overhead similar to training a floating-point full precision network, unlike RL-based mixed quantization techniques. DQ uses two quantization techniques, uniform rounding and power-of-two (i.e. codebook), where both are deterministic. Moreover, the framework's quantization techniques are under retraining.

The main motivation behind the DQ framework is to utilize quantizers that can be differentiated with respect to their parameters. The authors start by showing that each quantization technique (rounding/power-of-two) has its suitable parametrization (in terms of step size, dynamic range, or bitwidth). After that, they introduce memory constraints to the training of the differentiable quantizers. We follow the same order of description next.

First, it is shown that a uniform quantization can be represented in three parametrizations using a combination of two of the three parameters: step-size, dynamic-range, and bitwidth. We note that each parametrization has its own gradient and that all the parameters in a parametrization are related. The first parametrization -\textit{U1}- is in terms of step-size and bitwidth, the second -\textit{U2}- is in terms of bitwidth and maximum value (i.e. in terms of dynamic range since dynamic range is equal to the difference between the maximum and minimum quantized values), and the third -\textit{U3}- is in terms of maximum value quantized and step-size. Neither \textit{U1} nor \textit{U2} satisfies both of the ideal properties of gradients: having a bounded gradient magnitude that does not vary a lot and having the gradient vectors as unit vectors, but \textit{U3} shows the most potential in that sense. The best parametrization (as per the two experiments below) is found to be \textit{U3}. Similarly, for power-of-two quantization, it is shown that the parametrization that is well suited for training is that which depends on the dynamic range (i.e. on maximum and minimum quantized values), denoted hereon by \textit{P}. The choice of the best parametrization was made by conducting two experiments. The first experiment quantizes $10E4$ Gaussian data points (by rounding technique and then by power-of-two technique separately), by the use of the 3 proposed parametrizations of each quantization technique. Then, for each quantization technique, the error surface of each parametrization is plotted. The best parametrization is that which lies on the border of the flat region on the error surface graph. In the second experiment, ResNet-20 is trained on CIFAR-10, with quantized activations and weights. For the two quantization techniques, each parametrization is applied and the validation and training errors are plotted. The best parametrization is that which converges to the best local optimum and has the smallest validation error oscillations. In both of these experiments, \textit{U3} and \textit{P} were the best parametrizations for uniform and power-of-two quantizations respectively. It then follows that the bitwidths can be inferred from the best parametrizations.

Now we briefly describe what memory constraints are introduced with DQ in training. The authors focus on three types of memory constraints applied during training: "total memory to store all weights" denoted \textit{C1}, "total activation memory to store all feature maps" denoted \textit{C2}, and "maximum activation memory to store the largest feature map" denoted \textit{C3}. \textit{C1} is used to impose an upper weight memory size on the total memory requirement of weights and \textit{C2} is utilized to guarantee that the total memory for activation is below a maximum size. As for \textit{C3}, gives an upper limit on the maximum activation size. As such, with these constraints, the optimization problem is rendered as constrained. To solve the constrained optimization problem via the usual SGD, a penalty \cite{bertsekas2014constrained} is used to render it unconstrained.

DQ is evaluated with ResNet-20 on CIFAR-10 and with MobileNetV2 and ResNet-18 on ImageNet. In all the experiments, the best parametrization (see earlier) is used for rounding and power-of-two quantization, and for initialization, the pre-trained networks with full precision weights and activations are used. Two baselines are used: full precision network and fixed precision network across all layers. In some experiments, only weights are quantized, while in others both weights and activations are quantized. The model size (in terms of weights and activations) is reported, as well as the validation \% error. Moreover, the most restrictive quantization is reported (with maximum quantized value and bitwidth fixed), and results are reported when the DQ framework is trained with/without one or more memory constraint(s).

The reported results for the CIFAR-10 data set reveal that the maximum validation error obtained was 11.29\% when both weights and activations undergo power-of-two quantization with both \textit{C1} and \textit{C2} imposed. The error is smallest when all the quantization parameters are learned. For the experiments done on ImageNet, the best performing DQ has less than 0.5\% loss in accuracy compared to the full precision baseline. To sum up, DQ infers the per-layer bitwidths allocation from quantizers that are parametrized by step-size and dynamic range. It is competitive with other quantization schemes and does not require multiple retraining like RL-based mixed-precision works.

\subsubsection{Optimizing the Bit Allocation for Compression of Weights and Activations of Deep Neural Networks (OBACWA)}
The authors in \cite{zhe2019optimizing} formulate the mixed-precision problem of weights and activations as a Lagrangian optimization problem whose solution is the optimum joint precision allocation for all weights and activations. The quantization technique used is deterministic, post-training.

In particular, a relationship between the quantized weights and activations and the output error of a neural network is found. First, the output error due to quantizing the weights of layer $i$ ($i \in \{1,...,L\}$), $W_i$, is defined as follows.

\begin{equation}
    Error_{W_i}=\frac{E(D(O,O_m))}{||O||}
\end{equation}
where $E(.)$ is the expected value, $D(.)$ is the squared Euclidean distance, $O$ is the original output, $O_m$ is the modified output, and $||O||$ is the dimension of the original output. Similarly, $Error_{A_i}$ is defined as the output error due to quantizing the activations of layer $i$, $A_i$. Relying on the aforementioned notions, the following relationship is found.
\begin{equation}
    Error_t=Error_{W_1}+...+Error_{W_L}+Error_{A_1}+...+Error_{A_L}
\end{equation}
where $Error_t$ is the output error due to quantizing all the layers' weights and activations. The mixed-precision problem is then formulated as follows.
\begin{equation}
\begin{aligned}
    arg\min_{R_{W_i},R_{A_i}} \quad & Error_t\\
    \textrm{s.t.} \quad & R_{W_1}+...+R_{W_L}+R_{A_1}+...+R_{A_L}
\end{aligned}
\end{equation}
where $R_{W_i}$ and $R_{A_i}$ denote the total bit-rates of weights and activations in the $i$th layer respectively. The solution of the above optimization problem is the optimum bit allocation of weights and activations that minimizes the output error given a bit budget constraint for weights and activations. To solve the optimization problem, a classical Lagrangian rate-distortion formulation is utilized, where the Lagrangian cost function is as follows.
\begin{equation}
    C=\sum_i(Error_{W_i}+Error_{A_i})-\lambda*\sum_i(R_{W_i}+R_{A_i})
\end{equation}
where $\lambda$ is the Lagrange multiplier.

This framework is evaluated by quantizing VGG-16 and ResNet-50 on ImageNet. It is compared to other compression frameworks in literature; Binarized Weight Networks (BWN) \cite{rastegari2016xnor}, Ternary Weight Network (TWN) \cite{li2016ternary}, Incremental Network Quantization (INQ) \cite{zhou2017incremental}, Fine-grained Quantization (FGQ) \cite{mellempudi2017ternary}, Integer Arithmetic-only Inference (IAOI) \cite{jacob2018quantization}, Adaptive Quantization (AQ) \cite{zhou2018adaptive}, Compression Learning by In-parallel Quantization (CLIP-Q) \cite{tung2018clip}, Symmetric Quantization (SYQ) \cite{faraone2018syq}, and Two-bit Shift Quantization (TBSQ) \cite{leng2018extremely}. In all experiments, the top-1\% accuracies are reported along with the compression rates. Also, activations are not quantized in the last layer.

For VGG-16, the proposed framework incurs no accuracy loss (compared to the floating full precision model) and yields a 4x compression rate when weights and activations have an average bitwidth of 8 bits. When weights and activations have an average bitwidth of 4 bits, the framework achieves an 8x compression rate with only a 0.2\% drop in accuracy. For ResNet-50, the proposed framework incurs a 0.4\% and 1.0\% drop in accuracy while achieving a 4x and 8x compression rate when weights and activations have an average bitwidth of 8 and 4 bits respectively. In all experiments, OBACWA outperforms the other compression frameworks, except TBSQ for VGG-16. It is good to note that TBSQ relies on retraining.

In conclusion, OBACWA is a framework that jointly assigns mixed-precision for weights and activations by relating the quantization of weights and activations to the output error due to quantization, then relying on a Lagrangian formulation to solve the optimization problem. The results reported reveal that the proposed framework achieves high accuracies with high compression rates compared to floating full precision models and other compression frameworks.

\subsubsection{Bayesian Bits: Unifying Quantization and Pruning (Bayesian Bits)}
\cite{van2020bayesian} propose Bayesian Bits, a framework that co-optimizes - via gradient descend - per-tensor mixed-precision quantization and pruning (for weights and activations). After finding the effectively power-of-two bitwidths (through a novel quantization decomposition), the quantization technique employed is deterministic rounding. Since the framework starts from pre-trained models, the quantization is categorized as retraining.

Bayesian Bits first introduces a novel quantization decomposition to guarantee that the bitwidths are powers-of-two, which in turn facilitates the hardware implementation. The quantization decomposition is summarized as follows. The residual error between a current quantized value $q_{current}$ whose bitwidth is $b_{current}$ and the next quantized value,$q_{next}$, whose bitwidth is the next power-of-two value $b_{next}=b_{current}*2$ is denoted as $\epsilon_{next}=q_{next}-q_{current}$. The scheme recursively quantizes the residual error of lower bitwidth quantization as follows.
\begin{equation}
    \epsilon_{next}=s_{next}*RNI(\frac{x-q_{current}}{s_{next}})
\end{equation}
where RNI is a function that round to the nearest integer, $x$ is the input floating point value in the range $[t1,t2]$, $s_{current}$ is the step-size of the quantizer that is related to the bitwidth $b$ as: $s_{current}=\frac{t2-t1}{2^{b_{current}}-1}$, $s_{next}=\frac{s_{current}}{2^{b_{current}}+1}$. For this decomposition scheme, the inputs should belong to the quantization grid $[t1, t2]$. As such the inputs are clipped according to PACT \cite{choi2018pact} before applying the quantizer. 
Those quantized residual error tensors are added together to form the quantized approximation of the original tensor (hence quantizing to a specific bitwidth is seen as a gated addition of quantized residuals). After applying the aforementioned decomposition scheme, a learnable gate (i.e. a variable that $\in \{0,1\}$) is placed on each of the quantized residual error tensors. As such, the effective bitwidth can be controlled, optimized in a data-dependant manner, and learned jointly with the quantization scales and the DNN parameters. The gate of the lowest bitwidth possible allows pruning as well, since deactivating this gate (i.e. setting it to 0) will yield a quantized value having 0-bits i.e. pruned. The learning algorithm should utilize efficient gradient-based optimization while favoring gate configurations that yield efficient DNN. To have such a learning algorithm, Bayesian Bits relies on variational inference. As such, a gate regularizing term is derived via a prior which favors lower bitwidth configurations, and this term is added to the objective function. This objective function needs the expensive computation of an expectation of the log-likelihood with respect to the stochastic gates. To mitigate this complexity, Bayesian Bits approximates the gradient by exploiting the hard-concrete relaxations proposed by \cite{louizos2017learning}, thus allowing the optimization through the reparametrization trick proposed by \cite{kingma2013auto,rezende2014stochastic}. Moreover, the Straight-Through Estimator (STE) \cite{bengio2013estimating} is used for the gradients of the network parameters (for fine-tuning).

Bayesian Bits is evaluated by quantizing all the weights and activations (except the output logits) of LeNet-5 on MNIST, and VGG-7 on CIFAR-10. In addition ablation studies (some with combined pruning and mixed quantization, some with pruning with fixed precision quantization, and some with mixed-precision quantization without pruning) are done on ResNet-18 and MobileNet-V2 with ImageNet. The batch normalization layers are treated as in \cite{krishnamoorthi2018quantizing}. Even-though Bayesian Bits supports joint training of network parameters and mixed quantization with pruning, in the experiments on MNIST and CIFAR-10 only the quantization parameters are updated (includes pruning) on the pre-trained models, whereas the weights (pre-trained) are kept fixed. The Giga Bit Operations (GBOPs) (i.e. \# of multiplication operations x bit widths of the operands) relative to the full precision floating point models is reported along with the test accuracy, learned sparsity, and bitwidths of the models in several settings. The baselines used for the experiments are the full precision models, TWN \cite{li2016ternary}, LR-Net \cite{shayer2017learning}, RQ \cite{louizos2018relaxed}, WAGE \cite{wu2018training}, PACT \cite{choi2018pact}, LSQ \cite{esser2019learned}, AdaRound \cite{nagel2020up}, TQT \cite{jain2019trained}, and/or DQ \cite{uhlich2019mixed}.

The results reported on MNIST and CIFAR-10 reveal that Bayesian Bits offers the least relative GBOPs with a competitive accuracy compared to the other baselines. Different settings for Bayesian Bits are used and their results are reported, and these show that the regularization term can be used to balance the trade-off between accuracy and complexity. The ablation studies on ResNet-18 and MobileNet-V2 vary the regularization settings to highlight the trade-offs between accuracy and complexity and reveal that Bayesian Bits performs well compared to the baselines. In one ablation study, post-training mixed-precision is tested on a pre-trained ResNet-18 model (i.e. no finetuning is used, and only gates and/or quantization ranges are learned), and the Pareto fronts are provided.
Its results show that Bayesian Bits can serve as a midway method between post-training methods that do not require any backpropagation and other methods that require full model fine-tuning. To sum up, Bayesian Bits is a framework that relies on a variational inference to jointly optimize mixed-precision quantization and pruning, by proposing a novel quantization decomposition scheme and incorporating a regularization term in the objective function.

\subsection{Heuristic-Based Optimization}
\subsubsection{Adaptive Quantization for Deep Neural Network (AQ)}
\cite{zhou2018adaptive} introduced AQ, the very first framework to theoretically lay out the correlation between the quantization error of parameters at each layer and the overall model prediction accuracy. The purpose of this framework is to perform mixed bitwidth allocation for the weights (while activations are left in full precision) of each DNN layer according to the measure of correlation, taking an upper limit on the drop in model accuracy. The quantization technique utilized in this framework is deterministic rounding. Since AQ starts from pre-trained models without resorting to fine-tuning after quantization, the quantization is categorized as post-training.

Inspired by the work done by \cite{fawzi2016robustness} to analyze the effect of the input noise on the model, AQ uses a similar method to measure the effect of the noise induced by quantizing the DNN weights (according to optimal per-layer bitwidth allocation) on the overall model accuracy. First, the optimization problem is formulated as a minimization problem, where the objective function is defined as follows:

\begin{equation}
\begin{aligned}
& min\sum_{k=1}^{\#layers}w_k*b_k \\
\textrm{s.t.} \quad & accuracy_{M}-accuracy_{QM} <= C  \\
\end{aligned}
\end{equation}
where $w_{k}$ and $b_{k}$ are the weights and parameter bitwidths of layer $k$ respectively. $M$ is the model before quantization, $QM$ is the quantized model, and $C$ is the upper bound on the accuracy degradation that defines the performance penalty. When weight is quantized, the quantized weight is thought of as the unquantized weight added to some "noise" term. The noise on the weights is related to the noise on the last feature map, which is in turn related to the adversarial noise. Here the adversarial noise refers to the minimum noise that causes miss-classification and is calculated based on the input dataset. Using this fact, it is shown that if the weights of layer $k$ are quantized (i.e. if a quantization noise is added to a layer $k$), the noise through the layers undergoes an almost linear transformation to result with a noise on the last feature map is obtained, in turn causing degradation in model accuracy, $C$. Given this, AQ utilized an accuracy measurement, $m_{k}$, at layer $k$. The proposed measurement of accuracy is shown to be linear and additive which results in rewriting the optimization problem as follows.
\begin{equation}
\begin{aligned}
& min\sum_{k=1}^{\#layers}w_k*b_k \\
\textrm{s.t.} \quad & m_{total}=\sum_{k=1}^{\#layers}m_{k} <= C  \\
\end{aligned}
\end{equation}
where $m_{total}$ is the accuracy measurement across all layers. Also, the total accuracy measurement is shown to be represented as follows.
\begin{equation}
    m_{total}=\sum_{k=1}^{\#layers}\frac{p_{k}}{t_k}*e^{-\alpha*b_k}
\end{equation}
where $p_k$ is some parameter deduced from the mean of the noise on the last feature map, given a fixed $b_k$. $t_k$ is the ratio of the mean of the noise on the last feature map to the mean of the adversarial noise when the degradation in accuracy equals the upper bound, $C$. $\alpha=ln(4)$.
Given the new constrained optimization problem, and to calculate $t_k$, first the mean value of the adversarial noise is calculated from the given input dataset. Then, and with a fixed upper bound on accuracy degradation, $C$, the amount of noise on the weights for each layer is changed and added to the unquantized weights until the degradation in accuracy equals the upper bound, $C$. When the degradation accuracy and the upper bound are equal, the mean of the noise on the last feature map is calculated. The value of $t_k$ is then deduced. In order to calculate $p_k$, the bitwidth at each layer, $b_k$, is fixed and the mean of the noise of the last feature map is recorded, and then the value of $p_k$ is deduced. In order to calculate the optimal bitwidth at each layer, the bitwidth of the first layer is fixed and then $b_k$ can be deduced from $p_1$, $b_1$ and $w_1$. The choice of $b_1$ impacts the results of AQ as lower $b_1$ yields both a higher compression ratio and a higher accuracy degradation.

A set of experiments are conducted to prove the relationship between the proposed total measure of accuracy, its linearity, and its additivity. Moreover, to evaluate the performance of AQ: AlexNet, VGG-16, GoogLeNet, and ResNet-50 are quantized on ImageNet. For that purpose, only the convolutional layers are quantized while the fully connected layers are kept in 16-bit precision. The top-1\% accuracy is reported as a function of model size in 2D figures. AQ is compared to the fixed bitwidth optimization (without explicitly mentioning the fixed bitwidth is equal to which value) and to the optimization technique presented in SQNR framework \cite{lin2016fixed}.

The 2D figures reveal that AQ outperforms SQNR in terms of accuracy when the model size is fixed and achieves higher model size savings (higher compression ratio) for similar accuracy. In particular AQ used to quantize AlexNet, VGG-16, GoogLeNet and ResNet-50 on ImageNet achieves approximately 30\%, 40\%, 15\% and 20\% smaller model sizes respectively. To sum up, AQ follows a theoretical analysis method that relates the overall (total) model accuracy to quantization noise in order to find optimal bit precision for weights in DNNs.

\subsubsection{
Mixed-Precision Quantized Neural Networks with Progressively Decreasing Bitwidth (PDB) \protect\footnote{https://github.com/ariescts/mp-qnn}}
Progressively Decreasing Bitwidth \cite{chu2019mixed}, PDB, is a mixed-precision framework proposed to allocate per-layer weights precision based on a measure called input separability. In this work, the precision of activations is assumed to be fixed-point with a fixed bitwidth. The quantization adopted is training-aware, deterministic rounding.

By utilizing the internal feature distributions and network structure, PDB assigns weight precisions in layers heuristically based on the separability of feature representations. In particular, the feature separability of the dataset ${(x_k,y_k)}_{k=1}^{\#Layers}$ at layer $m$ is proposed as follows.
\begin{equation}
s^{(m)}=\frac{1}{\#Layers}\sum_{k=1}^{\#Layers}\frac{\frac{\sum_{l:y^{(l)}=y^{(k)}}d_{kl}^{(m)}}{\sum_{l}I(y^{(l)}=y^{(k)})}}{\frac{\sum_{l:y^{(l)}\neq y^{(k)}}d_{kl}^{(m)}}{\sum_{l}I(y^{(l)}\neq y^{(k)})}}   \label{pdb0}
\end{equation}
where $d_{kl}^{(m)}$ is the squared distance between the overall local representation and $I(.)$ is the indicator function. The feature separability aids as a hint to determine the weight precisions, and it is calculated by sampling a subset of the dataset. With deeper layers, the feature separability increases and hence the weights in these layers can be assigned lower precision. The heuristic used to determine the bitwidth, $b$, at any layer is as follows.
\begin{equation}
 b= \begin{cases} 
      4*h & s\leq 0.8 \\
      2*h & 0.8\textless s\leq 0.85 \\
      1*h & s\geq 0.85 
   \end{cases}
   \label{pdb}
\end{equation}
where $s$ is the feature separability and $h$ is a scaling integer used to adapt to different performance and memory requirements.

PDB is evaluated with ResNet-20 and VGG-7 on CIFAR-10 and CIFAR-100. Moroever, it is evaluated on ImageNet with AlexNet and ResNet-18 and on Pascal VOC with SSD-300 and ResNet-50 with a feature pyramid network based on faster R-CNN. In all cases, the precision of activations is fixed to a certain bitwidth. Also, all weights are quantized with progressively decreasing bitwidth according to the heuristic in Eqn. \ref{pdb} or set manually with a hint from Eqn. \ref{pdb0} except the weights of the output layer in the experiments on CIFAR-10\/100 and the weights of the input and output layers in the experiments on ImageNet. In all experiments, training is done from scratch, and the top-1\% and top-5\% accuracies are reported along with the average bitwidth of the parameters. The baselines used in these experiments are the associated floating-point full-precision networks, DoReFa, PACT, XNOR, Bi-Real, BNN and/or HWGQ  \cite{zhou2016dorefa,choi2018pact,rastegari2016xnor,liu2018bi,courbariaux2016binarized,cai2017deep}. It is worth noting that none of the reported baselines apply mixed-precision; all (except the full precision networks) adopt a homogeneous fixed precision for weights and/or activations.

In all experiments, PDB achieves an accuracy slightly less than that of the full precision models (except the experiment with VGG-7 on CIFAR-10 where PDB achieves an even higher accuracy than the full precision model) and higher than all the other baselines. For example with AlexNet on ImageNet, PDB's top-1\% accuracy drops by only 3.42\% compared to the full precision model. In some cases, the heuristic proposed in this framework achieves a better accuracy, while in others, PDB's manual assignment of precisions yields better accuracies. The model size, while not reported, is smaller than that of all the other baselines. In conclusion, PDB is a mixed-precision framework that devises a heuristic based on feature separability in order to determine the mixed-precision allocation of weights across different DNN layers. It achieves comparable results to the full precision models, with considerable model size savings when tested on ImageNet and Cifar-10\/100. Even with a sophisticated task like object detection and while having a notable drop in accuracy compared to the full precision models, the proposed framework still outperforms other frameworks that fix the precision across all layers like DoReFa in terms of both accuracy and model size savings.

\subsubsection{HAWQ-V2: Hessian Aware trace-Weighted Quantization of Neural Networks (HAWQ-V2) }

\cite{dong2019hawqv2} proposes HAWQ-V2, in which the problem of mixed-precision quantization (for weights and activations) is formulated as a second-order technique based on the per-block (block is a group of layers) Hessian spectrum. In particular, the HAWQ-V2 framework relies on the average Hessian trace as a measure of sensitivity, then uses the Pareto-frontier method to automatically determine the bit precision of different layers. Moreover, a fast algorithm based on Hutchinson’s algorithm \cite{avron2011randomized} in PyTorch is devised to compute the Hessian trace information. The quantization technique is deterministic rounding, under the category of retraining (it performs quantization-aware fine-tuning). The rounding used maps all real values in an interval to a quantized value, according to the bitwidths used in quantization.

The motivation behind HAWQ-V2 is that good sensitivity metrics can be obtained from the second-order information based on the Hessian matrix. Sensitivity is important to assess as it gives insights about which layer to quantize less/more aggressively. Technically, sensitivity is measured from the average Hessian trace. The average Hessian trace is the average of all Hessian eigenvalues. Since there is no direct access to the elements of the Hessian matrix, computing the Hessian trace may appear as a difficult task. Also, it is not feasible computationally to form the Hessian matrix explicitly. As such, the authors rely on the matrix-free Hutchinson algorithm \cite{avron2011randomized} to estimate the Hessian trace as follows. 
\begin{equation}
    Trace_{estimated}(H)=\frac{1}{n}\sum_{j=1}^n z_j^THz_j
    \label{hessian}
\end{equation}
where $H\in R^{d \times d}$ is the symmetric Hessian matrix, $z \in R^d$ is a random vector whose component is independent and identically distributed sampled Normal distribution ($\sim$ N(0,1)), and $n$ is the number of layers. In order to quantize activations with mixed-precision, the same theoretical formulations hold except that the Hessian is with respect to activations instead of model parameters. In fact, Eqn. \ref{hessian} still holds with the following modification. 
\begin{equation}
    z^TH_{a_j}z=\frac{1}{N}\sum_{i=1}^Nz_i^TH_{a_{j(x_i)}}z_i
\end{equation}
where $a_j$ is the activation of the j-th layer, $N$ is the number of inputs, and $\{z_i\}$ are the components of $z$ with respect to the i-th input, $x_i$. In short, HAWQ-V2 calculates the Hessian trace for the layer’s activations for one input at a time, and then averages the resulting Hessian traces of each block diagonal part. Authors claim that this trace computation converges quickly, averaging over the entire dataset is not required.

In order to automatically allocate the precision, HAWQ-V2 utilizes a Pareto frontier approach. Specifically, each candidate bit-precision setting $\in B$ (the set of all possible bitwidths) is sorted based on the total second-order perturbation that is caused, based on the following metric.

\begin{equation}
    \Gamma= \sum_{i=1}^L \Gamma_i=\sum_{i=1}^L Trace_{average}(H_i)*||Q(W_i)-W_i||_2^2
\end{equation}
where $L$ is the number of layers, $Trace_{average}(H_i)$ denotes the average Hessian trace at the $i$-th layer, and $||Q(Wi) − Wi||_2$ is the L2 norm of quantization perturbation. 
$\{W_1, W_2, · · · , W_L\}$ are learnable parameters, and $Q(.)$ is the quantization function that maps floating point values to low precision values. The idea behind this is that a bitwidth setting with minimal second-order perturbation to the model must yield good generalization after quantization-aware fine-tuning. Hence, given some target model size (the constraint), elements in $B$ are sorted depending on the $\Gamma$ values, and the bitwidth setting with minimal $\Gamma$ is chosen. After quantization, HAWQ-V2 performs a "quantization-aware" fine-tuning.

We note that HAWQ-V2 is a successor of HAWQ \cite{dong2019hawq}, where HAWQ is characterized by the following: 1) it utilizes a heuristic metric based on top Hessian eigenvalue as a measure of sensitivity, ignoring the remaining Hessian spectrum, 2) it only computes the relative sensitivity of different layers, and still requires a manual selection of the mixed-precision setting, and 3) it does not take into consideration mixed-precision activation quantization.

To evaluate HAWQ-V2, the authors quantized Inception-V3, ResNet-50, and SqueezeNext on ImageNet. As a baseline, the model with full precision for weights and activations is used. Moreover, the authors present a comparison with other frameworks in literature \cite{zhou2016dorefa,choi2018pact,zhang2018lq,han2015deep,wang2019haq,dong2019hawq, jacob2018quantization, park2018value, li2019fully}. Also, HAWQ-V2 is used to quantize RetinaNet-ResNet50 on Microsoft COCO 2017 \cite{lin2017focal, lin2014microsoft}. The top-1\% accuracy along with the weight compression ratio, and the model size are presented as performance metrics. For object detection tasks, mAP and activation compression ratio are also reported. In addition, the authors provide an ablation study of HAWQ-V3 to 1) show the importance of choosing the bit-precision setting that results in the smallest model perturbation, 2) measure the significance of utilizing the Hessian trace to weight the sensitivity, and 3) compare HAWQ-V2 with HAWQ approach where a sensitivity that is weighted by Top-1 Hessian eigenvalue is utilized.

The quantized Inception-V3 model tested on ImageNet achieves a x12.04 weight compression ratio and a 1.47\% loss in accuracy compared to the full precision model. Compared to \cite{jacob2018quantization}, \cite{dong2019hawq}, and \cite{park2018value}, HAWQ-V2 outperforms them all in terms of weight compression (weight compression in HAWQ is equivalent to HAWQ-V2) and accuracy. The results of quantizing ResNet-50 on image net reveal that HAWQ-V2 has a x12.24 compression ratio compared to the full precision model with a slight loss in accuracy. Compared to \cite{zhou2016dorefa,choi2018pact,zhang2018lq,han2015deep, dong2019hawq} and \cite{wang2019haq}, HAWQ-V2 achieves better accuracy. Moreover, the proposed framework outperforms the recent work \cite{wang2019haq} in terms of model compression as well. Furthermore, when HAWQ-V2 is used to quantize SqueezeNext on ImageNet. In this experiment, the activations are left in fixed 8-bit precision across all blocks (i.e. layers as well). Compared to the full precision model, the accuracy loss reported by HAWQ-V2 is 0.7\%, but with a x9.4 savings in model size (with an unprecedented model size of 1.07MB). On Microsoft COCO 2017, mixed-precision HAWQ-V2 has an mAP of 34.4, which exceeds that of HAWQ \cite{dong2019hawq} and FQN \cite{li2019fully}, with a slightly lower activation compression ratio. To sum up, HAWQ-V2 relies on two main steps to quantize pre-trained networks: First it utilizes second-order information (average Hessian trace in particular, calculated by matrix-free algorithms) in order to compute sensitivity, hence determining the relative precision order. Then, it utilizes a Pareto-frontier approach to automatically select the bit-precision. Moreover, mixed-precision activation is developed. The results of HAWQ-V2 are promising as it achieves good overall accuracy compared to the full precision model and some state-of-the-art models while providing high model size savings.

\subsubsection{Constructing energy-efficient mixed-precision neural networks through principal component analysis for edge intelligence (Hybrid-Net)}
\cite{chakraborty2020constructing} propose Hybrid-Net, is a mixed-precision framework that starts from a binary neural network (skeleton network used is proposed by \cite{rastegari2016xnor}) then relies on principle component analysis (PCA) to determine the significant layers (i.e. layers that contribute relevant transformations) in the network, and accordingly assign the parameters (weights and activations) in these layers higher precision. The quantization technique is post-training, deterministic rounding, and the quantization algorithm is inspired from XNOR-Net \cite{rastegari2016xnor}.

The core idea is that instead of using PCA to reduce dimensionality, Hybrid-Net starts from a low precision network (binary precision) and uses PCA to increase the precision of important layers in the DNN. As such, only significant layers are assigned higher precisions, while all other layers are kept in binary. To identify the significant layers that contribute relevant transformations on the input data, a DNN is viewed as an iterative projection of the input onto a successively higher-dimensional manifold at each layer, until the data can be separated linearly. A layer deemed significant is that in which the "significant dimensions" have increased from the preceding layer with a certain margin $\Delta$. The "significant dimensions" in a layer are identified by the number of filters needed to explain a cumulative 99\% of the total variance of the output activation map generated by that layer, and they are found by following the PCA algorithm proposed by \cite{garg2019low}. In particular, PCA is performed on the output tensor of each layer (before the nonlinear activation) resulting from convoluting the binarized input with the weight filters. PCA provides the ability to study the directions of maximum variance in the input data, and as such, it is used to identify redundancy. Based on the redundancy data obtained from the PCA, Hybrid-Net defines a significance metric that determines the significant layers whose weights and activations will be assigned $k_{b1}$ and $k_{b2}$ precisions respectively ($(k_{b1},k_{b2})\in \{2,4\}$).

To test Hybrid-Net: ResNet-\{20,32\} and VGG-15 are quantized on CIFAR-100, and ResNet-18 is quantized on ImageNet. In all of these experiments, the weights and activations of the first and last layers of the DNN are kept in full precision. The experiments performed were simulations for various random initializations, and the variation in the classification accuracy due to these initializations varies within a range of 2\%. for all cases considered. For ResNet models and as proposed by \cite{liu2018bi}, an additional design feature is included (adding identity shortcut connections at each layer instead of every two layers), and this additional feature enhances input representations through residual connections. However, in order to isolate the effect of the convolution layers on the activations, PCA on ResNets is performed on the plain version of a binary network without any residual connections. In all experiments, the top-1\% accuracy is reported. In addition, energy efficiency and memory compression are reported with respect to a full-precision network and normalized with respect to an XNOR-Net network. The baselines used for comparison are the full precision networks, XNOR-Net, Binary-Shortcut 1(same as XNOR-Net, except that it has residual connections in every layer), fixed 2-bit precision, XNOR-2x (modified XNOR-Net), Hybrid binary network, Bi-Real Net, LQ-Nets, PACT, DoReFa and/or Hybrid-Comp A $(k_{b1},k_{b2})$ (where the N-layer network is split into $N$-−$k$ binary precision (for weights and activations) layers and $k$ $(k_{b1},k_{b2})$ layers) \cite{rastegari2016xnor,choi2018pact,zhou2016dorefa,prabhu2018hybrid,liu2018bi,zhang2018lq}.

For the experiments on CIFAR-100 and with ResNet-\{20,32\} , Hybrid-Net(4,4) (4-bit precision for weights and activations in significant layers) enhances the classification accuracy by up to 13\% and 11\% respectively with up to 47\% degradation in memory compression when compared to XNOR-Net. Hybrid-Net(4,4) outperforms the accuracy reported by XNOR-Net on VGG-15 by 9.08\% with a 36\% degradation in energy efficiency compared to XNOR-Net. The experiments on ImageNet reveal that Hybrid-Net(2,2) achieves a 62.73\% accuracy outperforming XNOR-Net, Bi-Real Net, Binary-Shortcut 1, DoReFa, Hybrid-Comp A(2,2), and Hybrid binary network with a 20\% degradation in memory compression to XNOR-Net. To conclude, Hybrid-Net is a one-shot MXPDNN framework that starts with a binary network and allocates higher bit precisions for activations and weights in significant layers. Hybrid-Net outperforms the accuracy of extremely quantized neural networks, while still having comparable energy efficiency and memory compression.

\subsubsection{ZeroQ: A Novel Zero Shot Quantization Framework (ZeroQ) \protect\footnote{https://github.com/amirgholami/ZeroQ}}
ZeroQ, proposed in \cite{cai2020zeroq}, is a framework that optimizes, based on a proposed loss function and the batch normalization layers of the full precision model, an engineered distilled dataset to enable per-layer bitwidth allocation for weights and activations via Pareto Frontier without a need to access the training or validation data. The quantization technique is post-training, deterministic rounding.

Post-training quantization techniques usually suffer from the following: 1) performance degradation, 2) the need to access unlabeled data, and 3) the focus on standard DNNs. Motivated by those challenges, the authors propose ZeroQ as a solution. ZeroQ starts with a pre-trained full precision model, from which it creates synthetic data known as distilled data. There are two main challenges of not accessing the training data: 1) how to determine the activations' value range in order to do the clipping and 2) how to perform sensitivity analysis (see below) necessary to apply mixed-precision. ZeroQ proposes a novel method to distill data and tackle those challenges. To come up with an engineered distilled dataset that matches the statistics of the original training dataset, the authors rely on solving a distillation optimization problem (minimizing a loss). By doing so, a distribution of input data that matches the statistics of the batch normalization layer of the model is obtained. In particular, the distilled data is updated according to the collected statistics from the layers of the full precision model until the loss of the optimization problem is minimized. After the distilled data is available, it is used in order to compute the sensitivity metric based on the Kullback-Leibler (KL) divergence between the full precision model and its quantized counterpart (which is a function of the bitwidth). The sensitivity metric is used in order to avoid the exponentially large search space of bitwidths allocation. When the sensitivity metric at a certain layer with an "m"-bit precision is large, this means that quantizing this layer with "m"-bit precision will result in an output significantly different from the output of the full precision model, and hence this "m"-bit precision is not suitable for this layer but rather a higher precision is needed. This metric gives a relative ordering on the bitwidths, but to precisely set the precision, ZeroQ relies on the Pareto Frontier. In particular, for a target quantized model size and for each precision configuration, the computed sensitivity metrics (defined earlier) are summed across layers to measure the total model sensitivity that maintains the target quantized model size. The precision configuration that results in the minimum total model sensitivity is used. ZeroQ further utilizes dynamic programming in order to solve multiple such optimization problems for different target model sizes in parallel. In practice, given a certain model size constraint, the Pareto Frontier technique can be used to extract the best bit precision setting.

For evaluation, several experiments are done. ZeroQ is used to quantize ResNet-\{18,50,152\}, MobileNet-V2, SqueezeNext, Inception-V3 and ShuffleNet on ImageNet. Moreover, ResNet-20 is quantized on CIFAR-10, and RetinaNet is quantized on Microsoft COCO. For the baseline, the full precision model is used. ZeroQ is also compared with different quantization frameworks in literature \cite{choi2018pact,jacob2018quantization,park2018value,choukroun2019low,zhao2019improving,nagel2019data,li2019fully,haroush2020knowledge}. The model size, as well as top-1\% accuracy, are reported.

The results reported by quantizing ResNet-50 on ImageNet reveal that ZeroQ with weights quantized with mixed-precision and activations quantized with fixed bitwidth (6 bits) across all layers achieves x5.34 model size saving compared to the baseline with a slight loss in accuracy. ZeroQ with fixed 8-bit precision for weights and activations also outperforms \cite{choukroun2019low,choi2018pact} and \cite{zhao2019improving} in terms of accuracy and model size compression. With MobileNet-V2 on ImageNet, again the fixed 8-bit precision model of ZeroQ outperformed \cite{jacob2018quantization,nagel2019data} and \cite{park2017weighted} in terms of model size compression and accuracy. Moreover, this ZeroQ model has an accuracy drop of less than 0.12\% relative to the full precision baseline. With mixed-precision for weights and fixed 6-bit precision, the model has a 72.85\% accuracy while the full precision baseline has a 73.03\% accuracy, with ZeroQ having up to x5.35 model size savings.
For the object detection tasks, ZeroQ quantized RetinaNet on Microsoft COCO. Note that in RetinaNet, not all convolution layers are followed by batch normalization layers, but ZeroQ is still able to capture the distilled data. The ZeroQ model with mixed-precision for weights and fixed 6-bit precision for activations has a 0.5\% loss in accuracy compared to the full precision baseline, along with up to x6 model size savings. Compared to FQN \cite{li2019fully}, this ZeroQ model has 3.4\% higher accuracy with only x1.32 times larger model size. We note that in all experiments, the ZeroQ fixed 8-bit precision models for weights and activations achieve the highest accuracy (and closest to full precision models) among all ZeroQ reported precision configurations. When ZeroQ is used to quantize ResNet-20 on CIFAR-10, the mixed-precision weights/fixed 8-bit precision activations configuration experiences only a 0.16\% drop in accuracy with x5.20 savings in model size. In summary, ZeroQ is a new post-training framework that is able to generate distilled data statistically similar to the training data without access to the latter. It is also able to do mixed-precision quantization by utilizing the Pareto Frontier optimization formulation. The framework is thoroughly tested on different datasets and models, with a notable test on an object detection dataset. The time needed to execute ZeroQ is low, and it achieves competitive accuracy results while providing good model size savings compared to the state-of-the-art techniques reported. 

\subsection{HAWQ-V3: Dyadic Neural Network Quantization \protect\footnote{https://github.com/zhen-dong/hawq.git}}
In \cite{yao2021hawq}, the authors propose HAWQ-V3, a hardware-aware fixed low-precision and MXPDNN framework (for weights and activations) with integer-only inference by solving a constrained Integer Linear Programming (ILP) formulation. It relies on uniform quantization (with rounding). Also, channel-wise symmetric quantization is used for weights, layer-wise asymmetric quantization is utilized for activations, and static quantization for all the scaling factors. Results are presented with/without distillation, so this falls in the retraining/post-training category respectively.

HAWQ-V3 provides two major enhancements compared to HAWQ-V2. In particular, 1) it solves the problem of having the batch norm layers and residual connections kept in floating point precision in HAWQ-V2, whereby it uses integer-only multiplication, addition, and bit shifting with static quantization. In fact, HAWQ-V3 does not perform floating point nor integer division calculations throughout the whole inference framework. In addition, 2) it formulates the mixed precision problem as an ILP with latency, model size, and/or Bit Operations (BOPS) taken as constraints hence it is more flexible (towards hardware requirements) and hardware-aware than HAWQ-V2. HAWQ-V3 is open source and extends TVM \cite{chen2018tvm} by supporting 4-bit uniform/mixed-precision quantization in order to deploy the integer-only framework on hardware.  We will next elaborate on the two enhancements mentioned above.

Most of the existing quantization algorithms rely on "simulated" or "fake" quantization, where the parameters are stored with quantization. For inference, these parameters are cast into floating point, which incurs a hidden cost of conversion between floating point and quantized integer values. Moreover, there is a limitation on the peak computational capacity of the hardware since FP32 ALUs span a larger die area. To mitigate these issues, HAWQ-V3 quantizes the 1) matrix multiplication (convolution), 2) batch norm layers, and 3) residual connections as integer-only. In particular, for a layer with hidden activations, $a_h$, and weight tensor, $W$, followed by ReLU activation:

\begin{enumerate}
    \item $a_h$ is quantized to $S_{a_h}*q_{a_h}$, and $W$ is quantized to $S_W*q_W$ where $S$ and $q$ denote the real-valued quantization scales and corresponding quantized integer values respectively. The output result, $o$ is then computed as $o=S_W*S_{a_h}(q_W*q_{a_h})$, $q_W*q_{a_h}$ being the matrix multiplication (convolution) calculated with integer in low precision and accumulated in INT32 precision. Then this output is quantized into $q_o$ before being sent to the following layer as follows.
    \begin{equation}
        q_o=INT(\frac{S_W*S_{a_h}}{S_o}*(q_W*q_{a_h}))
        \label{hawq-v31}
    \end{equation}
    where $S_o$ is the scale factor pre-calculated for output activation. $\frac{S_W*S_{a_h}}{S_o}$ is a floating point scaling that is to be multiplied by the INT32 accumulated result. This can be achieved by making the scaling a dyadic number (for both low-precision and mixed-precision). Hence, the dyadic scaling in Eqn. \ref{hawq-v31} can be carried out by relying on INT32 multiplication and bit shifting. Moreover, relying on dyadic numbers eradicates the need for supporting division.
    
    \item Given an input activation $a_i$, the batch normalization is applied as follows
    \begin{equation}
        BatchNorm(a_i)=\gamma*\frac{a_i-mean(a_i)}{std(a_i)}+\lambda
    \end{equation}
    where $\gamma$ and $\lambda$ are trainable parameters. Since $\gamma$, $\lambda$, $mean(.)$, and $std(.)$ are fixed during inference, they are fused with the convolution and quantized with an integer-only approach. Moreover, HAWQ-V3 first keeps the convolution and batch normalization layer unfolded and allows the batch statistics to update. After several epochs, the running statistics in the batch normalization layer are frozen and the convolution and batch normalization layers are folded.
    
    \item HAWQ-V3 uses INT32 for the residual branch, whereby it ensures that the addition operation is done via dyadic arithmetic.
\end{enumerate}

HAWQ-V3 formulates the mixed precision bitwidth allocation problem for weights and activations as a constrained ILP problem. First, assuming that the perturbations for each layer are independent of each other, HAWQ-V3 precomputes the sensitivity of each layer separately. The sensitivity metric adopted is the Hessian-based perturbation proposed in HAWQ-V2. The ILP problem is formulated to optimize bitwidth allocation that minimizes this sensitivity, as follows.
\begin{equation}
    \begin{aligned}
\min_{{\{b_j\}}_{j=1}^N} \quad & \sum_{j=1}^{N}{\Gamma_j^{(b_j)}},\\
\textrm{s.t.} \quad & \sum_{j=1}^{N}{M_j^{(b_j)}} \leq A1, 
\quad & \sum_{j=1}^{N}{B_j^{(b_j)}} \leq A2, \text{and }
\quad & \sum_{j=1}^{N}{L_j^{(b_j)}} \leq A3
\end{aligned}
\end{equation}
where $\Gamma_j^{(b_j)}$ is the $j$-th layer perturbation with $b_j$ bit quantization, $M_j^{(b_j)}$ is the model size, $B_j^{(b_j)}$ is the BOPS required for computing the $j$-th layer, $L_j^{(b_j)}$ is the associated latency. $A1$, $A2$, and $A3$ are upper limits on model size, BOPS, and latency respectively. We note that HAWQ considers the hardware-specific latency constraint, which is important
since a layer’s latency does not necessarily decrease by half when the model quantized from INT8 to INT4 precision. This means that given a latency constraint, it is better (accuracy-wise) to keep some layers at high precision, even if they have low sensitivity. 
HAWQ-V3 does not require all the constraints to be set simultaneously. In fact, HAWQ-V3 is evaluated with one constraint set at a time (see below). The ILP problem is solved via the open source PULP library \cite{mitchell2011pulp}, which yields fast result for all cases that HAWQ-V3 is tested on (solution obtained in less than one second).

We mentioned that HAWQ-V3 extends the TVM \cite{chen2018tvm} framework to support INT4 inference with and without mixed precision. In particular, new features are added in both graph-level IR and operator schedules to render INT4 inference efficient. Eight 4-bit elements are packed into an INT32 data type and memory movements are performed in chunks. In the code generation stage, the data type and all memory accesses are adjusted as INT32. Moreover, HAWQ-V3 adopts scheduling strategies like Cutlass \cite{hawqv31}, whereby a new direct convolution schedule is implemented in TVM for Tensor Cores to support both 8-bit and 4-bit data. The TVM configuration knobs are set to the thread size, block size, and loop ordering. This enables the auto-tuner in TVM to search for the optimal latency settings. The back-end target consists of NVIDIA Turing Tensor Cores of T4 GPU.

HAWQ-V3 is evaluated by quantizing ResNet-\{18/50\} and Inception-V3 for ImageNet using INT4 (fixed across all layers), INT8 (fixed across all layers), and mixed-precision INT4/8 (mixed-precision with 4 and 8 bitwidths) with/without distillation, with/without ILP constraints. Experiments with distillation apply ResNet-101 \cite{he2016deep} as the teacher, and the quantized model as the student, where the naive distillation method \cite{hinton2015distilling} is utilized. In all experiments, the model size, BOPS, latency, and top-1\% accuracy are reported, and the first and last layer are kept in 8-bit integer. For experiments with constraints, the speeds compared to INT8 HAWQ-V3 are reported as well. Moreover, for all experiments, batch normalization folding is performed to expedite the inference. All the inference calculations are done by relying on dyadic arithmetic (i.e. integer addition, multiplication, and bit shifting), without any floating point or integer division usage. Mixed-precision HAWQ-V3 is compared to the baseline FP32 models, RVQuant \cite{park2018value}, PACT \cite{choi2018pact}, LQ-Nets \cite{zhang2018lq}, CalibTIB \cite{hubara2020improving}, HAQ \cite{wang2019haq}, Integer Only \cite{jacob2018quantization}, OneBitWidth \cite{chin2020one}, INT4 HAWQ-V3 (wights and activations quantized with 4 bitwidth), and INT8 HAWQ-V3 (wights and activations quantized with 8 bitwidth).

We will focus on the results reported for mixed precision. Without ILP constraints, all the mixed-precision results with distillation enhance the accuracy for all models while increasing the model size slightly when compared to all other frameworks, except the baseline (FP32) and the INT-8 HAWQ-V3. For ResNet-50, mixed-precision HAWQ-V3 with distillation has an accuracy boost of 1.34\% compared to the case without distillation. On ResNet-18 and ResNet-50, mixed precision HAWQ-V3 with distillation incurs a 1.09\% and 0.99\% decrease in top-1\% accuracy compared to the FP32 models, while yielding 6.66x and 5.23x model size savings, respectively.

For the experiments with one constraint set at a time (all results reported herein are with distillation): With model size limit set to 7.9MB on ResNet-18, mixed precision HAWQ-V3 achieves 71.09\% accuracy which is very close to that of INT8 HAWQ-V3. The mixed-precision HAWQ-V3 is also 6\% faster than INT8 HAWQ-V3. When ILP os requested to fine the bit-precision setting that results in 19\% faster latency on ResNet-18 as compared to the INT8 model, the resulting HAWQ-V3 model yields 70.55\% accuracy with a model size of 7.2MB. When the BOP limit is set to 197G on ResNet-50, the reported accuracy is 76.76\% on HAWQ-V3 with a model size of 22MB. In summary, HAWQ-V3 is a fixed/mixed-precision integer-only quantization framework, where inference is performed by relying on only integer multiplication, addition, and bit shifts. For the mixed-precision formulation, a hardware-aware ILP based method is proposed and its solution finds the optimal trade-off between model perturbation and constraints such as model size, inference speed, and total BOPS. In addition, TVM is extended to support INT4/8 with/without mixed precision. Its reported results are competitive with the state-of-the-art MXPDNNs and fixed-precision quantization frameworks.

\subsubsection{OPQ: Compressing Deep Neural Networks with One-shot Pruning-Quantization (OPQ)}
OPQ \cite{hu2021opq} is a layer-wise mixed-precision+pruning framework that utilizes a pre-trained model and performs a One-shot (O) pruning (P) quantization (Q), i.e. it analytically solves the compression allocation at the same time. Only weight parameters are pruned and quantized, while activations are left in full floating-point precision. The quantization scheme used is categorized as retraining, deterministic, and rounding.

We herein summarize OPQ's optimization process. Starting from a pre-trained full precision model, a target number of bits is required to store unpruned weights (for all layers), and a target pruning rate: OPQ first computes the pruning masks for all layers. This is done by reformulating the problem of finding which weights could be removed to find the pruning ratios of all layers. To minimize the errors caused by weight pruning while having a pruned model that is close in accuracy to the original unpruned model, the following pruning objective function is used.
\begin{equation}
    b_1^*,...,b_L^*=\argmin_{b_1,b_2,...,b_L}\frac{1}{\#weights}\sum_{j=1}^{L}E_j^b
    \label{pe1}
\end{equation}
where $b_1^*,...,b_L^*$ represent the range limits where the weights will be removed around zero in each layer. For layer 1, the weights would be removed in the symmetric range  $[-b_1^*, b_1^*]$ around zero. $E_j^b$ is the pruning error at layer $j$. Now given a target pruning rate, the objective function in Eqn. \ref{pe1} is solved through the Lagrange multiplier. By relying on the Newton-Raphson method, the Lagrange multiplier $\lambda$ is found, then the optimal range limits are derived analytically as follows.
\begin{equation}
    b_1^*=b_2^*=...=b_L^*=\sqrt{\lambda}
\end{equation}
The pruning ratio can then be derived as follows.
\begin{equation}
    R_{p_{j}}=\int_{-b_j^*}^{b_j^*}N_j*p_j(x)dx
\end{equation}
where $p_j(.)$ is a probability density function at layer $j$ and $N_j$ is the number of weight parameters in the weight tensor $W_j$ of the $jth$ layer. After that, the pruning masks, $M_j \in \{0,1\}^{|W_j|}$ which are used to remove the unimportant weights from a layer $j$ by $M_j \circ W_j$ ($\circ$ being the Hadamard product) are derived via magnitude-based thresholding. After computing the pruning masks, OPQ computes the quantization steps. For that, all channels of a layer are enforced to have a common scale factor and offset (i.e. common codebook) because having channel-wise codebooks would otherwise introduce overheads. That way all channels in a layer have one quantization step between two adjacent quantized bins/centers, and the channels in a certain layer $j$ share a common quantization function, $Q_j$. The mean-squared errors due to quantization are formulated as follows.
\begin{equation}
    E_q=\sum_{j=1}^L \frac{1}{N_{unpruned_{j}}} \sum_{i=1}^{N_j} M_{ji}(W_{ji}-Q_j(W_{ji}))^2
    \label{opq1}
\end{equation}
where $N_{unpruned_{j}}$ is the number of all unpruned weights in layer $j$, and $W_{ji}$ is the $ith$ element of the weight tensor at layer $j$. Now given a target number of bits, $B,$ required to store unpruned weights for all layers, Eqn. \ref{opq1} is minimized by a Lagrange multiplier, $\lambda_2$. When the Lagrange derivation is applied, the quantization step at layer $j$ is computed as follows.
\begin{equation}
    \Delta_j=\frac{1}{2^{B-1}*N_{unpruned}}\sum_{j=1}^L \sum_{i=1}^{C_j}N_{unpruned_{ji}}\alpha_{ji}
\end{equation}
where $N_{unpruned}$ represents the number of all unpruned weights in the model, $C_j$ is the number of channels in $jth$ layer, and $\alpha$ is a positive real value that represents the maximum for the $ith$ channel.
These steps are then used to quantize the DNN in the forward pass. It is worth noting that during fine-tuning, the compression module (comprising the pruning mask and the channel-wise quantizer of each layer) is fixed and only weight parameters are updated in order to alleviate the loss in accuracy due to the error in the model compression. Moreover and in the backward pass, the STE \cite{bengio2013estimating} is utilized on the quantized weights to compute the gradients. In particular, a two-stage fine-tuning is applied. In the first stage, the pruned networks are fine-tuned (the weights only) without quantization till they retrieve the performance of the original unpruned models. In the second stage, pruning and quantization are simultaneously applied while fine-tuning the weights.

OPQ is evaluated by compressing AlexNet, VGG-16, ResNet-50, and MobileNet-V1 on ImageNet. OPQ is compared with six pruning methods, five quantization methods- Q-CNN, Binary-Weight-Networks, ReNorm, HAQ, ACIQ \cite{wu2016quantized,rastegari2016xnor,he2018learning,wang2019haq,banner2018post}- and three pruning-quantization techniques: Deep Compression, CLIP-Q, and/or ANNC \cite{han2015deep,tung2018deep,yang2020automatic}. We will not focus on the comparison with the pruning-only baseline frameworks. In all experiments, the top-\{1,5\}\% accuracies, \% prune rate, average bitwidth, and model size savings.

Experiments with AlexNet reveal show that OPQ achieves 57.09\% top-1\% accuracy (0.46\% higher than the full precision uncompressed model) and model size savings of x138.96, outperforming all quantization-only baselines (excluding ReNorm which was not reported for this experiment) and pruning-quantization baselines. With VGG-16, OPQ achieves the highest accuracy compared to Q-CNN, Deep Compression, and CLIP-Q with a pruning rate of 94.41\% and x195.87 savings in model size. With MobileNet-V1, OPQ achieves higher top-1\% accuracy than HAQ, ReNorm, Deep Compression, CLIP-Q, and ANNC. Moreover, it enhances the accuracy of the uncompressed full precision model by 0.55\%. With ResNet-50, OPQ compressed model saves x38.03 in model size with a high 76.41\% accuracy outperforming HAQ, Deep Compression, ACIQ, and CLIP-Q. In conclusion, OPQ is a One-shot pruning-quantization framework that suggests that a pre-trained model is enough for determining the compression module prior to fine-tuning, and no complex/iterative optimization is required during the fine-tuning process. In addition, it outperforms many other compression frameworks in terms of accuracy and model size savings.

\subsubsection{Towards Mixed-Precision Quantization of Neural Networks via Constrained Optimization (MPQNNCO)}
The authors in \cite{chen2021towards} formulate the mixed-precision problem (for weights and activations) as a discrete constrained optimization problem which they solved by relying on second-order Taylor expansion, Hessian matrix computations, and a greedy search algorithm. In particular, the optimization problem is reformulated as a Multiple Knapsack problem which is solved by the proposed greedy search algorithm efficiently. This framework relies on deterministic rounding quantization for weights and activations (where the step size is found via solving an optimization problem). Since the framework starts with a pre-trained network and relies on fine-tuning after quantization, the technique is categorized as retraining.

This framework finds a middle ground between search-based techniques that rely on a small number of evaluations to reduce computational complexity and other methods that rely on some criteria that are easy to compute in order to reduce the time cost of performance evaluation. In particular, the proposed framework first formulates the MXPDNN allocation problem across the layers as a discrete optimization problem constrained by model compression. Solving the original optimization problem is computationally expensive as the network needs to be evaluated on the whole training dataset for each precision assignment. As such the authors approximate the objective function by relying on second-order Taylor expansion. The approximation consists of a constant zero-term, a first-order gradient, and a second-order Hessian matrix. The zero-term is omitted (the constant does not affect the optimization solution), and the first-order gradient is also omitted with the assumption that the pre-trained model will converge to a local minimum with a nearly zero gradient vector. As such, only the Hessian matrix is used as the objective function which approximates the loss perturbation from quantization. Solving this approximation at this stage, however; still suffers from large computational overhead as the Hessian matrix has a complexity quadratic to the number of parameters. To calculate the loss perturbation incurred due to quantization of a specific precision assignment efficiently, the Hessian matrix is further approximated by further omitting the term that represents the computational cost bottleneck. In addition to the computation becoming more efficient, there is also no need to traverse the complete dataset for calculating the loss perturbation since the loss converges rapidly as the number of images increases. 

To render the precision assignment automatic, the quantization across different layers is assumed to be independent, and the optimization problem is reformulated as follows.

\begin{equation}
\begin{aligned}
    \min_{\{b^{(l)}\}_{l=1}^N} &\quad \frac{1}{2}\sum_{l=1}^N(\Delta w^{(l)})^T H_w^{(l)} \Delta w^{(l)} \\
    \textrm{s.t.} & \quad \Delta w^{(l)}=Q(w^{(l)},b^{(l)})-w^{(l)} \; \text{and}
    \; \sum_{l=1}^{N}|w^{(l)}|*b^{(l)} \leq b_{t}*\sum_{l=1}^{N}|w^{(l)}| \\
     \text{where} & \quad  b^{(l)} \in B  \;\& \; l \in \{1,...,N\}
    \end{aligned}
    \label{opti}
\end{equation}
where $w=\{w^{(l)}\}_{l=1}^N$ is the set of flattened weight tensors of a CNN of $N$ layers, $b^{(l)}$ denotes the precision assignment at layer $l$, $Q(.)$ is the quantization function, and $b_{t}$ is the target average precision of the network. Moreover, $H_w^{(l)}$ denotes the approximated Hessian matrix, $B$ is the set of candidate bit-widths of each layer, and $|.|$ denotes the length of the corresponding vector. To solve this problem, it is finally written as a special variant of the Knapsack problem called the Multiple-Choice Knapsack Problem (MCKP). An "item" in MCKP terms is the bitwidth assignment of each layer in the newly reformulated problem. Since MCKP is NP-hard, a greedy search algorithm is proposed to solve it in an efficient manner. To summarize the greedy search algorithm, the \textbf{dominated} items (these are not considered in the solution of the MCKP) are filtered and each layer is initialized with the minimum available bitwidth. Then, the layer with the highest priority (based on the proposed greedy criterion) is chosen and its bitwidth is incremented until the target compression constraint is dissatisfied.

MPQNNCO is evaluated by starting with pre-trained models and quantizing (and fine-tuning) ResNet-\{18,50\} and MobileNet-V2 on ImageNet. The framework is compared with other fixed-precision frameworks (LQ-Nets \cite{zhang2018lq}, ABC-Net \cite{lin2017towards}, DoReFa \cite{zhou2016dorefa}, and PACT \cite{choi2018pact}) and MXPDNN frameworks (AutoQ \cite{lou2019autoq}, HAWQ \cite{dong2019hawq}, HAWQ-V2 \cite{dong2019hawqv2}, DC \cite{han2015deep}, and HAQ \cite{wang2019haq}). In all experiments, the top-1\% accuracies and drops in these accuracies (relative to full precision models) are reported, along with the weight and activation compression ratios.

The experiments on ResNet-18 show that MPQNNCO incurs a lower accuracy drop (relative to the full precision model) compared to LQ-Nets, ABC-Net, DoReFa, and PACT while achieving higher compression ratios (weight and activation) when both weights and activations are quantized with mixed-precision. For ResNet-50, the proposed framework has the same accuracy drop ($0.85\%$) compared to HAQ with larger compression ratios (weight and activation). Moreover, and compared to HAWQ and HAWQ-V2, MPQNNCO achieves a lower accuracy drop and similar compression ratios (weight and activation). The experiments on MobileNet-V2 reveal that the proposed framework has a lower drop in accuracy, similar weight compression ratio and higher activation compression ratio.

To sum up, this framework presents an interpretable, principled technique to tackle the mixed-precision problem. Eventually, a greedy search algorithm is used to solve the optimization problem which was reformulated as a MCKP relying on an approximation of the Hessian matrix. The experiments used to evaluate the proposed framework reveal its competitive advantage over other SOTA MXPDNN and fixed-precision frameworks.

\subsection{Meta-Heuristic-Based Optimization}
\subsubsection{Mixed-Precision Quantization of ConvNets via Differentiable Neural Architecture Search (DNAS)}
In order to allocate different bitwidths of weights and activations to different layers, \cite{wu2018mixed}  formulated the problem as a neural architecture search (NAS) problem with gradient-based optimization. In particular, an efficient differentiable NAS (DNAS) framework is proposed in order to efficiently navigate the search space while optimizing. The quantization technique used rounds a continuous value into its nearest neighbor, which means it is deterministic. Moreover, the quantization technique is retraining.

We start by describing the proposed DNAS framework (without quantization). To avoid the complexity imposed by NAS's exhaustive search, DNAS starts by constructing a directed acyclic graph, known as "supernet", to represent the search space. The supernet has a macro architecture that coincides with the target network. In the supernet, nodes and edges represent data tensors and operators respectively, where the operators are parametrized by their weights. To compute the data tensor at a specific node, the outputs which are computed by the operators of all the incoming edges are summed. As a result, the supernet can be used to represent any NN, which is in that case a subgraph of the supernet. Next, DNAS changes the supernet into a fully differentiable stochastic super net; where each edge is parametrized by $\theta$ such that the probability of executing an edge between nodes \textit{i} and \textit{j} is denoted $P_{\theta^{i,j}}$. From the distribution \textit{$P_{\theta}$}, a subgraph is deduced and the expected loss of the stochastic supernet is computed. Since the expected loss is not differentiable with respect to $\theta$, the Gumbel-Softmax method \cite{jang2016categorical} is used to estimate the gradient. Now instead of solving for the optimal architecture (like in ordinary NAS), DNAS relaxes the combinatorial optimization problem to solve for the optimal $\theta$ that minimizes the expected loss by Stochastic Gradient Descend (SGD). Given the optimal parameter $\theta$, the optimal architecture is directly obtained by sampling from \textit{$P_{\theta}$}. Throughout the training, sample architectures are drawn from \textit{$P_{\theta}$} and the dataset for architecture search is divided into two parts, one used to train weights and another used to train the $\theta$. The training process of the two parts on the drawn architectures is done in an alternating manner to ensure good generalization. 

The proposed DNAS framework is then used to tackle mixed-precision quantization. The goal is to allocate per-layer bitwidths for weights and activations. For that, the loss is redefined to include both the cross-entropy and a cost term. The cost term can take one of two forms, one with a goal to compress model size, and the other to reduce the computational complexity. The second cost term form (that to reduce computational complexity) is a function of bitwidths of activations and weights, while the first is a function of weights' bitwidths only. Now that the loss is redefined, the same framework discussed earlier is used for finding optimal bitwidths. After these are found, the authors quantize weights/activations by rounding them into the nearest neighbor according to the optimal bitwidths found inspired by \cite{zhou2016dorefa}.

For evaluation, two sets of experiments are conducted. The first set aims to quantize ResNet-\{20,56,110\} \cite{he2016deep} on CIFAR-10 \cite{krizhevsky2009learning}. In these experiments, the first and last layers are kept in full precision. Moreover, only weights are quantized (i.e. the first cost term form is used), and the quantization is carried on the block level; where a block has one or more layers. The second set of experiments comprises quantizing "ReLU-only preactivation" ResNet-\{18,34\} \cite{he2016identity} on ImageNet \cite{deng2009imagenet}. Similar to the first set of experiments, the first and last layers are not quantized, and the mixed-precision is conducted at the block level. Here the two cost terms forms are utilized in separate experiments, i.e. in some experiments only weights are quantized to reduce model size while in others both weights and activations are quantized to reduce computational complexity. In all experiments, accuracy and compression ratio (model size savings in comparison to 1) the equivalent benchmark but in full precision and 2) TTQ \cite{zhu2016trained}, ADMM \cite{leng2018extremely}, PACT \cite{choi2018pact}, DoReFa \cite{zhou2016dorefa}, QIP \cite{jung2018joint} and/or GroupNet \cite{zhuang2018training}) are adopted as measures of performance.

The first set of experiments on CIFAR-10, using the proposed framework, reveals that all of these experiments with the most accurate subgraphs outperform their full precision counterparts as well as the work done in \cite{zhu2016trained}. In addition, the most accurate models of the proposed framework achieve up to x12.5 compression in model size. On another note, the authors report the results of their "most-efficient" models (these have the highest compression ratio), and the results show that these models achieve even higher savings in model size while their accuracy drops to less than 0.39\%. The second set of experiments were on ImageNet. These experiments show that the proposed framework (with its most accurate models) achieves either superior accuracy or very close accuracy compared to the full precision counterparts and other works previously reported in the literature \cite{zhu2016trained,leng2018extremely,choi2018pact,zhou2016dorefa,jung2018joint,zhuang2018training}, while maintaining a high compression rate (i.e. low model size/computations). Accordingly, the authors contribute to a novel and efficient differentiable neural architecture search that can be applied to the problem of per-layer mixed-precision while still being generic to be applied on other applications. The proposed framework achieves up to x103.9 computational complexity savings and up to x21.1 model size savings while maintaining a competitive accuracy.

\subsubsection{Joint Neural Architecture Search and Optimization (JASQ)}
JASQ \cite{chen2018joint} is a framework that performs joint neural architecture search and mixed-precision quantization policy search (for weights) by relying on a multi-objective evolutionary search algorithm. Rounding is used as the quantization scheme so it is categorized as deterministic. Moreover, since no fine-tuning is performed on the quantized network, the scheme is categorized as post-training.

JASQ is an efficient framework that jointly searches for neural networks and quantization policies, by relying on a flexible multi-objective function that can be adjusted to produce models with different model sizes and accuracies. In particular, the search problem is defined as follows.
\begin{equation}
    \max_{\Theta}F(\Theta)=\max_{\Theta} A(\Theta)*(\frac{S(\Theta)}{S_t})^\Gamma
    \label{jasq}
\end{equation}
where $\Theta$ is the quantized model constructed by its network architecture and quantization policy and $F(\Theta)$ is the fitness function. $A(\Theta)$ and $S(\Theta)$ denote the validation accuracy and the model size of the quantized model respectively. Moreover, $S_t$ is the target size of the model and $\Gamma=0$ if $S(\Theta)\leq S_t$, or it is set to $-1$ otherwise. This multi-objective function reduces into a single objective function (relying on accuracy) when the model size meets the target model size. Since the quantized model depends on the network architecture and the quantization policy, the search task is to find an optimal neural architecture and an optimal quantization policy. For the neural architecture search, JASQ adopts the NASNet space \cite{zoph2018learning}. In particular, this is a cell-wise search space consisting of normal cells and reduction cells. Normal cells take in a feature map as an input and return another feature map of the same dimension, while reduction cells return a feature map with half the height and width. Each cell is a directed acyclic graph that has combinations where each combination is specified by two inputs and two operations (a combination takes two inputs and applies an operation to each of them). During the search, only the structure of these cells is altered and the resulting architecture is eventually determined by these cells' structures, the first convolution channels and the cell stacking number. As for the quantization policy, the search aims to optimize the bitwidth of each cell. Tournament selection \cite{goldberg1991comparative}, a classical evolutionary algorithm, is utilized for JASQ. The population of the evolutionary algorithm consists of models that are first initialized randomly. First, each model of the population (each individual) is trained on the training set, quantized using a quantization policy, and then evaluated on the validation set. The fitness $F(\Theta)$ is then computed as in Eqn. \ref{jasq}. During each evolutionary step, $N$ random models are sampled from the population; the best individual (according to fitness) becomes the parent while the worst individual (according to fitness) gets omitted from the population. The model that becomes a parent is then mutated to produce a child which is in turn trained, quantized, evaluated to compute fitness, and then pushed into the population. The mutation is applied to both the network architecture (by randomly choosing one of the 2 inputs and 2 operations from each combination and replacing what is chosen with a random substitute) and the quantization policy (by randomly picking one bitwidth assignment of one cell and replacing it by a random bitwidth). This search scheme allows random samples to repeatedly compete in iterations. We note that JASQ uses parameter sharing technique \cite{pham2018efficient} whereby the set of parameters is shared among all individuals of the population for acceleration purposes.

JASQ is evaluated on two sets of experiments. In the first set, the architecture is fixed, i.e. JASQ is applied on existing networks to find the quantization policy. In particular, JASQ is used to quantize ResNet-\{18,34,50,101,152\}, DenseNet-\{121,169,201\} \cite{huang2017densely}, MobileNet-\{V1,V2\}, and SqueezeNet on ImageNet. For this set of experiments, the top-1\% accuracy is reported along with the size of the quantized models, and the quantized models by JASQ are compared to 8-bit fixed precision models and the floating point models. For the second set of experiments, the joint search is applied on CIFAR-10 and ImageNet. The first and last layers of ImageNet (in both sets of experiments) are not quantized. $S_t$ is tweaked to obtain two models: JASQNet and JASQNet-Small. For this set of experiments, JASQ is compared to PNASNET-5 \cite{liu2018progressive}, NASNet-\{A,B,C\} \cite{zoph2018learning}, AmoebaNet-B \cite{real2019regularized}, ENAS \cite{pham2018efficient}, and DARTS \cite{liu2018darts}. Moreover, the floating point JASQNet and JASQNet models are included for comparison. For these experiments, the search cost, \#parameters in millions, size  in MB, and \%error (compared to floating point models) are reported.

For the first set of experiments and for all models, JASQ achieves better accuracy compared to floating point models and 8-bit fixed precision models while having competitive model size savings with the 8-bit fixed precision models. ResNet-18, JASQ achieves a 70.02\% accuracy higher than the floating point model (by 0.26\%) with a smaller size. For DenseNet-169, JASQ achieves an accuracy 0.79\% higher than the floating point model with 4.76x model size savings. For MobileNet-\{V1,V2\}, and SqueezeNet JASQ achieves higher accuracy than the 8-bit fixed precision models but has slightly less model size savings.  For the second set of experiments, JASQNet-Small has a model size up to 10x smaller than other frameworks on CIFAR-10 and up to 8x smaller than other frameworks on ImageNet. Compared to the floating point JASQNet and JASQNet-Small, the quantized JASQNet and JASQNet-Small achieve higher accuracies and smaller model sizes. Compared to the other frameworks that search for architecture, the quantized JASQNet and JASQNet-Small achieve similar \%error with the lowest reported model sizes (other frameworks do not quantize networks).

In conclusion, JASQ is an evolutionary-based search technique that jointly optimizes the network architecture and the mixed-precision quantization policy. JASQ can also be applied on existing models (only optimizes the quantization policy in that case). The results reported show that JASQ models (ones obtained by quantizing fixed architectures and ones obtained by joint search) achieve high accuracies and small model sizes. 

\subsubsection{Mixed Precision Neural Architecture Search for Energy Efficient Deep Learning (MPNASEE)}
\cite{pan2019} proposed MPNASEE; a framework that relies on energy-constrained differential NAS with a proxy task to co-optimize the architecture and block-wise precision of the weights and activations and efficiently maneuver the search space. The quantization is categorized as training-aware, deterministic, and rounding.

Since the goal is to jointly find an optimal energy-efficient architecture and to find optimal precision allocation across layers, we will henceforth discuss how the proposed framework achieves these two goals. To find the optimal energy-efficient architecture, MPNASEE can be viewed as a controller that interacts with two environments: the task and the hardware. The objective of the controller is to find architectures that minimize the loss associated with the task and satisfy energy constraints simultaneously. The training objective function is formulated as follows.
\begin{equation}
\begin{aligned}
\min_{\theta} \quad & \EX_c[Loss(M_{w_{1}}(c);D_1)]
\\
\textrm{s.t.} \quad & w_{1}=arg \min_{w}Loss(M_w(c),D_2,c)\\
  &\EX_{c}[Energy(c)]<A   \\
\end{aligned}
\end{equation}
where $\theta$ are internal parameters, and $M_{w}(c)$ is a DNN model with weights $w$ given a network configuration $c$ generated by the NAS controller. $D_1$ and $D_2$ are the validation and training sets respectively. $A$ is an energy constraint. In brief, during training, the controller finds $\theta$ that yield architectures with the minimum expected task-related loss evaluated on a validation set; such that the network weights $w_1$ are obtained by minimizing the loss function on a training set, and that the expected energy is constrained to guarantee an architecture search that is energy efficient. To obtain hardware performance metrics (energy, latency) throughout training, BitFusion \cite{sharma2018bit} simulator is utilized. In order to solve the objective function, the problem is relaxed into an unconstrained objective, $w_1$ is approximated via one-step gradient descent motivated by DARTS \cite{liu2018darts}, and the REINFORCE algorithm \cite{williams1992simple} is used to deal with the non-differentiable energy terms. In order to achieve the second goal (finding optimal precision allocation across layers), the search space is made up of a hierarchical stack of the base search unit: MobileNet-V2. By utilizing a skip operation (that skips an entire MobileNet-V2 block when selected) in the search space, the number of MobileNet-V2 units can be searched. Each MobileNet-V2 can choose the quantization precision adaptively. For each of the two types of layers in a MobileNet-V2 block (depthwise and bottleneck layers), a precision is chosen from a set of four possible bitwidths. Accordingly, each MobileNet-V2 block in the architecture will have two precisions, one for the depthwise layers and the other for bottleneck layers. We note that the authors rely on the Gumbel-Softmax distribution \cite{jang2016categorical} to replace the non-differentiable categorical network configuration samples with continuously differentiable samples. This in turn facilitates the use of vanilla back-propagation. To jointly search the architectures and mixed quantization precision for each of the layers in the DNN, a proxy task is adopted where the search is carried out on tiny ImageNet; which consists of 200 classes each of which contains 500 training images, 50 validation images, and 50 test images.

To evaluate the proposed framework, two discovered architectures are chosen (referred to as theirs-small and theirs-base henceforth) and trained from scratch on CIFAR-100 and ImageNet. The top-1\% and top-5\% classification errors on the validation set, energy, latency, and model size are reported in all experiments. The baseline models used are fixed 8-bit precision VGG-16, ResNet-50, MobileNet-V2 (as in HAQ), FBNet-B, fixed 3-bit precision FBNet-B, HAQ-small (HAQ framework applied on their small architecture), HAQ-base (HAQ framework applied on their base architecture), DenseNet-BC-190+Mixup, ENAS+Cutout, and/or NAO+Cutout \cite{simonyan2014very,he2016deep,zhang2017mixup,pham2018efficient,wang2019haq,wu2019fbnet,inproceedings}.

The reported results on ImageNet and CIFAR-100 reveal that MPNASEE achieves very high energy savings for both of the discovered architectures (theirs-small and theirs-base) while maintaining a low error in accuracy; competitive with the baselines. In particular, MPNASEE achieves lower accuracy errors and consumes less energy and latency compared to HAQ for both datasets and with both of the discovered architectures. On CIFAR-100, and compared to the fixed 8- bit precision MobileNet-V2 quantized model, theirs-small achieves a 5x reduction in model size, more than 5x reduction in energy consumption, and higher accuracy. In conclusion, MPNASEE is an MXPDNN framework that aims to co-optimize the architecture and quantization policy, while satisfying some energy constraints. The results of MPNASEE suggest that it is a very energy-efficient framework able to compete with advanced mixed and fixed quantization frameworks accuracy-wise.

\subsubsection{APQ: Joint Search for Network Architecture, Pruning and Quantization Policy (APQ) \protect\footnote{https://github.com/mit-han-lab/apq}}
APQ \cite{wang2020apq} is a framework that jointly optimizes the DNN architecture, pruning policy, and mixed-precision policy (for weights and activations) by relying on a pre-trained once-for-all network that generates DNN architectures and a quantization-aware predictor that demolishes the quantized data collection time. The quantization technique used by APQ is retraining, deterministic, and rounding.

To efficiently implement deep learning on a target hardware with some resource constraints, works in literature have proposed techniques to optimize the model architecture and other techniques to optimize model compression (i.e. pruning and quantization). The sequential optimization of these stages (architecture, pruning, quantization) has been shown to be efficient \cite{han2019design,han2015deep}, but poses a challenge with hyper-parameter tuning as the number of hyper-parameters grows exponentially \cite{he2018amc}. To mitigate the sub-optimality resulting due to optimizing each one of these stages separately, APQ reorganizes the pipeline of the three stages: architecture, pruning, and quantization into two stages: architecture search (coarse-grain architecture search and fine-grain channel search for channel-wise pruning) and mixed-precision search. The joint optimization objective of these two stages is formulated as follows.
\begin{equation}
    F^* = \argmax_{(F,w,P,Q)} ACC_{val}(Q(P(F,w)))
\end{equation}
where $ACC_{val}$ is a function that searches for the DNN architecture, $F$, with the best validation accuracy on the dataset, $Q$ is the mixed-precision quantization function, and $P$ is the function that prunes the channels. To solve this objective function,
first (1) a flexible once-for-all network that supports an extremely large and fine-grained search space is trained - via Progressive Shrinking (PS) algorithm \cite{cai2019once} - to support both operator change and a number of channel changes (PS is viewed as a generalized pruning method which shrinks multiple dimensions; depth, width, kernel size, and resolution of the full network) , so as any sub-network can be extracted from it and its accuracy can be approximately evaluated directly without re-training. MobileNet-V2 is used as the backbone block that builds the once-for-all network. Then, (2) a quantization-aware predictor is built to predict the accuracy of models with different architectures and different bitwidths after quantization without having to fine-tune the models (which is usually time and resource-consuming). The predictor built is a 3-layer feed-forward DNN. Since collecting data to train this predictor is prohibitively time-consuming, a predictor-transfer technique is proposed: First, a full-precision accuracy predictor is trained on cheap data points collected by evaluating the once-for-all network, and then the quantization-aware predictor is transferred from the full-precision predictor. The input to this quantization-aware predictor is the encoded network architecture, pruning strategy, and quantization policy. The network architecture is encoded block by block, where each block is the concatenation of the encoded one-hot vectors for kernel size, channel numbers, and weight/activation bitwidths. The quantization policy is also encoded as one-hot vectors. Having trained the quantization-aware predictor, the search is then deemed ultra-fast, and this predictor is used to estimate the predicted accuracy instead of the measured accuracy. Moreover, this predictor can be utilized for new hardware platforms and deployment scenarios without training the model again. After training the quantization-aware predictor, (3) a latency/energy lookup table is constructed to do a resource-constrained evolution search \cite{guo2020single} (whose evaluation process is replaced by the quantization-aware predictor) since evaluating each candidate policy on an actual hardware platform is costly. BitFusion \cite{sharma2018bit} is used to measure the resource consumption for the mixed-precision model and fill the values in the lookup table. First, a lookup table consisting of the latency and energy of each layer under different DNN architectures and bitwidths is constructed. Then for each candidate policy, it is broken down, and the lookup table can be queried at small cost to find the accurate inference cost. As such, the resource constraints can be verified by using the lookup table.

APQ is evaluated on ImageNet. The baselines used are one or a combination of fixed 8/6/4-bit precision MobileNet-V2, fixed bit precision ResNet-\{34/18\}, ProxylessNAS, AMC, HAQ, searched models by DNAS, and/or models searched by Single Path One-Shot \cite{cai2018proxylessnas,he2018amc,wang2019haq,wu2018mixed,guo2020single}. For the experiments, different architectures found by APQ (two with transfer and one without transfer) are compared with the baselines, and the top-1\% accuracy, latency, energy, Giga BitOps, design cost, marginal $CO_{2}$ emission, and marginal cloud compute cost are reported. Other experiments vary the latency/energy/BitOps constraint (each considered separately) and compare the accuracy of the models found by APQ to the accuracies of the other baselines.

The reported results show that one model found by APQ (whose architecture description is not shown or reported) with transfer outperforms all the other baselines in terms of accuracy (in particular it has a 0.5\% higher accuracy than the best performing baseline in term of accuracy: Single Path One-Shot), another model with transfer has the lowest latency, and a third model without transfer has the lowest BitOps. Moreover, the marginal carbon dioxide emission and the marginal cloud compute cost of the reported APQ models are way less than those reported for the baselines. Moreover, with decreasing the energy (latency) constraint, the accuracy of APQ's model decreases but remains in all cases higher than the top-1\% accuracy of HAQ with MobileNet-V2 and the fixed 6-bit (4-bit) precision MobileNet-V2 models. In particular, for the tightest latency/energy constraint, APQ's model achieves a higher top-1\% accuracy (+10.5\%/+11.3\%) compared to the MobileNet-V2 baseline. For the experiment that varies the BitOps, even with a tight BitOps constraint, APQ's model improves the top-1\% accuracy by more than 2\% compared with the searched model using Single Path One-Shot. In conclusion, APQ is a framework that jointly optimizes architecture, pruning, and mixed quantization by starting from a pre-trained once-for-all network and relying on a quantization-aware predictor along with a lookup table to conduct a budget-constrained evolutionary search efficiently. The reported results for APQ searched models are competitive with other top-performing baselines.

\subsubsection{Search What You Want: Barrier Panelty NAS for Mixed Precision Quantization (BP-NAS)}
\cite{yu2020search} propose BP-NAS, an MXPDNN framework that utilizes a differentiable soft barrier penalty-based NAS in order to automatically find the optimal per-block (all layers in one block share the same bitwidth) precision for weights and activations given a complexity constraint. The main idea behind adding the soft barrier to the loss term is to penalize the mixed-precision models that are near the complexity constraint barrier, hence increasing the chance of obtaining a mixed-precision model that satisfies the constraint and with good accuracy. Moreover, the soft barrier penalty focuses the search process on the valid search space, hence deeming the search more efficient. The quantization used is deterministic rounding. Moreover, the distribution reshaping method is used in training a floating-point model whose weights are uniformly distributed and that has no long-tailed activations, and this trained model is used as the pre-trained mixed-precision model, and fine-tuning is performed to enhance the accuracy after sampling the mixed-precision model. Accordingly, the quantization is categorized as retraining.

In BP-NAS, the mixed-precision quantization is treated as a constrained optimization task where the constraint imposed is the average BOPs. In particular and from a set of predefined configurations, BP-NAS seeks to find the optimal weight and activation precisions for all layers. The optimization problem is formulated as follows.
\begin{equation}
    C^*=\argmin_{C} L_{val}(N,C) \quad BOPs_{av}\leq A
    \label{bpnas}
\end{equation}
where $C^*$ is the optimal mixed-precision configuration, $N$ is the given network to quantize, and $L_{val}$ if the loss on the validation set. $A$ is the average BOPs constraint.
To solve Eqn. \ref{bpnas}, BP-NAS adopts the weight-sharing NAS also known as DARTS \cite{liu2018darts}. Accordingly, a supernet is constructed and trained. The edges of the supernet correspond to the layers in the mixed-precision model. To limit the search on the feasible search space (within the defined constraint) during training, the constraint in Eqn. \ref{bpnas} is rewritten as the barrier penalty regularizer: a differentiable regularization term inspired by the interior method \cite{alizadeh1995interior} as follows.
\begin{equation}
    L^*_{BP}(\theta)=-\mu log(log(A+1-E(supernet)))
\end{equation}
where $\theta$ is the architecture parameter and $E(supernet)$ is the expected BOPs cost of the sampled model which incorporates the importance factors for all layers. To ensure that these importance factors are highly selective; one element only approaches one while the other approach 0, BP-NAS uses a Prob-1 regularizer $L_{Prob-1}$. While training the supernet, the supernet weights $w$ and architecture parameter $\theta$ are trained alternately. In particular, the gradient of loss functions $L1$ on a training set and $L2$ on validation set update the weights and the architecture parameter of the supernet respectively. The loss functions $L1$ and $L2$ are represented as follows.
\begin{align}
\begin{split}
    L1=L_{training-set}(w,\theta) \\
    L2=L_{validation-set}(w,\theta)+L^*_{BP}(\theta)+L_{Prob-1}
\end{split}
\end{align}
After training the supernet for some epochs, the mixed-precision model is sampled and then fine-tuned (retrained) for some epochs to retrieve the accuracy.

BP-NAS is tested on image classification and object detection tasks under different BOPs constraints. In particular, ResNet-20 and ResNet-50 are quantized on CIFAR-10 and ImageNet respectively, and ImageNet-50 (faster R-CNN) is quantized on Microsoft COCO. In all experiments, the block-wise configurations for the quantized models are provided, as well as the top-1\% accuracy, the model size compression ratio, the bit operations compression ratio, and the average bit operations of the mixed-precision models. For the object detection task, the mean average precision (mAP) is reported. The baselines used for comparison are the full precision floating point models, models found by DNAS \cite{wu2018mixed}, HAWQ \cite{dong2019hawq}, DoReFa \cite{zhou2016dorefa}, PACT \cite{choi2018pact}, LQ-Nets \cite{zhang2018lq}, some fixed/mixed chosen baselines, FQN \cite{li2019fully}, and/or HAQ \cite{wang2019haq}. In addition, an ablation study is carried out to demonstrate the statistical significance of the proposed mixed-precision search, and the effectiveness of the proposed Prob-1 regularization term.

The results reported on CIFAR-10 show that with a maximum BOPs constraint of four, the top-1\% accuracy dropped by 0.31\% with a 10.19x compression in model size compared to the full precision baseline. BP-NAS with 4-bit BOPs constraint achieved a 76.67\% accuracy on ImageNet, outperforming PACT, HAWQ and HAQ accuracy-wise. On ImageNet and compared to the fixed four-bit quantized model, BP-NAS achieves 76.67\% Top-1 accuracy with similar average bit operations. For the object detection task and with BOPs constraint of four, BP-NAS achieves a 0.358 mAP (with a 1.5\% mAP reduction compared to the full precision baseline) and a 64.76x bit computation cost compression ratio  compared to the full precision model. To sum, BP-NAS is a gradient-based NAS framework for mixed-precision allocation with a predefined complexity constraint. It relies on a soft barrier penalty term added to the loss, and it achieves competitive results compared to the baselines.

\subsubsection{HMQ: Hardware Friendly Mixed Precision Quantization Block for CNNs (HMQ)}
HMQ \cite{habi2020hmq} is a mixed-precision quantization block that enables an efficient search for quantization parameters by utilizing the Gumbel-Softmax estimator \cite{jang2016categorical}. In this framework, an optimization method based on the HMQ blocks is used to search for the bitwidths (of weights and activations) and the threshold of each quantizer at the same time. Deterministic rounding is used for quantization after finding the optimal bitwidths. Since the optimization technique starts from a pre-trained full precision model then utilizes fine-tuning after quantization, the quantization technique is retraining.

The HMQ block is a hardware-friendly network block that uses SGD during training to learn a symmetric, uniform quantization scheme parametrized by a powers-of-two threshold and a bitwidth. To search for a pair of threshold and bitwidth, HMQ represents each pair (threshold, bitwidth) in the search space as a sample of a discrete categorical random variable of the Gumbel-Softmax. HMQs tie each pair of bit-width and threshold in their search space with a single trainable parameter. In particular and in feed-forward training, the pairs are sampled from the categorical distribution and used in the approximation of the joint discrete probability distribution of thresholds and bit-widths. Then, the expected threshold and step size, which rely on the aforementioned probability, are used to parametrize the quantizer of HMQ. After that in backpropagation, the gradients of the rounding operations are estimated using the STE to deal with the non-differentiability. We henceforth describe the optimization process that is based on the HMQ blocks. The search space of HMQ-based optimization is $TxB$ where bitwidths belong to $B$, and thresholds belong to $T$ which is defined as follows.
\begin{equation}
    T={2^k,2^{k-1},...,2^{k-8}}
\end{equation}
where $k=min{k:2^k\geq max(abs(I))}$ given that $I$ is the input tensor of weights to be quantized by some HMQ block. Furthermore, $B$ is predefined as a set of finite number of bitwidths in the search space of HMQs quantizing weights. First and starting from a pre-trained full precision floating point model, an HMQ block is added after every weight and every activation tensor per layer, including the first and last layers. After that, the fine-tuning optimization process can be seen as comprising two stages. The first stage is split into cycles with equal number of epochs each. In this stage, the model weights and HMQ parameters are trained (fine-tuned) jointly, where the model is optimized with respect to a target weight compression rate, $R_{wc}$, by relying on SGD to minimize the following loss function.
\begin{equation}
    Loss(\theta,\gamma)=J_{CrossEntropy}(\theta,\gamma)+\lambda(J_{R_{wc}}(\gamma_{w}))^2
\end{equation}
where $\theta$ is the set of weight tensors to be quantized, $\gamma$ is the set of HMQ parameters, and $\gamma_{w}$ is the subset of $\gamma$ consisting of all of the quantizing weight parameters. $J_{CrossEntropy}(\theta,\gamma)$ is the standard cross-entropy loss, $J_{R_{wc}}(\gamma_{w})$ is the loss with respect to the target compression rate $R_{wc}$ and $\lambda$ is a hyper-parameter controlling the trade-off between these two loss terms. Furthermore, $J_{R_{wc}}(\gamma_{w})$ is defined as follows.
\begin{equation}
J_{R_{wc}}(\gamma_{w})=\frac{max(0,(R_{wc}-R_{Ewc}(\gamma_{w})))}{R_{wc}}
\end{equation}
where $R_{Ewc}(\gamma_{w})$ is the expected weight compression rate induced by the values of $\gamma_{w}$. In this stage, the target weight compression rate $R_{wc}$ is gradually increased during the first few cycles, and the bitwidths across the whole model are gradually decreased in a mixed-precision fashion. For activations compression, the target activations compression rate is defined as follows.
\begin{equation}
    R_{ac}=\frac{32*max_{X_{j}}|X_j|}{A}
\end{equation}
where $A$ is an integer that represents a given size of memory, $X_j$ are the activation tensors to be quantized, $|X_j|$ are the number of elements in the activation tensors. This compression rate is increased gradually throughout the first few cycles of the first stage of the optimization process. But in order to quantize the activations, and instead of adding the target activations compression component to the loss, the precise bitwidths of every activation map are implied by $A$ and determined by the following equation. 
\begin{equation}
    bit(X)=floor(\frac{A}{|X|})
    \label{hmq0}
\end{equation}
where $X$ is the set of activation tensors to be quantized and $|X|$ is the number of elements in the activation tensor. Using Eqn. \ref{hmq0} to quantize the activation tensors ensures that the target activations compression rate is met and simplifies the loss function of the optimization process. Eventually, every activation tensor in the quantized model is bounded from above by a the value $A$. 
In the second stage of optimization, only the model weights are fine-tuned. In this stage, the quantizer of each HMQ block is parameterized by the pair (threshold, bitwidth) that corresponds to the maximal parameter drawn by the Gumbel-Softmax distribution and learned in the first stage.

To evaluate HMQ, ResNet-18 is quantized on CIFAR-10 and ResNet-50, EfficientNet-B0 \cite{tan2019efficientnet}, MobileNet-V1, and MobileNet-V2 are quantized on ImageNet. In all the experiments, the weight and activation HMQ blocks have a maximum bitwidth of eight. Moreover, the Weight Compression Rate (WCR, i.e. the ratio between the total number of bits of the weights in the full-precision model and the total number of bits of the weights in the quantized model), Activation Compression Rate (ACR, i.e. the ratio between the number of bits of the largest activation tensor in the full precision model and the number of bits of the largest activation tensor in the quantized model), and top-1\% accuracy are reported. ACRs and WCRs are reported with respect to the full precision model. For the experiments on CIFAR-10, $B = {1, 2, 3, 4, 5, 6, 7, 8}$, and in those on ImageNet $B = {2, 3, 4, 5, 6, 7, 8}$. The baselines models used for comparison are HAWQ, LQ-Nets, UNIQ, DNAS, DQ, HAQ, HAWQ-v2, and/or the full-precision models \cite{dong2019hawq,zhang2018lq,baskin2021uniq,wu2018mixed,uhlich2019mixed,wang2019haq,dong2019hawqv2}. But the comparison with the baselines is done when HMQs are used to quantize weights only.

The results reported on CIFAR-10 reveal that HMQ-based optimization outperforms LQ-Nets and UNIQ in terms of accuracy and with different WCRs and ACRs set. Quantizing MobileNet-V2 on ImageNet, with $ACR=8$, the HMQ-based framework outperformed DQ with a top-1\% accuracy of 70.9\% close to the full-precision model accuracy of 71.88\%. ResNet-50 experiments ($ACR=8$) reveal that HMQ-based optimization has a better accuracy than HAWQ, with a higher WCR of x13.1. With an $ACR=4$, the HMQ-based framework achieves a 74.6\% accuracy with a WCR of x11.97. Quantizing MobileNet-V1 on ImageNet, with activations given a fixed precision of 8 bits, the HMQ-based framework outperforms HAQ in terms of WCR and accuracy. With activation tensors having a single precision of 8 bits and on EfficientNet-B0, HMQs yield a 74.6\% accuracy with a WCR of x11.97. To sum, HMQs are hardware-friendly blocks proposed as a part of a two-stage fine-tuning gradient-based optimization that maintains competitive accuracy while quantizing the weight and activation tensors in a mixed-precision fashion.

\subsubsection{Rethinking Differentiable Search for Mixed-Precision Neural Networks (EdMIPS) \protect\footnote{https://github.com/zhaoweicai/EdMIPS}}
EdMIPS is proposed \cite{cai2020rethinking} as an efficient differentiable "mixed-precision network search" (MPS) that aims to find optimal per-layer precision of weights and activations without a proxy task, by solving a constrained optimization problem. The quantization technique is similar to that of HWGQ \cite{cai2017deep}. As such, the quantization technique is deterministic rounding. Since it trains from scratch, it is categorized as training-aware.

Even though EdMIPS is inspired by NAS, there are three main differences between regular neural architecture search and what is used in EdMIPS: MPS. 1) NAS usually depends on a proxy task in order to reduce the computational complexity of maneuvering the optimal network architecture search space, especially when the dataset is large. Such a proxy task is not relevant for MPS since the differences between layers in one network are significant and should be captured, and the optimal precision allocation between a small dataset and a large dataset is likely different. Moreover, 2) NAS usually relies on the minimization of classification loss to find the optimal architecture, but relying on that minimization will always lead to the solution whereby the highest bitwidth among the candidates is selected. This defies the reason behind using MPS. In addition, 3) general NAS uses different operators like convolution, skip connection, pooling and others. However, in MPS, only convolutions are needed. As such, EdMIPS aspires to resolve the three points above. To mitigate 1), it utilizes a differentiable search similar to what was done in \cite{liu2018darts}. To tackle 2) and 3), EdMIPS uses a complexity budget to constrain the minimization and an effective optimization where the architecture and network parameters are updated in a single forward-backward pass with efficient composite convolution. As such, the training in EdMIPS has a constant overhead independent of the search space size. Using a complexity budget to constrain minimization is denoted as "complexity-aware learning", during which the optimization problem of finding the optimal bitwidths allocation of weights and activations is rendered a constrained optimization problem. In "complexity-aware learning", the classification risk is minimized via minimizing Lagrangian such that there exists a constraint on the complexity risk. This formulation is generic and can accommodate any user-defined complexity be it memory, computation, energy, or others. In EdMIPS the complexity used is model complexity, characterized by the bit operations (BitOps) normalized to the floating-point operations (FLOPs) of the first layer. The minimization of the Lagrangian of the constrained formulation is not trivial since the search space is binary in nature; as EdMIPS seeks to find optimal weights and activations precision across the whole network. For a simpler optimization problem, binary search space is relaxed into a continuous one. This hence enables the following learning process for EdMIPS: the optimal bitwidths allocation and weight tensors are learned via gradient descent in continuous parameter space to minimize the Lagrangian. The weight tensors and architecture parameters are updated in a single forward-backward pass by default (another more computationally expensive update is also proposed where they are updated in an alternating fashion: 1st fix the tensors and update the parameters then fix the architecture parameters and update the tensors). This learning technique is a much less expensive procedure than combinatorial search. In addition, the composite convolution operation which is parametrized by the weighted sum of the weight tensor is used to replace multiple parallel convolutions, and weights are shared across filters to avoid under-training for filter with low probabilities. Eventually, after finding the optimal architecture parameters, EdMIPS needs to discretize the continuous selector variables into binary ones (as it has previously changed the binary search space into a continuous one to find the optimal parameters). For that, two strategies are proposed by the authors: "winner-take-all strategy" and "sampling strategy". In the first proposed strategy, the branch with the highest continuous selector is selected, which results in a single deterministic architecture. In the second strategy, the binary selectors are sampled from a multinomial distribution based on the continuous selectors, in which case a user can do multiple selections and hence sample multiple architectures. The choice of the best architecture in terms of complexity-accuracy trade-off is then manually made. The first strategy is adopted by EdMIPS by default.

EdMIPS is used to evaluate quantizing ResNet-\{18,50\}, AlexNet, GoogLeNet and Inception-V3 on ImageNet. EdMIPS is compared to HWGQ-Net \cite{cai2017deep} and LQ-Net \cite{zhang2018lq} that allocate a bitwidth of 2 for network parameters, as well to the full precision models. The optimal bitwidths found are reported along with the top-1-5\% accuracies and savings from composite convolution. In addition, ablation results are presented which ensure the correctness of the authors' choices to relax the optimization, use shared weights across the filter, adapt the multinomial sampling, utilize the single forward-backward pass optimization, and introducing efficient composite convolution.

For an average bitwidth of 2, EdMIPS outperforms the LQ-Net and HWGQ-Net in terms of top-1-5\% accuracies. Compared to the full precision models, EdMIPS performs decently, with a maximum of 4.9\% loss in top-1\% accuracy. With the help of efficient composite convolution, EdMIPS reported a maximum of around x3.85 savings in model size, x1.53 savings in memory, and x1.96 savings in computation. As such, EdMIPS presents itself as an efficient MPS framework that utilizes a differentiable search to find the optimal per-layer parameter bit precision.

\subsubsection{EvoQ: Mixed Precision Quantization of DNNs via Sensitivity Guided Evolutionary Search}
EvoQ \cite{evoq} is an MXPDNN framework (for weights) that relies on evolutionary search with a limited amount of data. The quantization scheme used is deterministic, uniform, asymmetric and categorized as post-training as no fine-tuning is utilized.

EvoQ starts with a pre-trained full precision model and tries to find the optimal mixed-precision quantization policy via evolutionary search, given some target bitwidth constraint. The evolutionary algorithm used is the classical tournament selection. The population of the algorithm is the quantization policies. We hereon summarize the used evolutionary algorithm. First, the population is initialized with a uniform quantization policy (all layers have the same precision) and its random perturbations. Then each individual (member of the population, i.e. a quantization policy) is evaluated using 50 unlabeled samples. To evaluate the quantization policy (this evaluation is needed as the fitness measure), the output difference is measured between the quantization model and the pre-trained full precision model using $N$ samples (data points) as follows.
\begin{equation}
    E(P(b_1...b_L))=\frac{1}{N}\sum_{i=1}^N(Q_{P(b_1...b_L)}(x_i)-M(x_i))^2
\end{equation}
where $P(b_1...b_L)$ is the quantization policy of the DNN with $L$ layers ($b_i$ being the bitwidth of the i-th layer), $Q_{P(b_1...b_L)}$ is the model quantized with policy P(.), M(.) is the pre-trained full precision model, and $x_i$ denotes the i-th input sample. Authors of EvoQ have set $N=50$ and this overcomes the time-consuming process of assessing the quantization model on the whole test dataset. The fitness of the quantization policy measured by EvoQ does not assume that the error of each layer is additive and independent and hence can work for low precision quantization. We note that the additivity assumption is not satisfied for low precision quantization (hence such an assumption hinders using low precision quantization) as the coupling effect of the layers cannot be ignored with low precision. After the policies are evaluated and at each evolutionary step, a number $K$ ($K$ controls the aggressiveness of the search) of quantization policies (individuals) are randomly sampled from the population, and that with the highest fitness measure is selected as the parent. Then a child (a new quantization policy) is constructed by mutation operation applied on the parent. This child is added to the population, while the policy with worst fitness measure out of the sampled individuals is excluded from the population. This scheme allows reaching an optimal quantization policy by allowing random individuals to repeatedly compete. It is important to note that EvoQ optimizes the mutation direction by utilizing the sensitivity of each layer. In particular, 50 samples are first used to evaluate the quantization error per-layer based on some bitwidth allocation, then the relative gain/loss per-layer is calculated as the quantization bitwidth is increased or decreased. Layers that are less sensitive to the current bitwidth will have the probability of having a lower bitwidth increased, which overcomes the problem of local minimum and hence renders the overall search efficient. In addition, EvoQ relies on the teacher-student framework to calibrate the features and hence improve the performance of the quantization model. In particular, the pre-trained full precision model is the teacher and the the quantized model is the student. The outputs and the intermediate features with more dimensions are utilized for calibration.

EvoQ is evaluated on image classification and object detection tasks, and in all experiments the batch normalization is folded into the adjacent layer prior to quantization. For image classification task, EvoQ is used to quantize ResNet-\{18,50,101\}, SqueezeNet, ShuffleNetV2, and MobileNet-v2 on ImageNet. EvoQ's performance is compared in that context to the following frameworks: UniQ (uniform quantization of 4 bits), OMSE \cite{hubara2016binarized}, and ACIQ \cite{real2019regularized}. For object detection, EvoQ is used to quantize SSD \cite{liu2016ssd} (trained with the training sets VOC 2007 and VOC 2012 and tested on VOC 2007 test set, and whose backbones are based on VGG-16 and ResNet-50) on PASCAL VOC. For object detection, EvoQ is compared to UniQ. For both types of tasks, 50 samples from the training set are used for the sensitivity analysis, quantization policy evaluation, activation range estimation, and feature calibration. Moreover, activations are quantized with fixed 8-bit precision across all layers. For the object detection task, the mean Average Precision (mAP) is used as a measure of performance. For classification task, the top-\{1,5\}\% are reported. For the two types of tasks, the size of the network is reported.

For image classification task, EvoQ outperforms OMSE, ACIQ, and UniQ in terms of accuracy and size for ResNet-\{18,50,101\}, SqueezeNet, ShuffleNet-V2, and MobileNet-V2 (with the exception of ResNet-101 on ACIQ). For ResNet-\{50,101\}, the drop in accuracy incurred by EvoQ compared to the full precision pre-trained model is less than 1\%. For object detection, EvoQ incurs a 0.74\% and a 1.14\% mAP degradation (when compared to the full precision models) for SSD with VGG-16 and SSD with ResNet-50 backbones respectively.

To sum up, EvoQ is a framework that tackles the problem of mixed-precision of weights by deploying evolutionary search with sensitivity analysis-guided mutation for faster search and feature calibration for better quantization model performance. EvoQ has competitive results compared to some frameworks used for object detection and image classification tasks.

\subsubsection{Bit-Mixer: Mixed-Precision networks with runtime bitwidth selection (Bit-Mixer)}
\cite{bulat2021bit} proposed Bit-Mixer, a framework for MXPDNN (for weights and activations) that allows -with high inference accuracy- at runtime modification in the per-layer precision by first training a meta-quantized network, optimized via a 3-stage optimization technique. Their quantization technique is based on that proposed in \cite{esser2019learned}, and hence it is in essence deterministic rounding. Since a pre-trained network is used, the quantization is categorized as retraining.

Motivated by the need for an MXPDNN framework that is compatible with multiple hardware platforms, and that can adjust the precision at runtime, Bit-Mixer trains a meta-quantized network (via the 3-stage optimization) which can change the network's precision (at any layer) at "test" time. To facilitate that, the authors propose learning a "Transitional Batch Normalization" (TBN) layer for each transition in precision that occurs between two consecutive layers to accommodate for the shift in the distribution that results when such a transition or change in precision occurs. Given two consecutive layers in which a transition of precision took place, the TBN layer in between ensures that the framework will converge with high accuracy. Two parameters are learned for the TBN layers, and these parameters only depend on the current layer's quantization level, so they are related to the bitwidth of this layer (not its preceding layer). The introduction of all the TBN layers in the network results in a meta-quantized network. In order to train the proposed meta-quantized network, first, a 2-stage efficient process is applied in order to allow the network to switch to a fixed precision configuration across all layers at runtime. After that, a third optimization stage is used in which the meta-network after stage two is trained gradually to allow mixed-precision transitions at run time. During stage 1, the weights of the networks are kept in full precision while the activations are quantized as follows: In each iteration, a bitwidth out of the pre-defined set of bitwidths is randomly selected with equal probability in order to quantize the activations across all layers. After this stage, the network will have quantized precision activations (with fixed bit precision across all layers), with the weights kept in full precision. During the second stage, the network obtained from the first stage is used as a starting network. Now both the weights and activations are quantized, where at the end the model obtained will have the same fixed precision for all weights and activations across all layers. During stage 3, again the network obtained from stage 2 is used as initialization to this stage. With a probability \textit{P}, the weights and activations are trained in the same manner as stage 2 (the same bitwidth is used for both weights and activations and across all layers), and with a probability \textit{1-P}, the bitwidth of each layer is selected randomly and independently from other layers so as at the end the network obtained will have different bit precision for different layers. As the training in this stage proceeds, \textit{P} is gradually decreased until $P=\frac{1}{4}$, after which typically the training continues for few epochs with this fixed probability. It is worth noting that the training scheduler is shared among the 3 stages.

For performance evaluation of Bit-Mixer, the framework is used to quantize ResNet-\{18,34,50\} and EBN \cite{bulat2020high} on ImageNet. EBN has 4 stages and each stage comprises 2 convolution blocks, and the group size per stage is denoted as "S1:S2:S3:S4". For Bit-Mixed the following group size combinations re used 4:8:8:16, 4:8:16:32, and 4:4:4:4. For comparison with state-of-the-art frameworks, different fixed and MXPDNN frameworks are used; \cite{zhou2016dorefa,zhang2018lq,choi2018pact,jung2019learning,li2019additive,cai2020rethinking,esser2019learned} and \cite{gong2019differentiable}. The \# of bits or average \# of bits is reported as well as the top-1\% accuracy. In addition, an ablation study is done in order to weigh different choices and justify the choices made for Bit-Mixer. In particular, the effect of TBN layers, the effect of knowledge distillation on Bit-Mixer, the effect of changing the quantization (symmetric/asymmetric, clipping/scaling), and the effect of considering different codebook sets from which the "bitwidth" is randomly drawn (This is different than codebook quantization where there is a set from which the "quantized value" is drawn). In addition, the ablation study includes a section where the authors propose a method to extract highly-performing (in terms of accuracy) sub-nets, given an average bitwidth budget, without further training based on HAWQ's second-order information.

Compared to AdaBits, when Bit-Mixer is used to quantize ImageNet on ResNet-\{18,34,50\} up to the second stage of the 3-stage optimization, the accuracy is very close. In fact, for ResNet-18, the accuracy reported by Bit-Mixer up to stage 2 optimization is higher than that reported for AdaBits for \{4,3,2\} bitwidths. Even after stage 3 optimization, and with random per-layer allocation, the accuracy drop experienced by Bit-Mixer compared to AdaBits is not severe. In all cases, Bit-Mixer was not compared to the full precision model baseline. Compared with the other state-of-the-art frameworks on ResNet-18, Bit-Mixer has a competitive edge on these, even when the "run-time mixed-precision facility" is not included in the comparison. To sum, Bit-Mixer is the very first framework that supports at runtime per layer mixed-precision by training one single meta-network. Its results are promising when compared to other frameworks.

\subsection{Reinforcement Learning-Based Optimization}
\subsubsection{A Reinforcement Learning Approach for Deep Quantization of Neural Networks (ReLeQ)}

ReLeQ \cite{elthakeb2018releq} is a framework proposed to automate learning the optimal bitwidths of weights across layers. The problem is formulated as an end-to-end Long Short-Term Memory (LSTM) based reinforcement learning (RL) approach that relies on optimizing a policy for optimal per-layer bitwidth allocation by efficiently exploring the huge hyper-parameter space of mixed-precision. By tuning a reward function, ReLeq guides the RL agent towards the optimized solution; a middle ground between accuracy and quantization. For the quantization technique, the authors rely on the method proposed in WRPN \cite{mishra2017wrpn}, which is based on rounding and is hence deterministic. Since ReLeq starts from a pre-trained network, the quantization method is categorized as retraining.

As in any RL framework, the notions of states, action space, reward, and policy/value-based agent network must be defined. As such, the authors first differentiate between "state of quantization" and "state of relative accuracy". The former is a network-specific parameter defined as the sum across network layers of the total computation cost (Multiply-Accumulate, MAC, operations) and memory cost (number of weights scaled by the ratio of the energy of memory accesses to the energy of MAC operations). It evaluates the advantages of quantization. "State of relative accuracy" is a metric that measures the shortcomings of quantization on performance. As such, it is defined as the ratio of current accuracy with the currently allocated bitwidths (quantized) across all layers during training, to the accuracy when the network trains without quantization. The two states defined are pivotal as they represent the observations that the RL agent grasps from the environment. These observations enable the agent to take actions. ReLeq adopts a flexible action space, where at each layer, the agent chooses to change the precision from any bitwidth from a discrete set of possible bitwidths to any other bitwidth from that set. Regarding the reward, ReLeq implements an asymmetric formulation for accuracy. The main reason behind needing an asymmetric formulation is the fact that the reward should reduce bitwidths across layers but maintain accuracy simultaneously, which results in an asymmetry between the accuracy and minimizing precision. ReLeq's reward formulation preserves that inherent asymmetry while signifying accuracy over precision minimization. Moreover, this formulation depends on the two states described earlier, which results in a smoother 2-Dimensional reward gradient as the agent gets closer to the optimal bitwidths allocation, and which speeds up the convergence of the agent. The framework penalizes the agent when it ventures to very low accuracy states by defining a threshold under which quantization is not possible since the loss in accuracy is not tolerated. As opposed to RL learning techniques that are based on policy only or value only, ReLeq utilizes Proximal Policy Optimization (PPO), which is an actor-critic technique whereby the agent comprises both policy and value networks \cite{schulman2017proximal}. The first hidden layer in both networks is an LSTM layer. While the policy network has two fully connected hidden layers -each comprising 128 neurons- and an output layer -with the number of neurons depending on the discrete possible quantizations-, the value network has two fully connected hidden layers after the LSTM layer with 128 and 64 neurons respectively.

Now that the essential components of an RL framework are defined, we summarize ReLeq as follows. Given a pre-trained network with full precision, the agent maneuvers the layers one by one, whereby at each layer the observation states are fed as an input to the policy and value networks which output the number of bitwidths. ReLeq then takes a stochastic action; it quantizes the weights by using the output bitwidth. If the network is shallow, short retraining is carried out to assess the effectiveness of the quantization, and afterward, the agent receives a reward; an estimated validation accuracy. Otherwise, if the network is deep, this short retraining is postponed until all the layers are allocated with predicted bitwidths in order to save time. Given the reward, the bitwidths, state of quantization, and state of accuracy are updated before the agent maneuvers the next layer. At the end of every episode, which is defined as one pass through the network, PPO is utilized to update the policy and value networks. When the agent completes the final episode (i.e. quantizes the last layer), a long retrain is carried out using the predicted precision across the layers. The final accuracy of the quantized network can then be determined.

To evaluate ReLeq, the authors choose to quantize AlexNet and MobileNet-V1 on ImageNet, ResNet-20, VGG-\{11,16\} and 5-layer SimpleNet on CIFAR-10. They also choose to quantize 10-layer SVHN on SVHN and LeNet on MNIST. This consists of a total of 8 experiments. Moreover, ReLeq (using 7 of the above 8 experiments; excluding VGG-16) is evaluated on a custom hardware accelerator, Stripes \cite{judd2016stripes} to evaluate energy and inference execution time, where the baseline is a network quantized to a fixed 8 bitwidth allocation across all layers. To evaluate the inference execution time of ReLeq (using 7 of the above 8 experiments; excluding VGG-16) on conventional hardware, the authors utilize the TVM compiler \cite{chen2018tvm} on Intel Core i7-4790 CPU with a baseline network similar to the one mentioned earlier. TVM and Stripes are also used to compare ReLeq (using AlexNet and LeNet) with ADMM \cite{ye2018unified} in terms of performance and energy. For all experiments, the initial (when the agent starts exploration) quantization bitwidth allocated for all layers is 8 bits. Then ReLeq handles assigning different bit precision across layers. The average bitwidth per network and the accuracy loss (compared to full precision networks) are reported.

The results reported by the authors, and in particular the average bitwidths, show that a fixed allocation of bitwidths across layers is not always optimal, which motivates the need for mixed precision. In the 8 experiments reported, the loss in accuracy of classification did not exceed 0.3\%. To further evaluate the bitwidth allocation resulting from ReLeq, Pareto analysis is carried out on the small networks (because it is infeasible on larger complex networks), and the bitwidths allocated by ReLeq were found to coincide with the Pareto frontier where the loss in accuracy can be recovered via fine-tuning. Compared to the baseline of fixed 8 bitwidth allocation across layers, ReLeq achieved an average of x2.2 speedup on conventional hardware using TVM compiler, and an average of x2.7 speedup on Stripes. Moreover, on Stripes and with the same 8 bitwidth fixed quantization baseline, ReLeq demonstrates an average of x2.7 energy savings. Finally, ReLeq provides speedups and benefits compared to ADMM on Stripes and conventional hardware using the TVM compiler. In conclusion, ReLeq presents itself as an efficient LSTM-based RL framework to automate mixed-precision, where it shows that it can understand the sensitivity of layers and make appropriate bitwidth assignments for the weights such that speedups and energy savings are realized.

\subsubsection{HAQ: Hardware-Aware Automated Quantization with Mixed-Precision (HAQ) \protect\footnote{https://github.com/mit-han-lab/haq}}

The purpose of HAQ \cite{wang2019haq} is to leverage RL in order to automate the quantization policy which dictates the bitwidths of weights and activations per layer, while taking the feedback of hardware accelerators into consideration. In particular, the authors develop a hardware simulator that feeds back the latency and energy to the RL agent. Deterministic rounding and k-mean clustering (vector quantization, for model size constraint; see below) are used as the quantization techniques. Quantization in HAQ is categorized as retraining.

One challenge in automating the learned quantization is the huge design search space. Maneuvering the design space intelligently by relying on rule-based heuristics calls for domain expertise, which renders these hand-crafted quantization policies unable to generalize across models. HAQ tries to tackle this challenge by automating the search through a "learning-based framework". On a lower abstraction level, another challenge lies in the efficiency of the quantization policy on real hardware platforms in terms of energy and latency. With the vast types of available hardware platforms, it becomes pivotal to incorporate those measures through a feedback loop into the design of the quantization policy. This is what HAQ does. Hereon, we briefly describe the essential components of RL used in HAQ. The observations in this framework are the statistics from the hardware platform and the layer configuration. In particular, a 10-Dimensional feature vector is defined as the observation. In fact, at each layer, the RL agent takes a step for weights and another for activations, where each step is an action to quantize the weights or activations. If the current layer at which the RL agent is trying to quantize is a convolution layer, it comprises different information: the index of the layer, the numbers of input/output channels, the action from the last step, the number of parameters, the kernel size, the stride, the input feature map size, an indicator for depth-wise convolutions, and a binary indicator for weights and activations. If the current layer is a fully connected layer, the 10-D observation vector has the index of the layer, the number of input/output hidden units, the size of the input feature, the number of parameters, an indicator for weights and activations, and the action from the previous step. A continuous action space, in the range of $[0,1]$ is adopted in HAQ in order to maintain the relative order. However, and in order to meet the energy and latency budgets, resource constraints are defined and the action space is limited. After all the layers have been given actions by the RL agent, the feedback from the hardware platform is used to determine whether the deployed currency meets the budget constraints or not. In the latter case, HAQ reduces the bitwidth of each layer in a sequential manner until the constraint is met. After all the parameters (weights and activations) in all layers are quantized in exploration, the model is fine-tuned via SGD for an extra epoch (for a short amount of time) and the validation (testing) accuracy is fed back to the RL agent as the reward. HAQ utilizes the Deep Deterministic Policy Gradient (DDPG) \cite{lillicrap2015continuous} for the agent. DDPG is an actor-critic algorithm (off-policy) optimized via ADAM \cite{kingma2014adam} during exploration. Both the actor and critic network share similar network architecture, where there are two fully connected layers with 400 hidden units each, followed by two fully connected layers with 300 and 1 hidden unit(s) respectively. The two networks differ only in that the actor-network has an extra sigmoid function to normalize the output between $0$ and $1$. An episode comprises many steps (recall that each step means one action is taken by the RL agent to quantize weights or activations at a layer), where actions are taken by the agent on all layers. It is worth noting that HAQ starts from a pre-trained full precision model, and when all the layers are quantized after exploration, the network is trained on the whole dataset for fine-tuning purposes.

HAQ is evaluated on several hardware platforms; cloud/edge and with spatial/temporal mixed-precision design with one constraint enforced on a budget at a time: \textit{C1} constraint on energy, \textit{C2} constraint on latency, and \textit{C3} constraint on model size. All experiments are performed using the ImageNet dataset \cite{deng2009imagenet}. Under latency constraints, HAQ is used with two hardware platforms: BitFusion \cite{sharma2018bit} (Cloud/Edge, ASIC, spatial) and BISMO \cite{umuroglu2018bismo} (Cloud/Edge, FPGA, temporal) architectures, and the models quantized are MobileNet-\{V1, V2\} \cite{howard2017mobilenets,sandler2018mobilenetv2}. In those experiments, HAQ fixes the first layer bitwidths to 8 bits. Under \textit{C2}, BitFusion is used as the hardware platform and MobileNet-V1 is the quantized model. And under \textit{C3}, no hardware feedback is needed and the quantized models are MobileNet-\{V1,V2\} and ResNet-50. The baselines used for experiments with \textit{C1} and \textit{C2} are PACT \cite{choi2018pact} and 8-bit fixed precision across layers, while those used with \textit{C3} experiments are Deep Compression \cite{han2015deep} and full precision networks. The authors reported the number of bits assigned per layer for most cases, and the budget values and top-1\%-5\% accuracies in all cases.

We focus on the top-1\% accuracies and budget savings herein. With \textit{C1} on BISMO and compared to PACT with similar latency, HAQ yields higher accuracies on cloud and on the edge and the bitwidths allocation across layers (for weights and activations) differs between edge and cloud and between pointwise layers and depthwise layers. Compared with the fixed 8 precision baseline, the accuracy of HAQ has little to no degradation (in some cases HAQ has better accuracy) but it yields up to x1.95 latency savings. With \textit{C1} on BitFusion and compared to PACT, HAQ outperforms the baseline with comparable latency. HAQ has up to x1.8 latency savings compared to the 8 fixed precision baseline with few losses in accuracy with close accuracies. When evaluated under \textit{C2}, HAQ has superior accuracy compared to PACT with similar energy and achieves up to x1.9 energy saving compared to the fixed 8 precision baseline. As for the evaluations under \textit{C3}, the proposed framework achieves higher accuracies when compared to the Deep Compression baseline with similar model sizes. Compared with the full precision model, HAQ achieves up to x8.03 memory size savings with close accuracies. Finally, HAQ contributes to automated, hardware-aware, specialized quantized models that are shown to be competitive with the baselines and under different budget constraints.

\subsubsection{AutoQ: Automated Kernel-Wise Neural Network Quantization}

AutoQ is proposed as an MXPDNN framework that relies on hierarchical deep reinforcement learning (DRL) to automatically and rapidly assign mixed-precision for weights kernels and activations across the layers of the DNN \cite{lou2019autoq}. Starting from a pre-trained network with full precision, AutoQ quantizes the weights (per kernel) and activations (per layer) based on the work done in \cite{zhang2018lq}. The quantization technique used is classified as training-aware, deterministic rounding. 

In order to choose the optimal bitwidths of weights and activations, AutoQ relies on two controllers; a high-level controller (HLC) and a low-level controller (LLC). The task of the HLC is choosing a bitwidth for each activation layer or the average bitwidth of all weight kernels of each convolutional layer, as a goal. The LLC produces a bitwidth for each weight kernel in a layer as an action. The two controls learn at the same time by trial and error. Since the HLC searches for the optimal bitwidths of activations layer-wise (and produces goals), and the LLC searches for the optimal bitwidths of the weight kernels in each layer (and produces actions), AutoQ is described as a two-level hierarchical DRL framework. We elaborate henceforth on the main components of AutoQ's DRL. The HLC and LLC are implemented using an off-policy correction (HIRO) \cite{nachum2018data} agent. The agent consists of an actor and a critic network, implemented similarly; two hidden layers, each having 300 units. The actor-network, however, has an additional sigmoid function that produces a normalized output. The observation or state is a vector of eleven elements as follows.
\begin{equation}
    observation^{(L_k,K_m)}=(L_k,K_m,ch_{in},ch_{o},s_{krnl},s_{strd},s_{ftr},b_{1},b_{2},g_{L_{k-1}},a_{(L_k,K_{m-1})})
\end{equation}
where $L_k$ and $K_m$ are the indices of the layer and weight kernel respectively. $ch_{in}$, $ch_{o}$, $s_{krnl}$, $s_{strd}$ and $s_{ftr}$ are the number of the input channels, number of the kernels, the kernel size, the stride, and the size of the input feature map respectively. While is a binary indicator for depthwise convolution, $b_2$ is a binary indicator whether the state describes a weight or activation. Moreover, $g_{L_{k-1}}$ and $a_(L_k,K_{m-1})$ represent the goal (see before) of the previous layer and the action bitwidth of the kernel in the previous layer respectively. The elements of the observation vector are normalized to $[0,1]$. As for the action and goal space, both the HLC and LLC use a continuous space. AutoQ designs an intrinsic reward and an extrinsic reward. When the agent arrives at a new state after taking an action, it gets the following extrinsic reward from the environment.
\begin{equation}
 R_e^{(L_k,K_m)}(NC,HWC)=log\frac{accuracy_{inference}(NC)^{t1}}{latency(NC,HWC)^{t2}*energy_{inference}(NC,HWC)^{t3}*area(NC,HWC)^{t4}}
\end{equation}
$NC$ and $HWC$ are the network and hardware configurations respectively. $t1$, $t2$, $t3$ and $t4$ are scaling factors that determine the impact of inference accuracy, latency, energy and FPGA area on the reward respectively. Given the maximum amount of hardware resources, i.e. for resource-constrained application, AutoQ sets $t1=1$, $t2=0$, $t3=0$ and $t4=0$ to get the best accuracy. For accuracy-guaranteed applications, AutoQ sets $t1=2$, $t2<1$, $t3<1$ and $t4<1$. In order to rapidly and accurately estimate the hardware latency, energy and area after each action while mitigating the synthesis time overhead on an FPGA, AutoQ relies on latency, area, and power models \cite{liu2013learning,zhou2019primal} instead of actually synthesizing on the FPGA. The HLC's job is to maximize the accumulative extrinsic reward. Alternatively, the LLC's job is to maximize the accumulative intrinsic reward. An intrinsic reward is received when the agent arrives at a state after the LLC generates actions based on goals produced by the HLC. The actions produced by the LLC should help complete HLC's goal and maximize the extrinsic reward, so the intrinsic reward is shaped as follows.
\begin{equation}
R_{i}^{(L_{k})}=(1-\phi)*(-||g_{L_{k}}*ch_{o}-\sum_{m=0}^{ch_o-1}a_{(L_k,K_m)}||_2)+\phi *\sum_{m=0}^{ch_o-1}R_e^{(L_k,K_m)}
\end{equation}
where $\phi$ is a factor that increases in the range $[0.1,0.8]$ as the number of training epochs increases.

AutoQ's workflow can be summarized as follows. After receiving an observation state from the environment (the quantized network model), the HLC either 1) produces a goal that is the bitiwdth for the activation layer or 2) generates a goal that is the average bitwidth of all weight kernels in a layer. Only in the latter case (2), the LLC produces an action for the wight kernel of the layer. In any goal case (1- or 2), the environment then estimates the hardware resources (energy, latency and area) by the models. Given these values (energy, latency and area) as well as the inference accuracy, the environment produces the extrinsic reward so that AutoQ can evaluate the LLC action. Based on all the actions of the LLC in a certain layer, the HLC then generates the intrinsic reward to evaluate how well the LLC implemented the goal.

For evaluation purposes, AutoQ quantizes ResNet-\{18/50\}, SqueezeNetV1 \cite{iandola2016squeezenet} and MobileNetV2 on ImageNet. The inference accuracy, energy consumption, and FPGA area are evaluated on a Xilinx Zynq-7020 FPGA that implements a temporal accelerator \cite{umuroglu2018bismo}. Two sets of experiments are carried out. In the first, resource-constrained ($t1=1$, $t2=0$, $t3=0$, and $t4=0$) kernel-wise quantized models by AutoQ (by setting a latency constraint) are compared to the layer-wise HAQ quantized models, to the fixed 4-bit (weights and activations) quantized models, and to the full precision models. For these experiments, the latency constraint is set to be equal to the inference latency of the fixed 4-bit quantized models, and AutoQ searches for the precision of weights and activations that guarantee the best inference accuracy given this latency constraint. The second set of experiments compares the accuracy-guaranteed ($t1=2$, $t2=0.5$, $t3=0$, and $t4=0$) quantized models by AutoQ to HAQ. For both sets of experiments, the top-\{1,5\}\% errors, average bitwidths of weights and activations, and the latency measures are reported. Moreover, another study shoes the energy and latency savings of the kernel-based AutoQ compared to HAQ and other quantization frameworks on both a temporal and spatial (BitFusion) accelerators.

The reported results on the resource-constrained experiments and with all models reveal that AutoQ achieves the lowest top-1\% error compared to HAQ and the fixed 4-bit precision quantized models. In the accuracy-guaranteed experiments, the top-1\% accuracy achieved by AutoQ still competes with the other baselines, but what is significant is that AutoQ yields the lowest average bitwidths of weights and activations in these experiments. In both sets of experiments, the top-1\% accuracy of AutoQ is close to the full precision model with a maximum of 2.5\% drop, and the inference latency is the lowest among all other baselines. In addition, it is reported that AutoQ yields a 39.04\% and 33.34\% average reduction in latency and energy respectively on the spatial accelerator platform (BitFusion) when compared to the layer-wise quantized models (using HAQ). To sum up, AutoQ is a hierarchical DRL kernel-wise mixed quantization framework that is able to reduce inference energy and latency while maintaining competitive accuracy.

\subsubsection{Simple Augmentation Goes A Long Way: ADRL For DNN Quantization (ADRL)}
\cite{ning2020simple} propose ADRL, an augmented (A) deep (D) reinforcement (R) learning (L) framework that allocates per-layer precisions for weights while keeping activations in full precision. The quantization scheme used is retraining, deterministic, rounding.

The motivation behind proposing ADRL is 1) the fact that deep reinforcement learning (DRL) suffers from slow convergence and sub-optimal results due to the overestimation bias and errors in the function approximation by the DRL agent, and 2) the challenges faced to have an efficient configuration of the mixed-precision search problem. To mitigate the variance and improve the convergence rate, ADRL introduces the "augmentation scheme" to the DRL networks as a complementary scheme. Consequently, ADRL makes up for the weaknesses in accuracy induced by the DRL agent approximations, especially in the early training stage. In particular, ADRL builds on top of the default actor-critic algorithm and is trained in the continuous action space with the off-policy agent DDPG as described henceforth. The process of mixed-precision quantization based on ADRL comprises three main stages: searching, fine-tuning, and evaluation where searching is the focus of ADRL. In short, the search stage has many episodes in which each comprises $L=\#OfLayers$ time steps, and at each time step, a bitwidth is selected for a certain layer. After finishing each episode, the network is quantized by the mixed-precision configuration that has the bit-width values for all the layers. After that, the quantized network is evaluated to compute the reward, and then the actor and critic parameters are updated. The algorithm then resets the target DNN network to the original one without quantization. After all episodes finish and in the following fine-tuning stage, the DNN network is quantized using the mixed-precision configuration selected at the end of the search stage. The network is then fine-tuned for some epochs. During the evaluation stage, the fine-tuned quantized model is run on the test dataset, and the resulting inference accuracy determines the quality of the quantized network.

Now we focus on the search stage. When searching for the optimal quantization configuration that satisfies a certain weight compression constraint, the environment is assumed to be fully observable, and a state is defined as a sequence of actions and observations as the environment is partially observed. The observation vector used is the same vector used in HAQ framework. In addition, the goal of the agent is to maximize the reward which is the expected return from the start state. In default actor-critic algorithm the actor-network- or policy approximator- maps a state into action and feeds the action into the critic network, while the critic network estimates the action and is updated by minimizing a loss function. In ADRL, however; the policy approximator is complemented with an augmented supplementary scheme constructed with domain knowledge or other mechanisms. The augmented policy approximator then comprises an extended actor-network and a refinement function. The extension of the actor-network allows the actor-network to generate multiple candidate actions instead of one, and the refinement function feeds the most promising action candidate to the critic network. ADRL modifies the last layer of the actor-network in order to get the extended version, whereby instead of having one neuron in the last layer, $N$ neurons are used for $N$ candidate actions. The most promising action is derived by the refinement function by selection. Given that the goodness of this selection relies on how well the critic network is trained, ADRL estimates a Q-value indicator (the indicator choice depends on the specific ADRL tasks, like mixed-precision)- whose output depends solely on the action and environment property rather than on the learnable actor and critic parameters- to complete the selection. Even after extending the actor-network and adding the refinement function, the actor-network parameters (like the critic network parameters) are still trained by DDPG. An effective augmentation leads to an estimated Q-value with lower variance, and it also leads to faster updates of the
critic network due to the larger step size. In order to incorporate mixed-precision, the choice of the Q-value is made such as to help choosing actions that are likely to yield higher quantization accuracy. Two types of Q-value indicators are used in this framework: profiling-based indicator and distance-based indicator. For the profiling-based indicator, at each time step (where an episode has $L$ timesteps and $L$ is the number of layers), the Q-value indicator computes the corresponding bit value for each one of the $N$ generated actions at a certain layer. Then if the bitwidth of a specific layer has a corresponding inference accuracy stored in a previously "memorized" dictionary, this stored accuracy is used as the indicated Q-value. Otherwise, the agent quantizes this layer while the others are kept unchanged (in full precision), ADRL evaluates the partially quantized network by relying on a test dataset, and the computed inference accuracy is added to the dictionary with its corresponding bitwidth. The inference accuracy resulting from the evaluation is used as an indicator for the fitness of the quantization level, and the refinement function chooses the action with the highest accuracy. The distance-based indicator measures the distance (L2 norm and KL-divergence) between the quantized and original weight, and the action with the least KL-divergence is chosen by the refinement function. Since the distance is a function of the quantized precision and the original weight, the distance computations are done ahead of time offline. When all the layers are allocated precisions, the agent verifies if the quantized network fulfills some compression ratio requirement. Otherwise, the precision is decreased layer by layer starting by the layer that yields the least accuracy loss. It is worth noting that an early termination mechanism is adopted to save time in the searching process, whereby this process is terminated when the inference accuracy has a variance among consecutive runs that is less than some threshold.

ADRL is tested by quantizing CifarNet and ResNet-20 on CIFAR-10, and it is tested as well by quantizing AlexNet and ResNet-50 on ImageNet. The profiling-based and distance-based Q-value indicators are used for ADRL, but we focus on the results of profiling-based ADRL because the reported results reveal that it outperforms its distance-based counterpart. HAQ, HAWQ, ReLeQ, and ZeroQ \cite{wang2019haq,dong2019hawq,elthakeb2018releq,cai2020zeroq} are used as baselines to compare with. The compression ratio and accuracy loss with respect to the full precision model are reported. Moreover, the learning speedups in the search and fine-tune stages and the overall speedup of the learning process of ADRL with respect to HAQ are reported.

The results reported reveal that ADRL (profiling-based) achieves good compression ratios with a zero accuracy loss with ResNet-\{20-50\} and AlexNet. Only ReLeQ has a zero accuracy loss on AlexNet among the baselines. ADRL provides higher overall speedups than HAQ in all experiments, and in particular, ADRL achieves up to 64x speedup compared to HAQ with ResNet-20 quantized on CIFAR-10. To sum up, ADRL is an MXPDNN framework proposed to boost the accuracy and speed of DRL, by introducing an augmentation scheme in the actor-network. The results reported by ADRL are competitive with other similar frameworks such as HAQ and ReLeQ.

\begin{table}[t]
\caption{Some results of RL-based and gradient-based (underlined) mixed precision frameworks, where at least weights have mixed precision.}
\label{tab:my-table11}
\resizebox{\textwidth}{!}
{
\begin{tabular}{|c|c|c|c|c|c|c|}
\hline
Framework & \begin{tabular}[c]{@{}c@{}}Quantization\\ Technique\end{tabular} & \begin{tabular}[c]{@{}c@{}}Activation\\ Quantization\end{tabular} & Dataset & Model & \begin{tabular}[c]{@{}c@{}}Top 1\%\\ Accuracy\end{tabular} & Savings \\ \hline
\multirow{3}{*}{ReLeQ*} & \multirow{3}{*}{\begin{tabular}[c]{@{}c@{}}Retraining\\ Deterministic\end{tabular}} & \multirow{3}{*}{None} & ImageNet & AlexNet & 63.22 & 3.95x (speedup) \tablefootnote{\label{fn1}On conventional hardware using TVM compiler. Baseline: 8-bit precision models for inference} \\ \cline{4-7} 
 &  &  & Cifar-10 & VGG-16 & 93.46 & - \\ \cline{4-7} 
 &  &  & MNIST & LeNet & 99.4 & 2.76x (speedup) \footref{fn1}\\ \hline
\multirow{3}{*}{HAQ} & \multirow{3}{*}{\begin{tabular}[c]{@{}c@{}}Retraining\\ Deterministic\end{tabular}} & \multirow{3}{*}{Mixed} & \multirow{3}{*}{ImageNet} & ResNet-50 & 76.14 \tablefootnote{With model size constraints} & 8.03x (Model Size) \tablefootnote{\label{fn2}Baseline: 8W8A models with energy and latency constraints, 32W32A models with model size constraint} \\ \cline{5-7} 
 &  &  &  & MobileNet-V1 \tablefootnote{BitFusion} & 70.90 \tablefootnote{With energy constraints} & 1.16x (Energy) \footref{fn2} \\ \cline{5-7} 
 &  &  &  & MobileNet-V2 \tablefootnote{BISMO Edge accelerator} & 71.89 \tablefootnote{With latency constraints} & 1.4x (Latency) \footref{fn2} \\ \hline
AutoQ & \begin{tabular}[c]{@{}c@{}}Retraining\\ Deterministic \end{tabular} & Mixed & ImageNet & ResNet-50 & 74.47 \tablefootnote{With resource-constraints} & 3.86x (Latency) \tablefootnote{Baseline: 16W16F models} \\ \hline
ADRL* & \begin{tabular}[c]{@{}c@{}}Retraining\\ Deterministic \end{tabular} & {None} & ImageNet & ResNet-50 & 83.2 & 10x (Model Size) \tablefootnote{\label{bl}Baseline: 32W32F} \\ \hline
\multirow{2}{*}{\underline{SLWP}} & \multirow{2}{*}{\begin{tabular}[c]{@{}c@{}}Retraining\\ Deterministic\end{tabular}} & \multirow{2}{*}{Mixed} & ImageNet & AlexNet \tablefootnote{1st layer is kept in full precision} & 52.54 \tablefootnote{Precision budget: 56} & - \\ \cline{4-7} 
 &  &  & MNIST & Similar to LeNet-5 & 98.86 \tablefootnote{Precision budget: 40} & - \\ \hline
\multirow{2}{*}{\underline{DQ}} & \multirow{2}{*}{\begin{tabular}[c]{@{}c@{}}Retraining\\ Deterministic\end{tabular}} & \multirow{2}{*}{Mixed} & ImageNet \tablefootnote{\label{memconst}Without constraints, and the networks are initialized with a pre-trained floating point network} & ResNet-18 \tablefootnote{\label{uniform}With best parametrization of uniform DQ} & 70.66 & 4.24x (Model Size)\tablefootnote{In terms of weight compression. Baseline: 32W32A models\label{w-s}}\\ \cline{4-7} 
 &  &  & Cifar-10 \footref{memconst} & ResNet-20 \footref{uniform} & 92.60 & - \\ \hline
\underline{OBACWA} & \begin{tabular}[c]{@{}c@{}}Post-training\\ Deterministic\end{tabular} & Mixed & ImageNet & ResNet-50 & 74.8 & 4x (Model Size) \footref{bl} \\ \hline
\multirow{3}{*}{\underline{Bayesian Bits}} & \multirow{3}{*}{\begin{tabular}[c]{@{}c@{}}Retraining\\ Deterministic\end{tabular}} & \multirow{3}{*}{Mixed} & ImageNet & MobileNet-V2 & 72.00 \tablefootnote{\label{logit}Output logits left in full precision} & 16x (GbOp) \footref{bl} \\ \cline{4-7} 
 &  &  & Cifar-10 & VGG-7 & 93.23 \footref{logit} & 196.08x (GbOp) \footref{bl} \\ \cline{4-7} 
 &  &  & MNIST & LeNet-5 & 99.30 \footref{logit} & 277.78x (GbOp)\footref{bl} \\ \hline
 \end{tabular}
}
\end{table}

\begin{table}[hbt!]
\caption{Some results of heuristic-based mixed precision frameworks, where at least weights have mixed precision.}
\label{tab:my-table12}
\resizebox{\textwidth}{!}{%
\begin{tabular}{|c|c|c|c|c|c|c|}
\hline
Framework & \begin{tabular}[c]{@{}c@{}}Quantization\\ Technique\end{tabular} & \begin{tabular}[c]{@{}c@{}}Activation\\ Quantization\end{tabular} & Dataset & Model & \begin{tabular}[c]{@{}c@{}}Top 1\%\\ Accuracy\end{tabular} & Savings \\ \hline
AQ & \begin{tabular}[c]{@{}c@{}}Post-training\\ Deterministic\end{tabular} & None & ImageNet & ResNet-50 \tablefootnote{Fully connected layers are kept in 16-bit precision} & 99.71 & 1.18x (Model Size)\tablefootnote{Baseline: Equal bit-width quantized model across all layers} \\ \hline
\multirow{3}{*}{PDB} & \multirow{3}{*}{\begin{tabular}[c]{@{}c@{}}Training-aware\\ Deterministic\end{tabular}} & \multirow{3}{*}{Fixed} & ImageNet & ResNet-18 \tablefootnote{The weights of
the input and output layers are left in full precision} & 65.03 & - \\ \cline{4-7} 
 &  &  & Cifar-10 & VGG-7 \tablefootnote{\label{pdb}The weights of the output layer are left in full precision} & 93.22 \tablefootnote{\label{manual}With manual fine-tuning} & - \\ \cline{4-7} 
 &  &  & Cifar-100 & VGG-7 \footref{pdb} & 71.53 \footref{manual} & - \\ \hline
HAWQ-V2 & \begin{tabular}[c]{@{}c@{}}Retraining\\ Deterministic\end{tabular} & Mixed & ImageNet & Inception-V3 & 75.98 & 12.04x (Model Size)\footref{w-s} \\ \hline
\multirow{3}{*}{HAWQ-V3} &\multirow{3}{*}{ \begin{tabular}[c]{@{}c@{}}Retraining/\\Post-training\\ Deterministic\end{tabular}} & \multirow{3}{*}{Fixed/Mixed} & \multirow{3}{*}{ImageNet} & \multirow{3}{*}{ResNet-50 \tablefootnote{All results reported with distillation}} & 77.58 & 1.15x (Model Size)\tablefootnote{Baseline: INT8 for weights and activations, with high model size constraint} \\ 
\cline{6-7} 
 &  &  &  &  & 76.97 & 1.13x (Speed)\tablefootnote{Baseline: INT8 for weights and activations, with high latency constraint} \\ 
\cline{6-7} 
 &  &  &  &  & 76.76 & 1.25x (BOPS)\tablefootnote{Baseline: INT8 for weights and activations, with high BOPS constraint} \\
\hline
\multirow{2}{*}{Hybrid-Net*} & \multirow{2}{*}{\begin{tabular}[c]{@{}c@{}}Post-training\\ Deterministic
\end{tabular}} & \multirow{2}{*}{Mixed} & ImageNet & ResNet-18 & 62.73 & 0.20x (Memory Compression)\tablefootnote{\label{mc1}Baseline: XNOR network}\\ \cline{4-7} 
 &  &  & Cifar-100 & ResNet-32 \tablefootnote{\label{identity}With identity shortcut connections at each layer} \tablefootnote{\label{res2}1st and last layer are kept in full precision} & 64.45 & 0.31x (Memory Compression)\footref{mc1}\\ \hline \multirow{2}{*}{ZeroQ} & \multirow{2}{*}{\begin{tabular}[c]{@{}c@{}}Post-training\\ Deterministic\end{tabular}} & \multirow{2}{*}{Fixed} & ImageNet & Inception-V3 & 78.76 & 5.34x (Model Size)\footref{bl} \\ \cline{4-7}
 &  &  & Cifar-10 & ResNet-20 & 93.87 & 5.20x (Model Size) \footref{bl} \\ \hline
OPQ & \begin{tabular}[c]{@{}c@{}}Retraining\\ Deterministic\end{tabular} & None & ImageNet & ResNet-50 & 76.41 & 38.03x (Model Size)\footref{w-s} \\ \hline
MPQNNCO & \begin{tabular}[c]{@{}c@{}}Retraining\\ Deterministic\end{tabular} & Mixed & ImageNet & ResNet-50 & 75.28 & 12.24x (Model Size)\footref{w-s} \\ \hline
\end{tabular}%
}
\end{table}

\begin{table}[h!]
\caption{Some results of meta-heuristic-based mixed precision frameworks, where at least weights have mixed precision.}
\label{tab:my-table13}
\resizebox{\textwidth}{!}{%
\begin{tabular}{|c|c|c|c|c|c|c|}
\hline
Framework & \begin{tabular}[c]{@{}c@{}}Quantization\\ Technique\end{tabular} & \begin{tabular}[c]{@{}c@{}}Activation\\ Quantization\end{tabular} & Dataset & Model & \begin{tabular}[c]{@{}c@{}}Top 1\%\\ Accuracy\end{tabular} & Savings \\ \hline
\multirow{2}{*}{DNAS} & \multirow{2}{*}{\begin{tabular}[c]{@{}c@{}}Retraining\\ Deterministic\end{tabular}} & \multirow{2}{*}{None/Mixed} & ImageNet & ResNet-34 \tablefootnote{\label{res3}ReLU-only preactivation}\tablefootnote{\label{res1}Activations, the 1st and the last layer are kept in full precision} & 74.61 & 10.6x (ModelSize)\footref{bl} \\ \cline{4-7} 
 &  &  & Cifar-10 & ResNet-110\footref{res1} & 95.07 & 12.5x (Model Size) \footref{bl} \\ \hline
\multirow{2}{*}{JASQ} & \multirow{2}{*}{\begin{tabular}[c]{@{}c@{}}Post-training\\ Deterministic\end{tabular}} & \multirow{2}{*}{None} & ImageNet & ResNet-152 \footref{res2} & 78.86 & 5.16x(Model Size) \footref{w-s} \\ \cline{4-7} 
 &  &  & Cifar-10 & JASQNet & 97.1 & - \\ \hline
\multirow{2}{*}{MPNASEE} & \multirow{2}{*}{\begin{tabular}[c]{@{}c@{}}Training-aware\\ Deterministic\end{tabular}} & \multirow{2}{*}{Mixed} & ImageNet & Theirs-Base & 71.77 & \begin{tabular}[c]{@{}c@{}}2.67x (Energy)\\ 2.99x (Latency) \tablefootnote{\label{mbnet}Baseline: 8-bit precision quantized MobileNet-V2}\end{tabular} \\ \cline{4-7} 
 &  &  & Cifar-100 & Theirs-Base & 78.73 & \begin{tabular}[c]{@{}c@{}}4.21x (Energy)\\ 4.41x (Latency) \footref{mbnet}\end{tabular} \\ \hline
APQ & \begin{tabular}[c]{@{}c@{}}Retraining\\ Deterministic\end{tabular} & Mixed & ImageNet & Their model  & 75.1 & 1.23x (BitOp)\footref{mbnet}\\ \hline
\multirow{2}{*}{BP-NAS} & \multirow{2}{*}{\begin{tabular}[c]{@{}c@{}}Retraining \\ Deterministic\end{tabular}} & \multirow{2}{*}{Mixed} & ImageNet & ResNet-50 & 76.67 & 71.65x (BitOp)\footref{bl} \\ \cline{4-7} 
 &  &  & Cifar-10 & ResNet-20 & 92.30 & 10.19x (Model Size)\footref{w-s}\\ \hline
{HMQ} & {\begin{tabular}[c]{@{}c@{}}Retraining\\ Deterministic\end{tabular}} & {Mixed} & ImageNet & ResNet-50 & 75.73 & 11.1x (Model Size)\footref{w-s} \\ \hline
EdMIPS & \begin{tabular}[c]{@{}c@{}}Training-aware\\ Deterministic\end{tabular} & Mixed & ImageNet & Inception-V3 & 72.4 & 3.85x (Model Size)\tablefootnote{Using efficient composite convolution with respect to vanilla parallel convolutions} \\ \hline
EvoQ & \begin{tabular}[c]{@{}c@{}}Post-training\\ Deterministic\end{tabular} & Fixed & ImageNet & ResNet-101 & 76.76 & 8x (Model Size) \footref{w-s}\\ \hline
Bit-Mixer & \begin{tabular}[c]{@{}c@{}}Retraining\\ Deterministic\end{tabular} & Mixed & ImageNet & ResNet-50 \tablefootnote{Random per layer bit width selection} & 73.2 & - \\ \hline
\end{tabular}%
}
\end{table}

\section{Discussion}

Hereon, we first compare the aforementioned MXPDNN frameworks. We start comparing the frameworks that belong to the same group according to their optimization technique, then we compare the frameworks across different groups. Afterwards, we compare the MXPDNN frameworks to the binary neural networks. Then, we define some guidelines for future works in mixed-precision. 

\subsection{Comparison Among MXPDNN Frameworks}
In order to compare among such frameworks, we divide the comparison into classes; intra-group and inter-group comparisons. In the former class of comparisons, we compare between frameworks within the same optimization group (gradient-based, RL-based, heuristic-based or meta-heuristic based group). In the class of inter-group comparisons, we compare the different frameworks across all four optimization groups.

\subsubsection{Intra-group Comparison}
\begin{enumerate}
\item Gradient-based optimization group: The optimization problem in DQ, SLWP and OBACWA is formulated as a constrained heuristic, but in DQ memory constraints are introduced in the optimization problem, and two quantization techniques are shown to have their suitable parametrizations, unlike SLWP and OBACWA which use the number of bits as the budget constraint. Moreover, in DQ, the optimal bitwidth per-layer allocation is not found directly, but rather inferred from the dynamic range and step size. Bayesian Bits considers the hardware constraint that the number of bits should be a power of two. Among the gradient-based optimization frameworks, OBACWA is the framework with the fastest run-time as it is a post-training framework that does not require extra fine-tuning. However, OBACWA evaluated only on ImageNet while the other frameworks in this group demonstrate results on other datasets like Cifar-10 and/or MNIST. 

\item RL-based optimization group: HAQ incorporates only one budget constraint at a time; either energy, latency, or model size in order to guide the search of the RL agent. It relies on energy and latency feedback received from the hardware platform (the environment). AutoQ, in comparison, accommodates for multiple hardware constraints (energy, latency) at a time along with accuracy. It does not rely on explicit hardware feedback from the FPGA (the hardware platform it is implemented on), it however relies on models that estimate the hardware resources. In that sense, it is faster than HAQ. ReLeq does not take any hardware feedback in the loop, even though it is tested on multiple hardware platforms and its reward tries to balance between higher accuracy, lower compute, and reduced memory with an inclination to prioritize higher accuracy. ADRL only considers weight compression as a constraint when searching for the optimal bitwidth configuration, and it does not take hardware feedback into the loop. ADRL, however; is capable of generating multiple actions at each RL episode step, and then chooses the best action (accuracy-wise). Both ReLeQ and ADRL do not quantize the activations, so their model savings are lower than the other frameworks in this group. Releq is the only framework in this group that relies on LSTM-based RL. In addition, AutoQ and ADRL are the only frameworks in this group that utilize "deep" RL (DRL), but AutoQ's DRL approach is hierarchical. While AutoQ and HAQ are evaluated on ImageNet, ADRL and ReLeQ are evaluated on Cifar-10 as well.

\item Heuristic-based optimization group: While AQ follows a theoretical analysis method that relates the total model accuracy to the quantization noise, PDB relies on input separability for solving the problem of mixed-precision. PDB, HAWQ-V2, and ZeroQ are the only frameworks in this group that are evaluated on both image classification and object detection tasks. Unlike PDB which progressively decreases the bitwidths of weights only, Hybrid-Net starts from a binary neural network then relies on PCA to increase the bitwidths of weights and activations at significant layers. In order to carry out the mixed-precision quantization, ZeroQ, HAWQ-\{V2,V3\} rely on sensitivity analysis. However, ZeroQ relies on KL divergence while HAWQ-V2 and HAWQ-V3 rely on second-order information as a sensitivity metric. Like HAWQ-\{V2,V3\}, MPQNNCO relies on the Hessian spectrum. However, MPQNNCO further approximates the Hessian matrix and relies on a greedy algorithm to solve the problem now formulated as an MCKP. HAWQ-V3 is an integer-only mixed/fixed precision framework that relies on the same sensitivity metric as HAWQ-V2, but utilizes ILP formulation to tackle the problem of mixed precision allocation (unlike the Pareto-frontier approach utilized in HAWQ-V2). HAWQ-V3 relies on direct hardware constraints (see subsection below) like latency in the ILP formulation, unlike HAWQ-V2 which only has a target model size. While ZeroQ, Hybrid-Net and AQ are post-training techniques, PDB is training-aware. HAWQ-V3 reports results with and without distillation so it can be used as a post-training or retraining framework. The rest of the frameworks in this group are retraining, so they require more time. OPQ and AQ do not quantize the activations, so they yield lower model savings. ZeroQ and PDB quantize the activations, but they utilize the same bitwidth across all layers (fixed quantization), this too sacrifices model savings for better accuracy.  

\item Meta-heuristic-based optimization group: DNAS and EdMIPS utilize differentiable search to find the optimal mixed-precision. While DNAS manually prunes the high search space, EdMIPS decouples the search space for weights bitwidth and activations bitwidth. Moreover, EdMIPS proposes "winner-takes-all" as opposed to the "sampling" strategy used in DNAS. JASQ, MPNASEE, and APQ in this group jointly search for network architecture and quantization. BP-NAS relies on a differentiable soft barrier penalty-based NAS to perform per-block precision search given some complexity constraint. DNAS, BP-NAS, and APQ rely on supernets in their optimizations. While EdMIPS relies on proxy-less NAS, MPNASEE utilizes a proxy task. EvoQ and JASQ rely on evolutionary search (in particular the tournament selection algorithm \cite{goldberg1991comparative}), but EvoQ utilizes a smaller amount of data to carry out the search. Hence, EvoQ can potentially optimize faster than JASQ. However, no comparison between the two frameworks is carried out in either works. Similarly, APQ relies on an evolutionary search to perform a resource-constrained (latency/energy) search for the optimal quantization policy. We note that MPNASEE and APQ are the only frameworks in this group that consider energy constraints, but APQ can alternatively consider a latency constraint, and it relies on a look-up table for faster evaluations. EvoQ and BP-NAS are evaluated on both image recognition and object detection tasks. BitMixer is the only framework in this group that claims to do at run-time quantization while relying on a 3-stage optimization technique. Even though Bit-Mixer provides the ability to adjust bit precision per layer at test time, it has not been tested on a real hardware platform. Moreover, BitMixer does the run-time change in precision randomly, and not according to a budget constraint imposed by hardware (like energy/latency as in MPNASEE and APQ). In addition, Bit-Mixer does not report the computational overhead of the technique, and the authors focus only on accuracy. HMQ proposes the hardware-friendly (as the learned quantization is a powers-of-two) HMQ blocks that rely on the Gumbel-Softmax estimator to jointly search for optimal bitwidths and thresholds of each quantizer. DNAS and MPNASEE, like HMQ, also rely on the Gumbel-Softmax estimator. JASQ and EvoQ are post-training frameworks, while EdMIPS and MPNASEE are training-aware. The rest of the frameworks in this group require fine-tuning. In addition, JASQ does not quantize the activations while EvoQ has a fixed precision for activations across the layers. 
 \end{enumerate}

\subsubsection{Inter-group Comparison}
Across the different groups, the gradient-based methods and the heuristic-based are the fastest (in runtime) as they do not rely on a search technique like differentiable NAS, RL, or evolutionary algorithms. We note that since HAWQ-V3 does not rely on exhaustive search techniques, but rather on solving an ILP formulation, it is probably the fastest framework among all MXPDNN frameworks summarized in this survey. Moreover, the frameworks in the RL-based optimization group focused mostly on the hardware constraints, hence they are the most "hardware-aware" in that sense. In that context, we differentiate between two categories of hardware constraints: 1) explicit and 2) implicit. Explicit hardware constraints are those limiting the following hardware resources: latency, memory, and energy. Implicit hardware constraints are those that set a budget for the model size, bitwidth, complexity, or degradation in accuracy. Imposing the latter type of hardware constraints would indirectly decrease the latency, energy, and memory usage of the hardware. HAQ, MPNASEE, AutoQ, DQ, HAWQ-V3, and APQ are under the umbrella of the explicit hardware constraints. All other constrained frameworks (i.e. BP-NAS, EdMIPS, EvoQ, SLWP, OBACWA, AQ, ZeroQ, HAWQ-V2, and MPQNNCO) belong to the implicit hardware constraints category. 

\subsection{Comparison Against Binary Neural Networks}
 The main motivation behind the MXPDNN frameworks is their promise of high energy efficiency and throughput. Logically, the Binary Neural Networks (BNNs) offers the optimal results in both energy efficiency and throughput when both weights and activations are quantized using a bitwidth of one. However, due to the extremely low precision, a higher loss in accuracy is incurred by such BNNs. Hence, hereon we juxtapose the state-of-art BNN accuracy results (adapted from \cite{qin2020binary}, and where both weights and activations are quantized with a bitwidth of one) against the best accuracy results reported by the MXPDNN frameworks (where both weights and activations are quantized in a mixed precision manner). Table \ref{BNN results} summarizes the comparison results on ImageNet and Cifar-10 datasets. As one can observe from the Table, the accuracy achieved by mixed precision frameworks is slightly higher than that of the BNNs for both datasets and across all models, with the exception of quantizing AlexNet on ImageNet. TSQ a notably higher accuracy than that achieved by SLWP when AlexNet is quantized on ImageNet. In general, the difference in accuracies reported between BNNs and mixed precision frameworks is not very high, which calls for more effort in better optimizing the mixed precision frameworks.
 
\begin{table}[hbt!]
\centering
\caption{Comparison between the highest top-1\% accuracies reported on BNNs, where both weights and activations are quantized using a bitwidth of one, and the highest top-1\% accuracies reported on MXPDNN frameworks, where both weights and activations are quantized in a mixed-precision manner.}
\label{BNN results}
{%
\begin{tabular}{|c|c|c|c|c|c|}
\hline
\multirow{2}{*}{Dataset} & \multirow{2}{*}{DNN Model} & \multicolumn{2}{c|}{MXDPNNs}    & \multicolumn{2}{c|}{BNNs}    \\ \cline{3-6}
 &  & Framework & Top-1 Accuracy(\%) & Framework & Top-1 Accuracy(\%) \\ \hline
\multirow{2}{*}{ImageNet} & AlexNet & SLWP & 52.54 & TSQ \cite{wang2018two} & 58.0 \\ \cline{2-6} 
 & ResNet-18 & DQ & 70.66 & CBCN \cite{liu2019circulant} & 61.4 \\ \hline
\multirow{1}{*}{Cifar-10} & ResNet-20 & DQ & 92.6 & BBG \cite{shen2020balanced} & 92.5 \\  \hline
\end{tabular}%
}
\end{table}

\subsection{Pruning in MXPDNNs}
Some of the aforementioned MXPDNN frameworks rely on pruning as an additional way toward more efficient neural networks. Pruning happens by removing unnecessary parameters from a large network, in order to improve generalization \cite{pruningsurvey}.  APQ searches for channel-wise pruning policy along with the quantization policy in a joint fashion. In particular, APQ solves a joint optimization problem that comprises two stages: 1) architecture search (coarse-grain architecture search and fine-grain channel search for channel-wise pruning) and 2) mixed-precision search. OPQ performs one-shot quantization and pruning. For pruning, OPQ computes the pruning masks for all layers, where these masks are used to remove the unimportant weights from a layer. Specifically, it first reformulates the problem of finding which weights could be removed to find the pruning ratios of all layers. The reformulated problem is solved by the Lagrangian multiplier by relying on the Newton-Raphson method. After finding the pruning ratios, the pruning masks are derived via magnitude-based thresholding. In addition, Bayesian Bits incorporates pruning via the learnable gates. The learnable gate is a variable $\in \{0,1\}$ that is placed on each of the quantized residual error tensors to control and optimize the effective bitwidth while jointly learning the quantization scales and the DNN parameters. The gate of the lowest bitwidth possible is what allows pruning, since deactivating this gate (i.e. setting it to 0) will yield a quantized value having 0-bits, i.e. pruned.

\subsection{Guidelines for Future MXPDNN Frameworks}
In this subsection, we suggest some guidelines for future mixed-precision frameworks. In particular, we believe that an MXPDNN framework must try to incorporate all of the following aspects.

\begin{enumerate}
    \item Hardware-awareness: As the goal of mixed-precision quantization is to be hardware friendly and deployable on devices with limiting computing capabilities such as edge devices, the requirements of the target device should be taken into consideration when the bit allocation is performed. Each device has its own energy/latency/memory resource budget that must be incorporated into the loop while optimizing for precision. 
    \item Model accuracy vs model compression balance: While the main target is to render DNN models deployable on low-compute capabilities, accuracy should not be sacrificed a lot. There should always be a balance between model accuracy and model compression. This is especially true in embedded devices in the healthcare system, where the accuracy of the network is highly related to the safety of the patient. 
    \item Run-time speed: The speed of convergence of the algorithm behind the framework is significant, especially if the framework is to be deployed on edge resources with limited computing capabilities and with changing requirements (see item below). 
    \item Run-time configurability: The ability of a network to adapt to changing hardware resource requirements over time is necessary. This is achieved by rendering the precision allocation flexible to change at run-time and when the hardware resources require so.

\end{enumerate}

\section{Conclusion} 
In conclusion, the idea of making DNNs more hardware-friendly is gaining more momentum as the need for deploying these algorithms on real hardware platforms increases. In this survey, we put together a summary of the different quantization techniques, followed by an elaboration on the main MXPDNN frameworks. We dig back to the earliest works on MXPDNNs, and we elaborate on newly proposed frameworks as well. Moreover, we present insights and comments on each framework, compare them, and summarize guidelines for future MXPDNN frameworks. To the best of our knowledge, we are the first to present such a survey on mixed-precision frameworks.


\begin{thebibliography}{100}
\bibitem{szegedy2013deep}
C.~Szegedy, A.~Toshev, and D.~Erhan, ``Deep neural networks for object
  detection,'' \emph{Advances in neural information processing systems},
  vol.~26, 2013.

\bibitem{tian2018deeptest}
Y.~Tian, K.~Pei, S.~Jana, and B.~Ray, ``Deeptest: Automated testing of
  deep-neural-network-driven autonomous cars,'' in \emph{Proceedings of the
  40th international conference on software engineering}, 2018, pp. 303--314.

\bibitem{hadidi2020toward}
R.~Hadidi, J.~Cao, M.~S. Ryoo, and H.~Kim, ``Toward collaborative inferencing
  of deep neural networks on internet-of-things devices,'' \emph{IEEE Internet
  of Things Journal}, vol.~7, no.~6, pp. 4950--4960, 2020.

\bibitem{rentzsch2021cadd}
P.~Rentzsch, M.~Schubach, J.~Shendure, and M.~Kircher,
  ``Cadd-splice—improving genome-wide variant effect prediction using deep
  learning-derived splice scores,'' \emph{Genome medicine}, vol.~13, no.~1, pp.
  1--12, 2021.

\bibitem{mohammadi2018enabling}
M.~Mohammadi and A.~Al-Fuqaha, ``Enabling cognitive smart cities using big data
  and machine learning: Approaches and challenges,'' \emph{IEEE Communications
  Magazine}, vol.~56, no.~2, pp. 94--101, 2018.

\bibitem{molchanov2019importance}
P.~Molchanov, A.~Mallya, S.~Tyree, I.~Frosio, and J.~Kautz, ``Importance
  estimation for neural network pruning,'' in \emph{Proceedings of the IEEE/CVF
  Conference on Computer Vision and Pattern Recognition}, 2019, pp.
  11\,264--11\,272.

\bibitem{li2016pruning}
H.~Li, A.~Kadav, I.~Durdanovic, H.~Samet, and H.~P. Graf, ``Pruning filters for
  efficient convnets,'' \emph{arXiv preprint arXiv:1608.08710}, 2016.

\bibitem{han2015learning}
S.~Han, J.~Pool, J.~Tran, and W.~Dally, ``Learning both weights and connections
  for efficient neural network,'' \emph{Advances in neural information
  processing systems}, vol.~28, 2015.

\bibitem{lecun1989optimal}
Y.~LeCun, J.~Denker, and S.~Solla, ``Optimal brain damage,'' \emph{Advances in
  neural information processing systems}, vol.~2, 1989.

\bibitem{yang2017designing}
T.-J. Yang, Y.-H. Chen, and V.~Sze, ``Designing energy-efficient convolutional
  neural networks using energy-aware pruning,'' in \emph{Proceedings of the
  IEEE conference on computer vision and pattern recognition}, 2017, pp.
  5687--5695.

\bibitem{mao2017exploring}
H.~Mao, S.~Han, J.~Pool, W.~Li, X.~Liu, Y.~Wang, and W.~J. Dally, ``Exploring
  the regularity of sparse structure in convolutional neural networks,''
  \emph{arXiv preprint arXiv:1705.08922}, 2017.

\bibitem{iandola2016squeezenet}
F.~N. Iandola, S.~Han, M.~W. Moskewicz, K.~Ashraf, W.~J. Dally, and K.~Keutzer,
  ``Squeezenet: Alexnet-level accuracy with 50x fewer parameters and< 0.5 mb
  model size,'' \emph{arXiv preprint arXiv:1602.07360}, 2016.

\bibitem{sandler2018mobilenetv2}
M.~Sandler, A.~Howard, M.~Zhu, A.~Zhmoginov, and L.-C. Chen, ``Mobilenetv2:
  Inverted residuals and linear bottlenecks,'' in \emph{Proceedings of the IEEE
  conference on computer vision and pattern recognition}, 2018, pp. 4510--4520.

\bibitem{tan2019efficientnet}
M.~Tan and Q.~Le, ``Efficientnet: Rethinking model scaling for convolutional
  neural networks,'' in \emph{International Conference on Machine
  Learning}.\hskip 1em plus 0.5em minus 0.4em\relax PMLR, 2019, pp. 6105--6114.

\bibitem{gholami2018squeezenext}
A.~Gholami, K.~Kwon, B.~Wu, Z.~Tai, X.~Yue, P.~Jin, S.~Zhao, and K.~Keutzer,
  ``Squeezenext: Hardware-aware neural network design,'' in \emph{Proceedings
  of the IEEE Conference on Computer Vision and Pattern Recognition Workshops},
  2018, pp. 1638--1647.

\bibitem{han2017efficient}
S.~Han, ``Efficient methods and hardware for deep learning,'' Ph.D.
  dissertation, Stanford University, 2017.

\bibitem{howard2019searching}
A.~Howard, M.~Sandler, G.~Chu, L.-C. Chen, B.~Chen, M.~Tan, W.~Wang, Y.~Zhu,
  R.~Pang, V.~Vasudevan \emph{et~al.}, ``Searching for mobilenetv3,'' in
  \emph{Proceedings of the IEEE/CVF International Conference on Computer
  Vision}, 2019, pp. 1314--1324.

\bibitem{wu2019fbnet}
B.~Wu, X.~Dai, P.~Zhang, Y.~Wang, F.~Sun, Y.~Wu, Y.~Tian, P.~Vajda, Y.~Jia, and
  K.~Keutzer, ``Fbnet: Hardware-aware efficient convnet design via
  differentiable neural architecture search,'' in \emph{Proceedings of the
  IEEE/CVF Conference on Computer Vision and Pattern Recognition}, 2019, pp.
  10\,734--10\,742.

\bibitem{hinton2015distilling}
G.~Hinton, O.~Vinyals, J.~Dean \emph{et~al.}, ``Distilling the knowledge in a
  neural network,'' \emph{arXiv preprint arXiv:1503.02531}, vol.~2, no.~7,
  2015.

\bibitem{mishra2017apprentice}
A.~Mishra and D.~Marr, ``Apprentice: Using knowledge distillation techniques to
  improve low-precision network accuracy,'' \emph{arXiv preprint
  arXiv:1711.05852}, 2017.

\bibitem{polino2018model}
A.~Polino, R.~Pascanu, and D.~Alistarh, ``Model compression via distillation
  and quantization,'' \emph{arXiv preprint arXiv:1802.05668}, 2018.

\bibitem{yin2020dreaming}
H.~Yin, P.~Molchanov, J.~M. Alvarez, Z.~Li, A.~Mallya, D.~Hoiem, N.~K. Jha, and
  J.~Kautz, ``Dreaming to distill: Data-free knowledge transfer via
  deepinversion,'' in \emph{Proceedings of the IEEE/CVF Conference on Computer
  Vision and Pattern Recognition}, 2020, pp. 8715--8724.

\bibitem{morgan1991experimental}
N.~Morgan \emph{et~al.}, ``Experimental determination of precision requirements
  for back-propagation training of artificial neural networks,'' in \emph{Proc.
  Second Int’l. Conf. Microelectronics for Neural Networks}.\hskip 1em plus
  0.5em minus 0.4em\relax Citeseer, 1991, pp. 9--16.

\bibitem{bhalgat2020lsq+}
Y.~Bhalgat, J.~Lee, M.~Nagel, T.~Blankevoort, and N.~Kwak, ``Lsq+: Improving
  low-bit quantization through learnable offsets and better initialization,''
  in \emph{Proceedings of the IEEE/CVF Conference on Computer Vision and
  Pattern Recognition Workshops}, 2020, pp. 696--697.

\bibitem{chin2020one}
T.-W. Chin, P.~I.-J. Chuang, V.~Chandra, and D.~Marculescu, ``One weight
  bitwidth to rule them all,'' in \emph{European Conference on Computer
  Vision}.\hskip 1em plus 0.5em minus 0.4em\relax Springer, 2020, pp. 85--103.

\bibitem{hubara2016binarized}
I.~Hubara, M.~Courbariaux, D.~Soudry, R.~El-Yaniv, and Y.~Bengio, ``Binarized
  neural networks,'' in \emph{Proceedings of the 30th International Conference
  on Neural Information Processing Systems}, 2016, pp. 4114--4122.

\bibitem{jacob2018quantization}
B.~Jacob, S.~Kligys, B.~Chen, M.~Zhu, M.~Tang, A.~Howard, H.~Adam, and
  D.~Kalenichenko, ``Quantization and training of neural networks for efficient
  integer-arithmetic-only inference,'' in \emph{Proceedings of the IEEE
  Conference on Computer Vision and Pattern Recognition}, 2018, pp. 2704--2713.

\bibitem{rastegari2016xnor}
M.~Rastegari, V.~Ordonez, J.~Redmon, and A.~Farhadi, ``Xnor-net: Imagenet
  classification using binary convolutional neural networks,'' in
  \emph{European conference on computer vision}.\hskip 1em plus 0.5em minus
  0.4em\relax Springer, 2016, pp. 525--542.

\bibitem{zhang2018lq}
D.~Zhang, J.~Yang, D.~Ye, and G.~Hua, ``Lq-nets: Learned quantization for
  highly accurate and compact deep neural networks,'' in \emph{Proceedings of
  the European conference on computer vision (ECCV)}, 2018, pp. 365--382.

\bibitem{zhou2017incremental}
A.~Zhou, A.~Yao, Y.~Guo, L.~Xu, and Y.~Chen, ``Incremental network
  quantization: Towards lossless cnns with low-precision weights,'' \emph{arXiv
  preprint arXiv:1702.03044}, 2017.

\bibitem{sharma2018bit}
H.~Sharma, J.~Park, N.~Suda, L.~Lai, B.~Chau, V.~Chandra, and H.~Esmaeilzadeh,
  ``Bit fusion: Bit-level dynamically composable architecture for accelerating
  deep neural network,'' in \emph{2018 ACM/IEEE 45th Annual International
  Symposium on Computer Architecture (ISCA)}.\hskip 1em plus 0.5em minus
  0.4em\relax IEEE, 2018, pp. 764--775.

\bibitem{dong2019hawq}
Z.~Dong, Z.~Yao, A.~Gholami, M.~W. Mahoney, and K.~Keutzer, ``Hawq: Hessian
  aware quantization of neural networks with mixed-precision,'' in
  \emph{Proceedings of the IEEE/CVF International Conference on Computer
  Vision}, 2019, pp. 293--302.

\bibitem{bulat2021bit}
A.~Bulat and G.~Tzimiropoulos, ``Bit-mixer: Mixed-precision networks with
  runtime bit-width selection,'' \emph{arXiv preprint arXiv:2103.17267}, 2021.

\bibitem{fiesler1990weight}
E.~Fiesler, A.~Choudry, and H.~J. Caulfield, ``Weight discretization paradigm
  for optical neural networks,'' in \emph{Optical interconnections and
  networks}, vol. 1281.\hskip 1em plus 0.5em minus 0.4em\relax International
  Society for Optics and Photonics, 1990, pp. 164--173.

\bibitem{cai2020zeroq}
Y.~Cai, Z.~Yao, Z.~Dong, A.~Gholami, M.~W. Mahoney, and K.~Keutzer, ``Zeroq: A
  novel zero shot quantization framework,'' in \emph{Proceedings of the
  IEEE/CVF Conference on Computer Vision and Pattern Recognition}, 2020, pp.
  13\,169--13\,178.

\bibitem{horowitz20141}
M.~Horowitz, ``1.1 computing's energy problem (and what we can do about it),''
  in \emph{2014 IEEE International Solid-State Circuits Conference Digest of
  Technical Papers (ISSCC)}.\hskip 1em plus 0.5em minus 0.4em\relax IEEE, 2014,
  pp. 10--14.

\bibitem{cai2020rethinking}
Z.~Cai and N.~Vasconcelos, ``Rethinking differentiable search for
  mixed-precision neural networks,'' in \emph{Proceedings of the IEEE/CVF
  Conference on Computer Vision and Pattern Recognition}, 2020, pp. 2349--2358.

\bibitem{qin2020binary}
H.~Qin, R.~Gong, X.~Liu, X.~Bai, J.~Song, and N.~Sebe, ``Binary neural
  networks: A survey,'' \emph{Pattern Recognition}, vol. 105, p. 107281, 2020.

\bibitem{zhou2016dorefa}
S.~Zhou, Y.~Wu, Z.~Ni, X.~Zhou, H.~Wen, and Y.~Zou, ``Dorefa-net: Training low
  bitwidth convolutional neural networks with low bitwidth gradients,''
  \emph{arXiv preprint arXiv:1606.06160}, 2016.

\bibitem{choi2018pact}
J.~Choi, Z.~Wang, S.~Venkataramani, P.~I.-J. Chuang, V.~Srinivasan, and
  K.~Gopalakrishnan, ``Pact: Parameterized clipping activation for quantized
  neural networks,'' \emph{arXiv preprint arXiv:1805.06085}, 2018.

\bibitem{kravchik2019low}
E.~Kravchik, F.~Yang, P.~Kisilev, and Y.~Choukroun, ``Low-bit quantization of
  neural networks for efficient inference,'' in \emph{Proceedings of the
  IEEE/CVF International Conference on Computer Vision Workshops}, 2019, pp.
  0--0.

\bibitem{banner2018post}
R.~Banner, Y.~Nahshan, E.~Hoffer, and D.~Soudry, ``Post-training 4-bit
  quantization of convolution networks for rapid-deployment,'' \emph{arXiv
  preprint arXiv:1810.05723}, 2018.

\bibitem{krishnamoorthi2018quantizing}
R.~Krishnamoorthi, ``Quantizing deep convolutional networks for efficient
  inference: A whitepaper,'' \emph{arXiv preprint arXiv:1806.08342}, 2018.

\bibitem{jin2019towards}
Q.~Jin, L.~Yang, and Z.~Liao, ``Towards efficient training for neural network
  quantization,'' \emph{arXiv preprint arXiv:1912.10207}, 2019.

\bibitem{mellempudi2017ternary}
N.~Mellempudi, A.~Kundu, D.~Mudigere, D.~Das, B.~Kaul, and P.~Dubey, ``Ternary
  neural networks with fine-grained quantization,'' \emph{arXiv preprint
  arXiv:1705.01462}, 2017.

\bibitem{han2015deep}
S.~Han, H.~Mao, and W.~J. Dally, ``Deep compression: Compressing deep neural
  networks with pruning, trained quantization and huffman coding,'' \emph{arXiv
  preprint arXiv:1510.00149}, 2015.

\bibitem{park2018value}
E.~Park, S.~Yoo, and P.~Vajda, ``Value-aware quantization for training and
  inference of neural networks,'' in \emph{Proceedings of the European
  Conference on Computer Vision (ECCV)}, 2018, pp. 580--595.

\bibitem{huang2021mxqn}
C.~Huang, P.~Liu, and L.~Fang, ``Mxqn: mixed quantization for reducing
  bit-width of weights and activations in deep convolutional neural networks,''
  \emph{Applied Intelligence}, pp. 1--14, 2021.

\bibitem{elthakeb2018releq}
A.~T. Elthakeb, P.~Pilligundla, F.~Mireshghallah, A.~Yazdanbakhsh, and
  H.~Esmaeilzadeh, ``Releq: A reinforcement learning approach for deep
  quantization of neural networks,'' \emph{arXiv preprint arXiv:1811.01704},
  2018.

\bibitem{allen2018learning}
Z.~Allen-Zhu, Y.~Li, and Y.~Liang, ``Learning and generalization in
  overparameterized neural networks, going beyond two layers,'' \emph{arXiv
  preprint arXiv:1811.04918}, 2018.

\bibitem{yao2021hawq}
Z.~Yao, Z.~Dong, Z.~Zheng, A.~Gholami, J.~Yu, E.~Tan, L.~Wang, Q.~Huang,
  Y.~Wang, M.~Mahoney \emph{et~al.}, ``Hawq-v3: Dyadic neural network
  quantization,'' in \emph{International Conference on Machine Learning}.\hskip
  1em plus 0.5em minus 0.4em\relax PMLR, 2021, pp. 11\,875--11\,886.

\bibitem{guo2018survey}
Y.~Guo, ``A survey on methods and theories of quantized neural networks,''
  \emph{arXiv preprint arXiv:1808.04752}, 2018.

\bibitem{gholami2021survey}
A.~Gholami, S.~Kim, Z.~Dong, Z.~Yao, M.~W. Mahoney, and K.~Keutzer, ``A survey
  of quantization methods for efficient neural network inference,'' \emph{arXiv
  preprint arXiv:2103.13630}, 2021.

\bibitem{courbariaux2015binaryconnect}
M.~Courbariaux, Y.~Bengio, and J.-P. David, ``Binaryconnect: Training deep
  neural networks with binary weights during propagations,'' \emph{arXiv
  preprint arXiv:1511.00363}, 2015.

\bibitem{gupta2015deep}
S.~Gupta, A.~Agrawal, K.~Gopalakrishnan, and P.~Narayanan, ``Deep learning with
  limited numerical precision,'' in \emph{International conference on machine
  learning}.\hskip 1em plus 0.5em minus 0.4em\relax PMLR, 2015, pp. 1737--1746.

\bibitem{wu2018training}
S.~Wu, G.~Li, F.~Chen, and L.~Shi, ``Training and inference with integers in
  deep neural networks,'' \emph{arXiv preprint arXiv:1802.04680}, 2018.

\bibitem{choi2016towards}
Y.~Choi, M.~El-Khamy, and J.~Lee, ``Towards the limit of network
  quantization,'' \emph{arXiv preprint arXiv:1612.01543}, 2016.

\bibitem{wu2016quantized}
J.~Wu, C.~Leng, Y.~Wang, Q.~Hu, and J.~Cheng, ``Quantized convolutional neural
  networks for mobile devices,'' in \emph{Proceedings of the IEEE Conference on
  Computer Vision and Pattern Recognition}, 2016, pp. 4820--4828.

\bibitem{park2017weighted}
E.~Park, J.~Ahn, and S.~Yoo, ``Weighted-entropy-based quantization for deep
  neural networks,'' in \emph{Proceedings of the IEEE Conference on Computer
  Vision and Pattern Recognition}, 2017, pp. 5456--5464.

\bibitem{gong2014compressing}
Y.~Gong, L.~Liu, M.~Yang, and L.~Bourdev, ``Compressing deep convolutional
  networks using vector quantization,'' \emph{arXiv preprint arXiv:1412.6115},
  2014.

\bibitem{li2016ternary}
F.~Li, B.~Zhang, and B.~Liu, ``Ternary weight networks,'' \emph{arXiv preprint
  arXiv:1605.04711}, 2016.

\bibitem{leng2018extremely}
C.~Leng, Z.~Dou, H.~Li, S.~Zhu, and R.~Jin, ``Extremely low bit neural network:
  Squeeze the last bit out with admm,'' in \emph{Proceedings of the AAAI
  Conference on Artificial Intelligence}, vol.~32, no.~1, 2018.

\bibitem{khoram2018adaptive}
S.~Khoram and J.~Li, ``Adaptive quantization of neural networks,'' in
  \emph{International Conference on Learning Representations}, 2018.

\bibitem{hou2016loss}
L.~Hou, Q.~Yao, and J.~T. Kwok, ``Loss-aware binarization of deep networks,''
  \emph{arXiv preprint arXiv:1611.01600}, 2016.

\bibitem{hwang2014fixed}
K.~Hwang and W.~Sung, ``Fixed-point feedforward deep neural network design
  using weights+ 1, 0, and- 1,'' in \emph{2014 IEEE Workshop on Signal
  Processing Systems (SiPS)}.\hskip 1em plus 0.5em minus 0.4em\relax IEEE,
  2014, pp. 1--6.

\bibitem{zhu2016trained}
C.~Zhu, S.~Han, H.~Mao, and W.~J. Dally, ``Trained ternary quantization,''
  \emph{arXiv preprint arXiv:1612.01064}, 2016.

\bibitem{tang1993multilayer}
C.~Z. Tang and H.~K. Kwan, ``Multilayer feedforward neural networks with single
  powers-of-two weights,'' \emph{IEEE Transactions on Signal Processing},
  vol.~41, no.~8, pp. 2724--2727, 1993.

\bibitem{achterhold2018variational}
J.~Achterhold, J.~M. Koehler, A.~Schmeink, and T.~Genewein, ``Variational
  network quantization,'' in \emph{International Conference on Learning
  Representations}, 2018.

\bibitem{muller2015rounding}
L.~K. Muller and G.~Indiveri, ``Rounding methods for neural networks with low
  resolution synaptic weights,'' \emph{arXiv preprint arXiv:1504.05767}, 2015.

\bibitem{lin2015neural}
Z.~Lin, M.~Courbariaux, R.~Memisevic, and Y.~Bengio, ``Neural networks with few
  multiplications,'' \emph{arXiv preprint arXiv:1510.03009}, 2015.

\bibitem{soudry2014expectation}
D.~Soudry, I.~Hubara, and R.~Meir, ``Expectation backpropagation:
  Parameter-free training of multilayer neural networks with continuous or
  discrete weights.'' in \emph{NIPS}, vol.~1, 2014, p.~2.

\bibitem{shayer2017learning}
O.~Shayer, D.~Levi, and E.~Fetaya, ``Learning discrete weights using the local
  reparameterization trick,'' \emph{arXiv preprint arXiv:1710.07739}, 2017.

\bibitem{jordan1999introduction}
M.~I. Jordan, Z.~Ghahramani, T.~S. Jaakkola, and L.~K. Saul, ``An introduction
  to variational methods for graphical models,'' \emph{Machine learning},
  vol.~37, no.~2, pp. 183--233, 1999.

\bibitem{wu2018mixed}
B.~Wu, Y.~Wang, P.~Zhang, Y.~Tian, P.~Vajda, and K.~Keutzer, ``Mixed precision
  quantization of convnets via differentiable neural architecture search,''
  \emph{arXiv preprint arXiv:1812.00090}, 2018.

\bibitem{jung2018joint}
S.~Jung, C.~Son, S.~Lee, J.~Son, Y.~Kwak, J.-J. Han, and C.~Choi, ``Joint
  training of low-precision neural network with quantization interval
  parameters,'' \emph{arXiv preprint arXiv:1808.05779}, vol.~2, 2018.

\bibitem{zhuang2018training}
B.~Zhuang, C.~Shen, and I.~Reid, ``Training compact neural networks with binary
  weights and low precision activations,'' \emph{arXiv preprint
  arXiv:1808.02631}, 2018.

\bibitem{seide20141}
F.~Seide, H.~Fu, J.~Droppo, G.~Li, and D.~Yu, ``1-bit stochastic gradient
  descent and its application to data-parallel distributed training of speech
  dnns,'' in \emph{Fifteenth Annual Conference of the International Speech
  Communication Association}, 2014.

\bibitem{strom2015scalable}
N.~Strom, ``Scalable distributed dnn training using commodity gpu cloud
  computing,'' in \emph{Sixteenth Annual Conference of the International Speech
  Communication Association}, 2015.

\bibitem{alistarh2017qsgd}
D.~Alistarh, D.~Grubic, J.~Li, R.~Tomioka, and M.~Vojnovic, ``Qsgd:
  Communication-efficient sgd via gradient quantization and encoding,''
  \emph{Advances in Neural Information Processing Systems}, vol.~30, pp.
  1709--1720, 2017.

\bibitem{dryden2016communication}
N.~Dryden, T.~Moon, S.~A. Jacobs, and B.~Van~Essen, ``Communication
  quantization for data-parallel training of deep neural networks,'' in
  \emph{2016 2nd Workshop on Machine Learning in HPC Environments
  (MLHPC)}.\hskip 1em plus 0.5em minus 0.4em\relax IEEE, 2016, pp. 1--8.

\bibitem{evoq}
Y.~Yuan, C.~Chen, X.~Hu, and S.~Peng, ``Evoq: Mixed precision quantization of
  dnns via sensitivity guided evolutionary search,'' in \emph{2020
  International Joint Conference on Neural Networks (IJCNN)}, 2020, pp. 1--8.

\bibitem{cai2017deep}
Z.~Cai, X.~He, J.~Sun, and N.~Vasconcelos, ``Deep learning with low precision
  by half-wave gaussian quantization,'' in \emph{Proceedings of the IEEE
  conference on computer vision and pattern recognition}, 2017, pp. 5918--5926.

\bibitem{zhou2018explicit}
A.~Zhou, A.~Yao, K.~Wang, and Y.~Chen, ``Explicit loss-error-aware quantization
  for low-bit deep neural networks,'' in \emph{Proceedings of the IEEE
  conference on computer vision and pattern recognition}, 2018, pp. 9426--9435.

\bibitem{mckinstry2018discovering}
J.~L. McKinstry, S.~K. Esser, R.~Appuswamy, D.~Bablani, J.~V. Arthur, I.~B.
  Yildiz, and D.~S. Modha, ``Discovering low-precision networks close to
  full-precision networks for efficient embedded inference,'' \emph{arXiv
  preprint arXiv:1809.04191}, 2018.

\bibitem{migacznvidia}
S.~Migacz, ``Nvidia 8-bit inference with tensorrt,'' in \emph{Proceedings of
  the GPU Technology Conference}.

\bibitem{wu2020integer}
H.~Wu, P.~Judd, X.~Zhang, M.~Isaev, and P.~Micikevicius, ``Integer quantization
  for deep learning inference: Principles and empirical evaluation,''
  \emph{arXiv preprint arXiv:2004.09602}, 2020.

\bibitem{habi2020hmq}
H.~V. Habi, R.~H. Jennings, and A.~Netzer, ``Hmq: Hardware friendly mixed
  precision quantization block for cnns,'' in \emph{Computer Vision--ECCV 2020:
  16th European Conference, Glasgow, UK, August 23--28, 2020, Proceedings, Part
  XXVI 16}.\hskip 1em plus 0.5em minus 0.4em\relax Springer, 2020, pp.
  448--463.

\bibitem{yu2020search}
H.~Yu, Q.~Han, J.~Li, J.~Shi, G.~Cheng, and B.~Fan, ``Search what you want:
  Barrier panelty nas for mixed precision quantization,'' in \emph{European
  Conference on Computer Vision}.\hskip 1em plus 0.5em minus 0.4em\relax
  Springer, 2020, pp. 1--16.

\bibitem{wang2020apq}
T.~Wang, K.~Wang, H.~Cai, J.~Lin, Z.~Liu, H.~Wang, Y.~Lin, and S.~Han, ``Apq:
  Joint search for network architecture, pruning and quantization policy,'' in
  \emph{Proceedings of the IEEE/CVF Conference on Computer Vision and Pattern
  Recognition}, 2020, pp. 2078--2087.

\bibitem{chen2018joint}
Y.~Chen, G.~Meng, Q.~Zhang, X.~Zhang, L.~Song, S.~Xiang, and C.~Pan, ``Joint
  neural architecture search and quantization,'' \emph{arXiv preprint
  arXiv:1811.09426}, 2018.

\bibitem{gong2019differentiable}
R.~Gong, X.~Liu, S.~Jiang, T.~Li, P.~Hu, J.~Lin, F.~Yu, and J.~Yan,
  ``Differentiable soft quantization: Bridging full-precision and low-bit
  neural networks,'' in \emph{Proceedings of the IEEE/CVF International
  Conference on Computer Vision}, 2019, pp. 4852--4861.

\bibitem{chen2021towards}
W.~Chen, P.~Wang, and J.~Cheng, ``Towards mixed-precision quantization of
  neural networks via constrained optimization,'' in \emph{Proceedings of the
  IEEE/CVF International Conference on Computer Vision}, 2021, pp. 5350--5359.

\bibitem{hu2021opq}
P.~Hu, X.~Peng, H.~Zhu, M.~M.~S. Aly, and J.~Lin, ``Opq: Compressing deep
  neural networks with one-shot pruning-quantization,'' in \emph{Proceedings of
  the Thirty-Fifth AAAI Conference on Artificial Intelligence (AAAI-21),
  Vancouver, VN, Canada}, 2021, pp. 2--9.

\bibitem{chakraborty2020constructing}
I.~Chakraborty, D.~Roy, I.~Garg, A.~Ankit, and K.~Roy, ``Constructing
  energy-efficient mixed-precision neural networks through principal component
  analysis for edge intelligence,'' \emph{Nature Machine Intelligence}, vol.~2,
  no.~1, pp. 43--55, 2020.

\bibitem{zhou2018adaptive}
Y.~Zhou, S.-M. Moosavi-Dezfooli, N.-M. Cheung, and P.~Frossard, ``Adaptive
  quantization for deep neural network,'' in \emph{Proceedings of the AAAI
  Conference on Artificial Intelligence}, vol.~32, no.~1, 2018.

\bibitem{chu2019mixed}
T.~Chu, Q.~Luo, J.~Yang, and X.~Huang, ``Mixed-precision quantized neural
  network with progressively decreasing bitwidth for image classification and
  object detection,'' \emph{arXiv preprint arXiv:1912.12656}, 2019.

\bibitem{ning2020simple}
L.~Ning, G.~Chen, W.~Zhang, and X.~Shen, ``Simple augmentation goes a long way:
  Adrl for dnn quantization,'' in \emph{International Conference on Learning
  Representations}, 2020.

\bibitem{wang2019haq}
K.~Wang, Z.~Liu, Y.~Lin, J.~Lin, and S.~Han, ``Haq: Hardware-aware automated
  quantization with mixed precision,'' in \emph{Proceedings of the IEEE/CVF
  Conference on Computer Vision and Pattern Recognition}, 2019, pp. 8612--8620.

\bibitem{van2020bayesian}
M.~van Baalen, C.~Louizos, M.~Nagel, R.~A. Amjad, Y.~Wang, T.~Blankevoort, and
  M.~Welling, ``Bayesian bits: Unifying quantization and pruning,'' \emph{arXiv
  preprint arXiv:2005.07093}, 2020.

\bibitem{uhlich2019differentiable}
S.~Uhlich, L.~Mauch, K.~Yoshiyama, F.~Cardinaux, J.~A. Garcia, S.~Tiedemann,
  T.~Kemp, and A.~Nakamura, ``Differentiable quantization of deep neural
  networks,'' \emph{arXiv preprint arXiv:1905.11452}, vol.~2, no.~8, 2019.

\bibitem{lacey2018stochastic}
G.~Lacey, G.~W. Taylor, and S.~Areibi, ``Stochastic layer-wise precision in
  deep neural networks,'' \emph{arXiv preprint arXiv:1807.00942}, 2018.

\bibitem{zhe2019optimizing}
W.~Zhe, J.~Lin, V.~Chandrasekhar, and B.~Girod, ``Optimizing the bit allocation
  for compression of weights and activations of deep neural networks,'' in
  \emph{2019 IEEE International Conference on Image Processing (ICIP)}.\hskip
  1em plus 0.5em minus 0.4em\relax IEEE, 2019, pp. 3826--3830.

\bibitem{bengio2013estimating}
Y.~Bengio, N.~L{\'e}onard, and A.~Courville, ``Estimating or propagating
  gradients through stochastic neurons for conditional computation,''
  \emph{arXiv preprint arXiv:1308.3432}, 2013.

\bibitem{srivastava2014dropout}
N.~Srivastava, G.~Hinton, A.~Krizhevsky, I.~Sutskever, and R.~Salakhutdinov,
  ``Dropout: a simple way to prevent neural networks from overfitting,''
  \emph{The journal of machine learning research}, vol.~15, no.~1, pp.
  1929--1958, 2014.

\bibitem{jang2016categorical}
E.~Jang, S.~Gu, and B.~Poole, ``Categorical reparameterization with
  gumbel-softmax,'' \emph{arXiv preprint arXiv:1611.01144}, 2016.

\bibitem{lecun1998gradient}
Y.~LeCun, L.~Bottou, Y.~Bengio, and P.~Haffner, ``Gradient-based learning
  applied to document recognition,'' \emph{Proceedings of the IEEE}, vol.~86,
  no.~11, pp. 2278--2324, 1998.

\bibitem{krizhevsky2012imagenet}
A.~Krizhevsky, I.~Sutskever, and G.~E. Hinton, ``Imagenet classification with
  deep convolutional neural networks,'' \emph{Advances in neural information
  processing systems}, vol.~25, pp. 1097--1105, 2012.

\bibitem{bertsekas2014constrained}
D.~P. Bertsekas, \emph{Constrained optimization and Lagrange multiplier
  methods}.\hskip 1em plus 0.5em minus 0.4em\relax Academic press, 2014.

\bibitem{tung2018clip}
F.~Tung and G.~Mori, ``Clip-q: Deep network compression learning by in-parallel
  pruning-quantization,'' in \emph{Proceedings of the IEEE conference on
  computer vision and pattern recognition}, 2018, pp. 7873--7882.

\bibitem{faraone2018syq}
J.~Faraone, N.~Fraser, M.~Blott, and P.~H. Leong, ``Syq: Learning symmetric
  quantization for efficient deep neural networks,'' in \emph{Proceedings of
  the IEEE Conference on Computer Vision and Pattern Recognition}, 2018, pp.
  4300--4309.

\bibitem{louizos2017learning}
C.~Louizos, M.~Welling, and D.~P. Kingma, ``Learning sparse neural networks
  through $ l\_0 $ regularization,'' \emph{arXiv preprint arXiv:1712.01312},
  2017.

\bibitem{kingma2013auto}
D.~P. Kingma and M.~Welling, ``Auto-encoding variational bayes,'' \emph{arXiv
  preprint arXiv:1312.6114}, 2013.

\bibitem{rezende2014stochastic}
D.~J. Rezende, S.~Mohamed, and D.~Wierstra, ``Stochastic backpropagation and
  approximate inference in deep generative models,'' in \emph{International
  conference on machine learning}.\hskip 1em plus 0.5em minus 0.4em\relax PMLR,
  2014, pp. 1278--1286.

\bibitem{louizos2018relaxed}
C.~Louizos, M.~Reisser, T.~Blankevoort, E.~Gavves, and M.~Welling, ``Relaxed
  quantization for discretized neural networks,'' \emph{arXiv preprint
  arXiv:1810.01875}, 2018.

\bibitem{esser2019learned}
S.~K. Esser, J.~L. McKinstry, D.~Bablani, R.~Appuswamy, and D.~S. Modha,
  ``Learned step size quantization,'' \emph{arXiv preprint arXiv:1902.08153},
  2019.

\bibitem{nagel2020up}
M.~Nagel, R.~A. Amjad, M.~Van~Baalen, C.~Louizos, and T.~Blankevoort, ``Up or
  down? adaptive rounding for post-training quantization,'' in
  \emph{International Conference on Machine Learning}.\hskip 1em plus 0.5em
  minus 0.4em\relax PMLR, 2020, pp. 7197--7206.

\bibitem{jain2019trained}
S.~R. Jain, A.~Gural, M.~Wu, and C.~Dick, ``Trained uniform quantization for
  accurate and efficient neural network inference on fixed-point hardware,''
  \emph{arXiv preprint arXiv:1903.08066}, vol.~6, p.~6, 2019.

\bibitem{uhlich2019mixed}
S.~Uhlich, L.~Mauch, F.~Cardinaux, K.~Yoshiyama, J.~A. Garcia, S.~Tiedemann,
  T.~Kemp, and A.~Nakamura, ``Mixed precision dnns: All you need is a good
  parametrization,'' \emph{arXiv preprint arXiv:1905.11452}, 2019.

\bibitem{fawzi2016robustness}
A.~Fawzi, S.-M. Moosavi-Dezfooli, and P.~Frossard, ``Robustness of classifiers:
  from adversarial to random noise,'' \emph{arXiv preprint arXiv:1608.08967},
  2016.

\bibitem{lin2016fixed}
D.~Lin, S.~Talathi, and S.~Annapureddy, ``Fixed point quantization of deep
  convolutional networks,'' in \emph{International conference on machine
  learning}.\hskip 1em plus 0.5em minus 0.4em\relax PMLR, 2016, pp. 2849--2858.

\bibitem{liu2018bi}
Z.~Liu, B.~Wu, W.~Luo, X.~Yang, W.~Liu, and K.-T. Cheng, ``Bi-real net:
  Enhancing the performance of 1-bit cnns with improved representational
  capability and advanced training algorithm,'' in \emph{Proceedings of the
  European conference on computer vision (ECCV)}, 2018, pp. 722--737.

\bibitem{courbariaux2016binarized}
M.~Courbariaux, I.~Hubara, D.~Soudry, R.~El-Yaniv, and Y.~Bengio, ``Binarized
  neural networks: Training deep neural networks with weights and activations
  constrained to+ 1 or-1,'' \emph{arXiv preprint arXiv:1602.02830}, 2016.

\bibitem{dong2019hawqv2}
Z.~Dong, Z.~Yao, Y.~Cai, D.~Arfeen, A.~Gholami, M.~W. Mahoney, and K.~Keutzer,
  ``Hawq-v2: Hessian aware trace-weighted quantization of neural networks,''
  \emph{arXiv preprint arXiv:1911.03852}, 2019.

\bibitem{avron2011randomized}
H.~Avron and S.~Toledo, ``Randomized algorithms for estimating the trace of an
  implicit symmetric positive semi-definite matrix,'' \emph{Journal of the ACM
  (JACM)}, vol.~58, no.~2, pp. 1--34, 2011.

\bibitem{li2019fully}
R.~Li, Y.~Wang, F.~Liang, H.~Qin, J.~Yan, and R.~Fan, ``Fully quantized network
  for object detection,'' in \emph{Proceedings of the IEEE/CVF Conference on
  Computer Vision and Pattern Recognition}, 2019, pp. 2810--2819.

\bibitem{lin2017focal}
T.-Y. Lin, P.~Goyal, R.~Girshick, K.~He, and P.~Doll{\'a}r, ``Focal loss for
  dense object detection,'' in \emph{Proceedings of the IEEE international
  conference on computer vision}, 2017, pp. 2980--2988.

\bibitem{lin2014microsoft}
T.-Y. Lin, M.~Maire, S.~Belongie, J.~Hays, P.~Perona, D.~Ramanan,
  P.~Doll{\'a}r, and C.~L. Zitnick, ``Microsoft coco: Common objects in
  context,'' in \emph{European conference on computer vision}.\hskip 1em plus
  0.5em minus 0.4em\relax Springer, 2014, pp. 740--755.

\bibitem{garg2019low}
I.~Garg, P.~Panda, and K.~Roy, ``A low effort approach to structured cnn design
  using pca,'' \emph{IEEE Access}, vol.~8, pp. 1347--1360, 2019.

\bibitem{prabhu2018hybrid}
A.~Prabhu, V.~Batchu, R.~Gajawada, S.~A. Munagala, and A.~Namboodiri, ``Hybrid
  binary networks: optimizing for accuracy, efficiency and memory,'' in
  \emph{2018 IEEE Winter Conference on Applications of Computer Vision
  (WACV)}.\hskip 1em plus 0.5em minus 0.4em\relax IEEE, 2018, pp. 821--829.

\bibitem{choukroun2019low}
Y.~Choukroun, E.~Kravchik, F.~Yang, and P.~Kisilev, ``Low-bit quantization of
  neural networks for efficient inference.'' in \emph{ICCV Workshops}, 2019,
  pp. 3009--3018.

\bibitem{zhao2019improving}
R.~Zhao, Y.~Hu, J.~Dotzel, C.~De~Sa, and Z.~Zhang, ``Improving neural network
  quantization without retraining using outlier channel splitting,'' in
  \emph{International conference on machine learning}.\hskip 1em plus 0.5em
  minus 0.4em\relax PMLR, 2019, pp. 7543--7552.

\bibitem{nagel2019data}
M.~Nagel, M.~v. Baalen, T.~Blankevoort, and M.~Welling, ``Data-free
  quantization through weight equalization and bias correction,'' in
  \emph{Proceedings of the IEEE/CVF International Conference on Computer
  Vision}, 2019, pp. 1325--1334.

\bibitem{haroush2020knowledge}
M.~Haroush, I.~Hubara, E.~Hoffer, and D.~Soudry, ``The knowledge within:
  Methods for data-free model compression,'' in \emph{Proceedings of the
  IEEE/CVF Conference on Computer Vision and Pattern Recognition}, 2020, pp.
  8494--8502.

\bibitem{chen2018tvm}
T.~Chen, T.~Moreau, Z.~Jiang, H.~Shen, E.~Q. Yan, L.~Wang, Y.~Hu, L.~Ceze,
  C.~Guestrin, and A.~Krishnamurthy, ``Tvm: end-to-end optimization stack for
  deep learning,'' \emph{arXiv preprint arXiv:1802.04799}, vol.~11, p.~20,
  2018.

\bibitem{mitchell2011pulp}
S.~Mitchell, M.~OSullivan, and I.~Dunning, ``Pulp: a linear programming toolkit
  for python,'' \emph{The University of Auckland, Auckland, New Zealand},
  vol.~65, 2011.

\bibitem{hawqv31}
``Nvidia. cutlass library,'' \url{https://github. com/NVIDIA/cutlass},
  accessed: 2022-05-21.

\bibitem{he2016deep}
K.~He, X.~Zhang, S.~Ren, and J.~Sun, ``Deep residual learning for image
  recognition,'' in \emph{Proceedings of the IEEE conference on computer vision
  and pattern recognition}, 2016, pp. 770--778.

\bibitem{hubara2020improving}
I.~Hubara, Y.~Nahshan, Y.~Hanani, R.~Banner, and D.~Soudry, ``Improving post
  training neural quantization: Layer-wise calibration and integer
  programming,'' \emph{arXiv preprint arXiv:2006.10518}, 2020.

\bibitem{he2018learning}
X.~He and J.~Cheng, ``Learning compression from limited unlabeled data,'' in
  \emph{Proceedings of the European Conference on Computer Vision (ECCV)},
  2018, pp. 752--769.

\bibitem{tung2018deep}
F.~Tung and G.~Mori, ``Deep neural network compression by in-parallel
  pruning-quantization,'' \emph{IEEE transactions on pattern analysis and
  machine intelligence}, vol.~42, no.~3, pp. 568--579, 2018.

\bibitem{yang2020automatic}
H.~Yang, S.~Gui, Y.~Zhu, and J.~Liu, ``Automatic neural network compression by
  sparsity-quantization joint learning: A constrained optimization-based
  approach,'' in \emph{Proceedings of the IEEE/CVF Conference on Computer
  Vision and Pattern Recognition}, 2020, pp. 2178--2188.

\bibitem{lin2017towards}
X.~Lin, C.~Zhao, and W.~Pan, ``Towards accurate binary convolutional neural
  network,'' \emph{Advances in neural information processing systems}, vol.~30,
  2017.

\bibitem{lou2019autoq}
Q.~Lou, F.~Guo, L.~Liu, M.~Kim, and L.~Jiang, ``Autoq: Automated kernel-wise
  neural network quantization,'' \emph{arXiv preprint arXiv:1902.05690}, 2019.

\bibitem{krizhevsky2009learning}
A.~Krizhevsky, G.~Hinton \emph{et~al.}, ``Learning multiple layers of features
  from tiny images,'' 2009.

\bibitem{he2016identity}
K.~He, X.~Zhang, S.~Ren, and J.~Sun, ``Identity mappings in deep residual
  networks,'' in \emph{European conference on computer vision}.\hskip 1em plus
  0.5em minus 0.4em\relax Springer, 2016, pp. 630--645.

\bibitem{deng2009imagenet}
J.~Deng, W.~Dong, R.~Socher, L.-J. Li, K.~Li, and L.~Fei-Fei, ``Imagenet: A
  large-scale hierarchical image database,'' in \emph{2009 IEEE conference on
  computer vision and pattern recognition}.\hskip 1em plus 0.5em minus
  0.4em\relax Ieee, 2009, pp. 248--255.

\bibitem{zoph2018learning}
B.~Zoph, V.~Vasudevan, J.~Shlens, and Q.~V. Le, ``Learning transferable
  architectures for scalable image recognition,'' in \emph{Proceedings of the
  IEEE conference on computer vision and pattern recognition}, 2018, pp.
  8697--8710.

\bibitem{goldberg1991comparative}
D.~E. Goldberg and K.~Deb, ``A comparative analysis of selection schemes used
  in genetic algorithms,'' in \emph{Foundations of genetic algorithms}.\hskip
  1em plus 0.5em minus 0.4em\relax Elsevier, 1991, vol.~1, pp. 69--93.

\bibitem{pham2018efficient}
H.~Pham, M.~Guan, B.~Zoph, Q.~Le, and J.~Dean, ``Efficient neural architecture
  search via parameters sharing,'' in \emph{International Conference on Machine
  Learning}.\hskip 1em plus 0.5em minus 0.4em\relax PMLR, 2018, pp. 4095--4104.

\bibitem{huang2017densely}
G.~Huang, Z.~Liu, L.~Van Der~Maaten, and K.~Q. Weinberger, ``Densely connected
  convolutional networks,'' in \emph{Proceedings of the IEEE conference on
  computer vision and pattern recognition}, 2017, pp. 4700--4708.

\bibitem{liu2018progressive}
C.~Liu, B.~Zoph, M.~Neumann, J.~Shlens, W.~Hua, L.-J. Li, L.~Fei-Fei,
  A.~Yuille, J.~Huang, and K.~Murphy, ``Progressive neural architecture
  search,'' in \emph{Proceedings of the European conference on computer vision
  (ECCV)}, 2018, pp. 19--34.

\bibitem{real2019regularized}
E.~Real, A.~Aggarwal, Y.~Huang, and Q.~V. Le, ``Regularized evolution for image
  classifier architecture search,'' in \emph{Proceedings of the aaai conference
  on artificial intelligence}, vol.~33, no.~01, 2019, pp. 4780--4789.

\bibitem{liu2018darts}
H.~Liu, K.~Simonyan, and Y.~Yang, ``Darts: Differentiable architecture
  search,'' \emph{arXiv preprint arXiv:1806.09055}, 2018.

\bibitem{pan2019}
C.~Gong, Z.~Jiang, D.~Wang, Y.~Lin, Q.~Liu, and D.~Z. Pan, ``Mixed precision
  neural architecture search for energy efficient deep learning,'' in
  \emph{2019 IEEE/ACM International Conference on Computer-Aided Design
  (ICCAD)}, 2019, pp. 1--7.

\bibitem{williams1992simple}
R.~J. Williams, ``Simple statistical gradient-following algorithms for
  connectionist reinforcement learning,'' \emph{Machine learning}, vol.~8,
  no.~3, pp. 229--256, 1992.

\bibitem{simonyan2014very}
K.~Simonyan and A.~Zisserman, ``Very deep convolutional networks for
  large-scale image recognition,'' \emph{arXiv preprint arXiv:1409.1556}, 2014.

\bibitem{zhang2017mixup}
H.~Zhang, M.~Cisse, Y.~N. Dauphin, and D.~Lopez-Paz, ``mixup: Beyond empirical
  risk minimization,'' \emph{arXiv preprint arXiv:1710.09412}, 2017.

\bibitem{inproceedings}
R.~Luo, F.~Tian, T.~Qin, E.~Chen, and T.-Y. Liu, ``Neural architecture
  optimization,'' 12 2018.

\bibitem{han2019design}
S.~Han, H.~Cai, L.~Zhu, J.~Lin, K.~Wang, Z.~Liu, and Y.~Lin, ``Design
  automation for efficient deep learning computing,'' \emph{arXiv preprint
  arXiv:1904.10616}, 2019.

\bibitem{he2018amc}
Y.~He, J.~Lin, Z.~Liu, H.~Wang, L.-J. Li, and S.~Han, ``Amc: Automl for model
  compression and acceleration on mobile devices,'' in \emph{Proceedings of the
  European conference on computer vision (ECCV)}, 2018, pp. 784--800.

\bibitem{cai2019once}
H.~Cai, C.~Gan, T.~Wang, Z.~Zhang, and S.~Han, ``Once-for-all: Train one
  network and specialize it for efficient deployment,'' \emph{arXiv preprint
  arXiv:1908.09791}, 2019.

\bibitem{guo2020single}
Z.~Guo, X.~Zhang, H.~Mu, W.~Heng, Z.~Liu, Y.~Wei, and J.~Sun, ``Single path
  one-shot neural architecture search with uniform sampling,'' in
  \emph{European Conference on Computer Vision}.\hskip 1em plus 0.5em minus
  0.4em\relax Springer, 2020, pp. 544--560.

\bibitem{cai2018proxylessnas}
H.~Cai, L.~Zhu, and S.~Han, ``Proxylessnas: Direct neural architecture search
  on target task and hardware,'' \emph{arXiv preprint arXiv:1812.00332}, 2018.

\bibitem{alizadeh1995interior}
F.~Alizadeh, ``Interior point methods in semidefinite programming with
  applications to combinatorial optimization,'' \emph{SIAM journal on
  Optimization}, vol.~5, no.~1, pp. 13--51, 1995.

\bibitem{baskin2021uniq}
C.~Baskin, N.~Liss, E.~Schwartz, E.~Zheltonozhskii, R.~Giryes, A.~M. Bronstein,
  and A.~Mendelson, ``Uniq: Uniform noise injection for non-uniform
  quantization of neural networks,'' \emph{ACM Transactions on Computer Systems
  (TOCS)}, vol.~37, no. 1--4, pp. 1--15, 2021.

\bibitem{liu2016ssd}
W.~Liu, D.~Anguelov, D.~Erhan, C.~Szegedy, S.~Reed, C.-Y. Fu, and A.~C. Berg,
  ``Ssd: Single shot multibox detector,'' in \emph{European conference on
  computer vision}.\hskip 1em plus 0.5em minus 0.4em\relax Springer, 2016, pp.
  21--37.

\bibitem{bulat2020high}
A.~Bulat, B.~Martinez, and G.~Tzimiropoulos, ``High-capacity expert binary
  networks,'' \emph{arXiv preprint arXiv:2010.03558}, 2020.

\bibitem{jung2019learning}
S.~Jung, C.~Son, S.~Lee, J.~Son, J.-J. Han, Y.~Kwak, S.~J. Hwang, and C.~Choi,
  ``Learning to quantize deep networks by optimizing quantization intervals
  with task loss,'' in \emph{Proceedings of the IEEE/CVF Conference on Computer
  Vision and Pattern Recognition}, 2019, pp. 4350--4359.

\bibitem{li2019additive}
Y.~Li, X.~Dong, and W.~Wang, ``Additive powers-of-two quantization: An
  efficient non-uniform discretization for neural networks,'' \emph{arXiv
  preprint arXiv:1909.13144}, 2019.

\bibitem{mishra2017wrpn}
A.~Mishra, E.~Nurvitadhi, J.~J. Cook, and D.~Marr, ``Wrpn: Wide
  reduced-precision networks,'' \emph{arXiv preprint arXiv:1709.01134}, 2017.

\bibitem{schulman2017proximal}
J.~Schulman, F.~Wolski, P.~Dhariwal, A.~Radford, and O.~Klimov, ``Proximal
  policy optimization algorithms,'' \emph{arXiv preprint arXiv:1707.06347},
  2017.

\bibitem{judd2016stripes}
P.~Judd, J.~Albericio, T.~Hetherington, T.~M. Aamodt, and A.~Moshovos,
  ``Stripes: Bit-serial deep neural network computing,'' in \emph{2016 49th
  Annual IEEE/ACM International Symposium on Microarchitecture (MICRO)}.\hskip
  1em plus 0.5em minus 0.4em\relax IEEE, 2016, pp. 1--12.

\bibitem{ye2018unified}
S.~Ye, T.~Zhang, K.~Zhang, J.~Li, J.~Xie, Y.~Liang, S.~Liu, X.~Lin, and
  Y.~Wang, ``A unified framework of dnn weight pruning and weight
  clustering/quantization using admm,'' \emph{arXiv preprint arXiv:1811.01907},
  2018.

\bibitem{lillicrap2015continuous}
T.~P. Lillicrap, J.~J. Hunt, A.~Pritzel, N.~Heess, T.~Erez, Y.~Tassa,
  D.~Silver, and D.~Wierstra, ``Continuous control with deep reinforcement
  learning,'' \emph{arXiv preprint arXiv:1509.02971}, 2015.

\bibitem{kingma2014adam}
D.~P. Kingma and J.~Ba, ``Adam: A method for stochastic optimization,''
  \emph{arXiv preprint arXiv:1412.6980}, 2014.

\bibitem{umuroglu2018bismo}
Y.~Umuroglu, L.~Rasnayake, and M.~Sj{\"a}lander, ``Bismo: A scalable bit-serial
  matrix multiplication overlay for reconfigurable computing,'' in \emph{2018
  28th International Conference on Field Programmable Logic and Applications
  (FPL)}.\hskip 1em plus 0.5em minus 0.4em\relax IEEE, 2018, pp. 307--3077.

\bibitem{howard2017mobilenets}
A.~G. Howard, M.~Zhu, B.~Chen, D.~Kalenichenko, W.~Wang, T.~Weyand,
  M.~Andreetto, and H.~Adam, ``Mobilenets: Efficient convolutional neural
  networks for mobile vision applications,'' \emph{arXiv preprint
  arXiv:1704.04861}, 2017.

\bibitem{nachum2018data}
O.~Nachum, S.~Gu, H.~Lee, and S.~Levine, ``Data-efficient hierarchical
  reinforcement learning,'' \emph{arXiv preprint arXiv:1805.08296}, 2018.

\bibitem{liu2013learning}
H.-Y. Liu and L.~P. Carloni, ``On learning-based methods for design-space
  exploration with high-level synthesis,'' in \emph{Proceedings of the 50th
  annual design automation conference}, 2013, pp. 1--7.

\bibitem{zhou2019primal}
Y.~Zhou, H.~Ren, Y.~Zhang, B.~Keller, B.~Khailany, and Z.~Zhang, ``Primal:
  Power inference using machine learning,'' in \emph{Proceedings of the 56th
  Annual Design Automation Conference 2019}, 2019, pp. 1--6.

\bibitem{wang2018two}
P.~Wang, Q.~Hu, Y.~Zhang, C.~Zhang, Y.~Liu, and J.~Cheng, ``Two-step
  quantization for low-bit neural networks,'' in \emph{Proceedings of the IEEE
  Conference on computer vision and pattern recognition}, 2018, pp. 4376--4384.

\bibitem{liu2019circulant}
C.~Liu, W.~Ding, X.~Xia, B.~Zhang, J.~Gu, J.~Liu, R.~Ji, and D.~Doermann,
  ``Circulant binary convolutional networks: Enhancing the performance of 1-bit
  dcnns with circulant back propagation,'' in \emph{Proceedings of the IEEE/CVF
  Conference on Computer Vision and Pattern Recognition}, 2019, pp. 2691--2699.

\bibitem{shen2020balanced}
M.~Shen, X.~Liu, R.~Gong, and K.~Han, ``Balanced binary neural networks with
  gated residual,'' in \emph{ICASSP 2020-2020 IEEE International Conference on
  Acoustics, Speech and Signal Processing (ICASSP)}.\hskip 1em plus 0.5em minus
  0.4em\relax IEEE, 2020, pp. 4197--4201.

\bibitem{pruningsurvey}
R.~Reed, ``Pruning algorithms-a survey,'' \emph{IEEE Transactions on Neural
  Networks}, vol.~4, no.~5, pp. 740--747, 1993.

\end{thebibliography}
\end{document}